\documentclass{article} 
\usepackage{main,times}

\usepackage{amsmath,amsfonts,bm}

\def\eqref#1{equation~\ref{#1}}

\def\1{\bm{1}}

\DeclareMathAlphabet{\mathsfit}{\encodingdefault}{\sfdefault}{m}{sl}
\SetMathAlphabet{\mathsfit}{bold}{\encodingdefault}{\sfdefault}{bx}{n}

\usepackage{hyperref}
\usepackage{url}
\usepackage{graphicx} 
\usepackage{amsmath}  
\usepackage{subcaption} 
\usepackage{amssymb}    
\usepackage{algorithm}    
\usepackage{algorithmic}
\usepackage{color}
\usepackage{multirow}

\newtheorem{lemma}{Lemma}
\newtheorem{definition}{Definition}
\newenvironment{proof}{{\noindent\bf Proof}}{\hfill $\square$\par}
\usepackage[encapsulated]{CJK}
\usepackage{makecell}

\usepackage{soul}
\usepackage{comment}
\usepackage{stackengine}

\usepackage{pifont}
\usepackage{wrapfig}
\usepackage{bbding}
\usepackage{threeparttable}
\usepackage{tcolorbox}
\usepackage{verbatim}
\usepackage{listings}
\usepackage{courier}
\tcbuselibrary{listings, breakable}
\usepackage{xcolor} 
\definecolor{cyan}{RGB}{154,201,219}
\definecolor{red}{RGB}{255,0,0}
\definecolor{flesh}{RGB}{192,0,0}
\definecolor{myb}{RGB}{46, 152, 140}
\definecolor{myg}{RGB}{26, 173, 25}
\definecolor{orange}{RGB}{237,125,49}
\usepackage{booktabs}
\lstdefinelanguage{plainjson}{
  basicstyle=\ttfamily\fontsize{5}{2},
  breaklines=true,
  showstringspaces=false,
  escapeinside={(*@}{@*)} 
}
\usepackage[title,titletoc]{appendix}
\usepackage{titletoc}
\usepackage[table]{xcolor}
\usepackage{colortbl}

\title{Local-First Intelligence via Proactive SLM Collaborating with Feedback-Providing LLM}

\author{Hang Zeng$^{1}$ \quad
Xiangyu Liu$^{2}$ \quad
Yong Hu$^{2}$ \quad
Chaoyue Niu$^{1}$\thanks{Correspondence to: Chaoyue Niu  (\texttt{rvince@sjtu.edu.cn})} \quad
Jiarui Zhang$^{1}$ \\
\bf
Shaojie Tang$^{3}$ \quad
Fan Wu$^{1}$ \quad
Guihai Chen$^{1}$ \\[6pt]
$^{1}$Shanghai Jiao Tong University, Shanghai, China \\
$^{2}$WeChat, Tencent, Beijing, China \\
$^{3}$State University of New York at Buffalo, NY, USA}

\iclrfinalcopy
\begin{document}

\maketitle

\begin{abstract}
Personal intelligence requires strong capabilities under strict privacy and efficiency constraints. Unlike cloud-first intelligence, which relies on powerful but costly and privacy-vulnerable large language models (LLMs), we focus on local-first intelligence, emphasizing on-device small language models (SLMs), which enable efficient, private inference but have limited capacity. To reconcile the capability gap with privacy and efficiency, we propose a dynamic collaboration framework, where an SLM learns to proactively decide how to consult an LLM during multi-step reasoning, while the LLM provides adaptive feedback instead of acting as a passive tool, modeling them as heterogeneous agents under information asymmetry. We further systematically investigate how collaboration strategies are shaped by SLM and LLM capabilities as well as efficiency and privacy constraints. Evaluation results reveal a distinct scaling effect: stronger SLMs become more self-reliant, while stronger LLMs enable fewer and more informative interactions. In addition, the learned dynamic collaboration strategies significantly outperform various baselines and transfer robustly to unseen LLMs as well as to real-world personal tasks without additional training.
\end{abstract}

\section{Introduction}

Personal intelligence is emerging as a key paradigm for delivering context-aware assistance grounded in user-specific data, requiring strong performance, robust privacy preservation, and efficient interaction, as reflected in commercial systems like Gemini Personal Intelligence \citep{google2026gemini} and Apple Intelligence \citep{apple2026intelligence}. Existing cloud-first intelligence prioritizes powerful large language models (LLMs), achieving strong performance but introducing significant challenges in privacy and cost. This motivates local-first intelligence, where on-device small language models (SLMs) serve as the first point of contact for user queries, offering lower cost and stronger privacy guarantees. However, local SLMs generally lack the strong capability and broad knowledge required to handle complex queries.

To enable powerful local-first intelligence, a natural solution is to integrate on-device SLMs with cloud-based LLMs. Existing approaches typically rely on static interactions \citep{LLM_cascade,collaborate3,query_rewriting1}, which can be categorized by how user queries are exposed to the LLM. The first category directly sends unresolved queries or sub-queries to the LLM, as in AutoMix \citep{AutoMix} and Router-R1 \citep{router_r1}. However, these methods either ignore efficiency or rely on predefined heuristics, leading to unnecessary or missed LLM calls while overlooking privacy concerns. The second category masks private information before forwarding queries to the LLM, as in Self-Disclosures \citep{selfdisclosures}, yet static privacy filtering may either remove critical information and degrade answer quality, or fail to fully filter sensitive content. The third category rewrites queries with SLM before invoking the LLM and post-processes its responses, as in PAPILLON \citep{papillon}. However, fixed interaction policies and limited awareness of the LLM's capabilities prevent the SLM from adaptively deciding when and what to consult, causing irrelevant or insufficient LLM calls. Overall, these methods treat the LLM as an external passive tool invoked through predefined pipelines, precluding the bidirectional interaction where the SLM adaptively requests assistance and the LLM provides targeted knowledge or feedback.

\begin{figure}[!t]
  \centering
         \centering
         \begin{subfigure}{0.49\linewidth}
             \centering
            \begin{minipage}{\linewidth}
             \includegraphics[width=\textwidth]{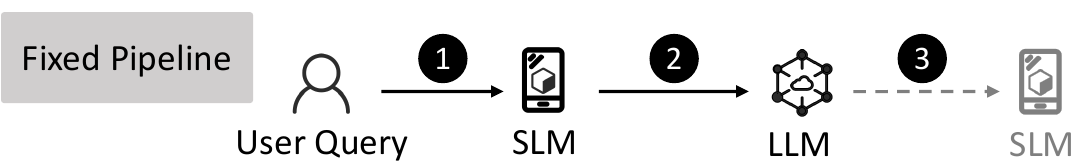}
            \end{minipage}
             \caption{Static Interaction Framework}
             \label{fig:static_workflow}
        \end{subfigure}
        \hfill
        \begin{subfigure}{0.49\linewidth}
             \centering
            \begin{minipage}{\linewidth}
             \includegraphics[width=\textwidth]{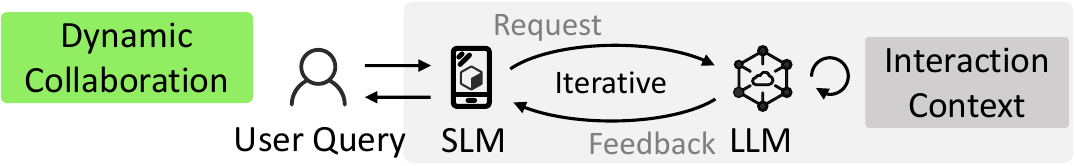}
            \end{minipage}
             \caption{Dynamic Collaboration Framework}
             \label{fig:dynamic_workflow}
        \end{subfigure}
    \caption{Static interaction framework in existing work vs. Our dynamic collaboration framework.}\label{fig:workflow_comapre} 
    \vspace{-1.2em}
\end{figure}

In this work, we propose a novel dynamic collaboration framework where both on-device SLMs and cloud-based LLMs are modeled as inherently heterogeneous agents with different capabilities and asymmetric information. As illustrated in Figure \ref{fig:workflow_comapre}, the SLM serves as the lead coordinator, first receiving the user query, conducting iterative reasoning, and adaptively deciding whether to proceed locally or consult the LLM. Ideally, the SLM formulates targeted requests only when assistance is needed and completes the reasoning independently otherwise. Crucially, the LLM is not treated as a black-box tool but as a flexible source of feedback and knowledge that participates in a multi-turn interaction with the SLM. 
At each interaction step, the LLM can take one of two actions: (1) provide feedback when the request is ambiguous, or (2) return relevant and useful information.
We train the SLM through end-to-end online reinforcement learning (RL), enabling it to autonomously learn interaction timing and request formulation, conditioned on task difficulty and LLM feedback.

To understand the factors shaping the collaboration strategies, we systematically investigate different research questions. We evaluate how the SLM and LLM capabilities affect the effectiveness and dynamics of collaboration. We also examine whether the learned strategies enable richer and more effective interaction patterns than static interaction frameworks, whether they transfer to real-world personal tasks, whether SLMs can acquire privacy-preserving collaboration behaviors without sacrificing performance, and how external efficiency and privacy constraints affect strategies.

Our findings reveal several key insights. SLM and LLM capabilities shape collaboration strategies: stronger SLMs become more selective and self-reliant, while more capable LLMs support more concise and effective interactions. Notably, the learned strategies transfer effectively to unseen LLMs and further generalize to real-world personal tasks. Moreover, compared to static interaction frameworks, dynamic collaboration enables more efficient behavior by selectively invoking the LLM only when necessary, leading to improved performance and stronger privacy preservation.

We summarize our key contributions as follows:
(1) We formulate local-first personal intelligence as a novel dynamic collaboration problem between on-device SLMs and cloud-based LLMs with heterogeneous capabilities and asymmetric information.
(2) We propose a proactive SLM that adaptively decides when and how to collaborate, a collaborative LLM with a closed-loop feedback mechanism, and a multi-objective reward jointly optimizing response quality, efficiency, and privacy.
(3) We construct the PrivQA, a benchmark of 153K privacy-augmented QA spanning general and multi-hop questions, and conduct systematic experiments, providing actionable insights into collaboration behaviors. Our method consistently improves EM by 3.0\% -- 15.6\%, reduces interaction turns by 0.11 -- 0.27, and lowers privacy leakage by 20.9\% -- 74.1\% over static baselines. The learned policy also directly transfers to real-world personal tasks without additional training, improving quality by 3.1\% -- 6.9\% and lowering privacy leakage by 5.4\% -- 26.0\% over baselines.
\section{Related Work}

Collaboration between SLM and LLM has been studied mainly under two paradigms: cloud-first and local-first intelligence. Cloud-first approaches rely on LLMs as the primary inference engine, augmenting them with tools or domain-specific small models \citep{search_r1} or routing among multiple LLMs based on query characteristics \citep{router1,router2,router4,router5}. While effective, these methods remain centered on cloud-based LLMs and therefore inherit their privacy risks and efficiency concerns. 
Local-first approaches instead place an SLM on the user side as the primary interface. Most existing methods adopt a fixed pipeline in which the SLM preprocesses the query once and forwards it to the LLM \citep{Reducto,collaborate1,collaborate3,CoGenesis}, either through query rewriting, as in PAPILLON \citep{papillon} and PrivacyRestore \citep{query_rewriting1}, or through privacy filtering \citep{privacy_preprocessing1,privacy_preprocessing2}, exemplified by Self-Disclosures \citep{selfdisclosures}. A more recent line of work trains the SLM as an agentic router: Privacy-R1 \citep{privacy_r1} dispatches non-sensitive chunks to the LLM, and Router-R1 \citep{router_r1} routes sub-queries across multiple LLMs. Despite differing implementations, existing methods rely on a fixed interaction framework, treat the LLM as a passive tool, and address at most two of the three coupled requirements: response quality, interaction efficiency, and privacy preservation. 
In contrast, we propose a dynamic framework with closed-loop feedback in which the proactive SLM adaptively decides when and how to consult, and the LLM can either provide information or ask for context, jointly optimizing all three requirements under information asymmetry between heterogeneous agents. Appendix~\ref{sec:baseline_delta} elaborates these methodological deltas.

Another related line of work is Multi-Agent Systems (MASs), where autonomous entities interact in a shared environment to achieve individual or collective goals \citep{Cooperative_ai, mas22, mas23, mas_survey}. LLM-driven MASs have enabled agents to reason, plan, and communicate in natural language \citep{llm_agent1,llm_aget_survey,mas_plan}, as exemplified by MetaGPT \citep{MetaGPT} and AgentVerse \citep{AgentVerse}. These systems typically require agents to share task descriptions, execution traces, or environmental information. For local-first collaboration, however, models are heterogeneous in capability and subject to information asymmetry, as the on-device SLM and cloud-based LLM differ in capability, efficiency, and privacy exposure, and the LLM only accesses information selectively disclosed by the SLM. Appendix \ref{sec:existing_framework} further provide a broader comparison with existing work.

\begin{figure}[!t]
\centering

\begin{minipage}[t]{0.55\linewidth}
    \centering
    \includegraphics[width=\linewidth]{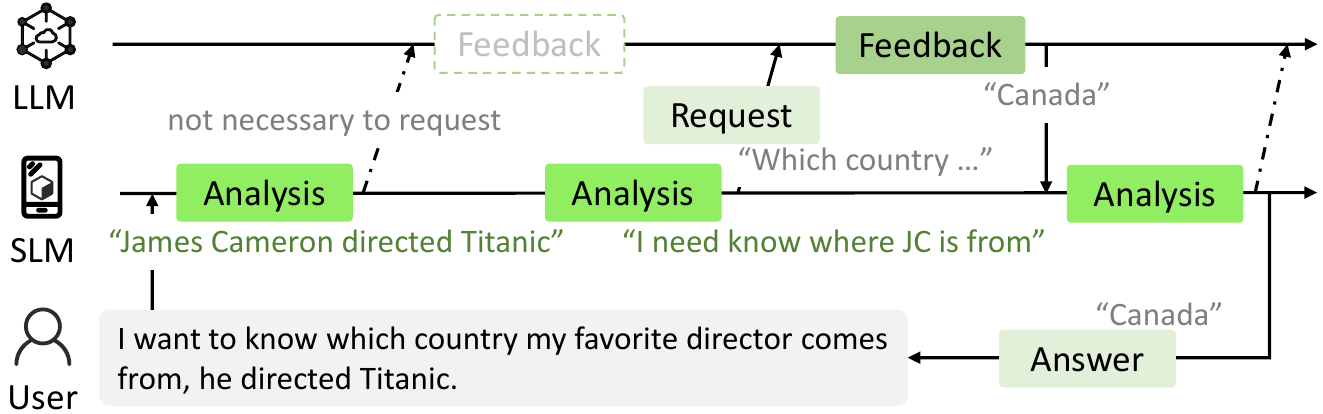}
    \caption{Illustration of dynamic collaboration workflow between proactive SLM and cloud-based LLM.}
    \label{fig:workflow}
\end{minipage}
\hfill
\begin{minipage}[t]{0.44\linewidth}
    \centering
    \includegraphics[width=\linewidth]{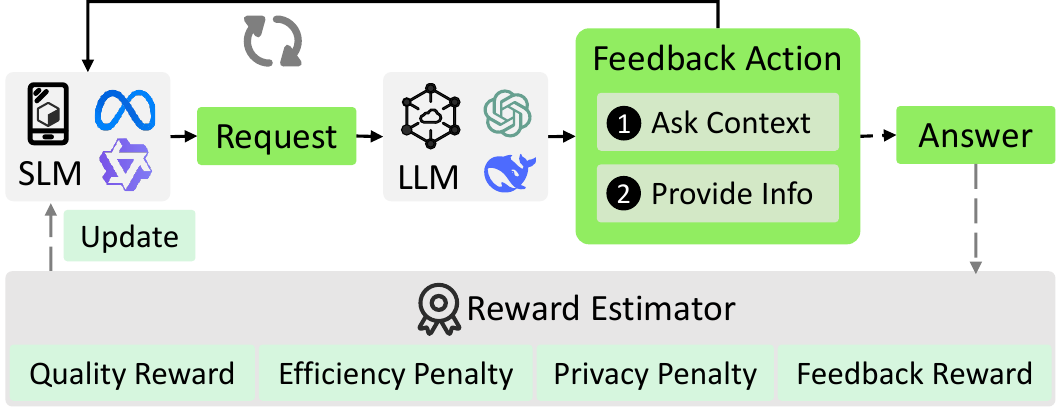}
    \caption{Training with LLM feedback via outcome- and process-based rewards.} %
    \label{fig:collaboration}
\end{minipage}
\vspace{-1.2em}
\end{figure}

\section{Problem Formulation}\label{sec:problem_formulation}

\subsection{Workflow Formulation}
We focus on local-first intelligence, a paradigm in which on-device SLMs act as the primary interface for user queries.
As shown in Figure \ref{fig:workflow}, when the SLM receives a query, it enters an iterative multi-step reasoning loop: at each step, it assesses whether the current subproblem can be solved locally with high confidence, and otherwise requests the required information from a more capable cloud-based LLM. After the LLM responds and provides feedback, the SLM integrates the returned information and continues solving subsequent steps until it produces a final answer for the user.
For example, given the privacy-sensitive query ``I want to know which country my favorite director comes from, he directed Titanic,'' the SLM first identifies the subproblem ``Who directed Titanic?'' and answers ``James Cameron'' locally, then forwards the next subproblem ``Which country is James Cameron from?'' to the LLM. After receiving ``Canada,'' the SLM integrates the information and returns the final response.
Throughout this process, only the SLM has access to the user's query and decides how to interact with the LLM.

We model the workflow in which an SLM autonomously requests assistance from an LLM as a collaborative process between heterogeneous agents with asymmetric capabilities and information availability.
Formally, let $M_s$ denote the on-device SLM and $M_l$ denote the cloud-based LLM. Given a user query $Q$, which is available only to $M_s$, the SLM’s behavior is modeled as a sequence of triplets $\langle \tau, a, o \rangle$. At step $t$, $\tau_t$ is the reasoning process, $a_t$ denotes an action available to $M_s$, including making structured requests as well as generating the answer, and $o_t$ is the observation received after each action, which consists of feedback provided by $M_l$.
The trajectory terminates when $M_s$ produces the final answer.
For $M_s$, the trajectory history at step $t$ is defined as
\begin{equation*}
\mathcal{H}_t = \{\tau_0, a_0, o_0, \tau_1, a_1, o_1, \dots, \tau_{t-1}, a_{t-1}, o_{t-1}\}.
\end{equation*}
$M_s$ generates $\tau_t$ and $a_t$ according to the collaboration policy given $\mathcal{H}_t$ and $Q$, and $M_l$ generates $o_t$ based on $a_t$ issued by $M_s$ and the interaction context between $M_s$ and $M_l$. 

\subsection{Overview of Collaboration Goal}

Local-first intelligence naturally poses three coupled objectives for the autonomous collaboration between SLMs and LLMs, namely delivering high-quality responses, interacting efficiently, and minimizing privacy leakage. Response quality is evaluated by comparing the SLM’s predictions with ground-truth answers for each user query. Interaction efficiency is analyzed in terms of the average number of interaction turns and the necessity of the SLM’s requests to the LLM. Privacy leakage is measured by assessing whether any requests issued by the SLM expose information from the set of user private data.
Beyond these optimization objectives, we further investigate how different SLMs autonomously learn effective collaboration strategies with different LLMs under tasks with varying difficulty and constraints. We systematically examine how various factors influence the collaboration strategies adopted by SLMs.

\section{Proactive SLM with LLM Feedback}

We design dynamic collaboration between heterogeneous SLM and LLM under information asymmetry, where SLM learns to proactively consult LLM with active feedback, as shown in Figure~\ref{fig:collaboration}.

\subsection{Outcome Quality and Interaction Necessity Driven SLM Request}

The SLM is trained to learn a collaboration policy $\pi_{\theta}(y \mid x, M_l)$, where $x = Q$ is the user query and $y = {(\tau_t, a_t, o_t)}_{t=0}^{T}$ is the interaction trajectory, with $\tau_t$ and $a_t$ generated by $M_s$ and $o_t$ returned by $M_l$. It must decide when to query the LLM and how to formulate each request, based on the respective capabilities of $M_s$ and $M_l$ and the information required for the task, jointly optimizing response quality and interaction efficiency.

For response quality, we adopt an outcome-based quality reward $r_{\text{quality}}$ to encourage the SLM to produce high-quality answers. Formally, $r_{\text{quality}}$ is defined as the Exact Match (EM) score between the prediction $ans_{\text{pred}}$ and the ground-truth $ans_{\text{gt}}$ as follows.
\begin{equation*}
    r_{\text{quality}} = \text{EM}(ans_{\text{pred}}, ans_{\text{gt}})
\end{equation*}

For efficiency, the SLM is encouraged to issue requests only when necessary, i.e., when it genuinely lacks the required knowledge. We design a difficulty-aware efficiency penalty $r_{\text{efficiency}}$. {\bf For easy queries, the SLM is expected to answer without issuing any requests. For hard queries, the SLM may issue requests but should minimize the number of interaction turns to obtain the correct answer.} We employ a simple but effective difficulty judgment strategy: if the SLM can correctly answer the user query without relying on external information, the query is considered easy; otherwise, it is considered hard. Formally, $r_{\text{efficiency}}$ is defined as
\begin{equation*}
\small
r_{\text{efficiency}} = -\mathbb{I}(Q \notin D_{\text{e}})\frac{n_r}{n_{max}} - \mathbb{I}(Q \in D_{\text{e}} \land n_r > 0),
\end{equation*}
where $Q \in D_{\text{e}}$ indicates that the query is easy for the current SLM, $n_r$ denotes the total number of interaction turns, and $n_{\max}$ denotes the predefined maximum number of interaction turns.

\subsection{Active LLM Feedback}

Prior collaboration frameworks treat the LLM as a passive tool that returns information for whatever request the SLM sends. This is problematic when the SLM, constrained by its limited capability, issues requests that are vague or under-specified: the LLM may return superficially relevant but ultimately noisy information, causing errors to accumulate over multiple turns. We instead model the LLM as an active collaborator whose response adapts to the precision of the SLM's request.
At each interaction step $t$, the LLM takes the SLM's request $a_t$ together with the interaction history and selects one of two actions: {\bf (1) Clarify if $a_t$ is under-specified, asking the SLM to provide the missing context; or (2) Answer if $a_t$ is well-specified, returning the relevant information}. The selected action yields the observation $o_t$ that is fed back into the SLM's trajectory. The action decision is made via a prompted rubric detailed in Appendix~\ref{sec:prompt_framework} and the LLM is not further trained.

Furthermore, the LLM's feedback also serves as a process-level learning signal for the SLM: {\bf a request that prompts a clarification is by construction under-informative, so we penalize it through a feedback reward $r_{\text{feedback}}$.} For example, the request ``I need to know where the user's cousin lives'' would elicit a clarification and receive $r_{\text{feedback}}=-1$, teaching the SLM to issue more concrete requests in subsequent turns. Formally, $r_{\text{feedback}}$ is defined as 
\begin{equation*}
r_{\text{feedback}} = \left\{ 
\begin{aligned}
-1,&\ \text{LLM asks contextual information}\\
 0,&\ \text{otherwise}
\end{aligned}
\right.
\end{equation*}

\subsection{Privacy Awareness with Dataset Synthesis}

To ensure privacy-preserving collaboration, the SLM needs to learn to formulate requests without disclosing sensitive user information. We address this by introducing a privacy penalty, $r_{\text{privacy}}$, which is applied to the interaction whenever an SLM request results in a privacy leak: 
\begin{equation*}
    r_{\text{privacy}} = \left\{ 
    \begin{aligned}
    -1,&\ \text{if the request leaks privacy}\\
     0,&\ \text{otherwise}.
    \end{aligned}
    \right.
\end{equation*}
Whether privacy is leaked is judged by a separate evaluator LLM \citep{llm_as_a_judge}. Given the labeled private information in the user query and the requests generated by the SLM across all interaction turns, the evaluator LLM determines whether any private information can be inferred, ensuring multi-turn, compositional, and indirect leakage is covered. Under an honest-but-curious LLM adversary, Appendix~\ref{sec:eval_metric_def} provides the full formulation. Human evaluation on 200 instances agrees with the evaluator's judgments 98.5\% of the time. Appendix~\ref{sec:human_eval} details the protocol and further shows that judgments are consistent across evaluators from different model families.

To train the SLM's privacy awareness, it requires data containing user privacy information. However, existing privacy QA datasets \citep{medqa, papillon} contain only a few hundred samples primarily designed for evaluation and are limited in diversity. We therefore build a privacy injection pipeline that {\bf augments existing QA data by prompting a synthesis LLM to replace explicit entities with privacy-related indirect references, simulating inadvertent user disclosure}. As shown in Figure~\ref{fig:dataset}, given an original QA pair, we first generate logically consistent personal information related to the core entities; we then reformulate the underlying factual statement so that the answer term is replaced by an indirect reference through this personal information; finally, we combine the reformulated fact with a new question derived from the personal information to obtain a privacy-infused QA pair with an unchanged correct answer. The injected information is thus a necessary premise for reaching the answer, not a separable decoration. The synthesis LLM subsequently filters cases where the answer is altered or explicitly revealed in the question. We further conduct human evaluation on 200 randomly sampled instances, confirming that 99\% are correctly reformulated with accurate answers and appropriate privacy infusion.

Applying this pipeline, we build the Privacy QA (PrivQA) dataset from six public QA sources: three general-domain datasets (NQ \citep{nq}, TriviaQA \citep{triviaqa}, PopQA \citep{popqa}) and three multi-hop datasets (HotpotQA \citep{hotpotqa}, 2WikiMultihopQA \citep{2wiki}, MuSiQuE \citep{musique}), which are collected from genuine user queries such as Google search logs and TREC. Each PrivQA instance contains the privacy-augmented question, its ground-truth answer, and the synthesized private information used for quantitative leakage evaluation. Appendix~\ref{sec:dataset_statistics} presents the pipeline details and dataset statistics.

\subsection{Collaboration Strategy Learning}\label{sec:collab_strategy_learning}

We finally integrate the goals of response quality, efficiency, and privacy into a multi-objective reward function. We additionally include a format reward $r_{\text{format}} \in \{-1, 0\}$ that penalizes format violations, as shown in Appendix \ref{sec:foramt_slm}. The final reward function is defined as 
\begin{equation*}
    R = r_{\text{format}} + \underbrace{r_{\text{quality}} + r_{\text{feedback}}}_{\text{answer reward}} + \underbrace{\alpha r_{\text{privacy}} + \beta r_{\text{efficiency}}}_{\text{constraint penalty}}.
\end{equation*}
where $\alpha, \beta \geq 0$ trade off quality against privacy and efficiency. Larger weights induce more privacy-preserving or more efficient collaboration, allowing the framework to adapt to different deployment requirements.
Appendix~\ref{sec:noise_robust} proves that reward judgment errors do not change the gradient direction.

Given the reward, the SLM is optimized using the latest algorithm, VAPO \citep{vapo}. The policy model is updated to maximize the following objective:
\begin{equation*}
\small
\begin{aligned}
\mathcal{J_{\mathrm{VAPO}}}(\theta) = \mathbb{E}_{x\sim \mathcal{D}, y\sim \pi_{\theta_\text{old}}(\cdot\mid x;o)}
\Bigg[\frac{1}{\sum_{t=1}^{|y|}\mathbb{I}(y_t \neq o)} \sum_{t=1}^{|y|} 
\min \Big( p_{t}(\theta) \hat{A}_{t}, \text{clip} \Big( p_{t}(\theta), 1 - {\varepsilon_{low}}, 1 + {\varepsilon_{high}} \Big) \hat{A}_{t} \Big) \Bigg],
\label{eq:vapoloss}
\end{aligned}
\end{equation*}
where $p_{t}(\theta) = \frac{\pi_\theta\bigl(y \mid x,\,y_{<t}; o\bigr)}{\pi_{\theta_{\mathrm{old}}}\bigl(y \mid x,\,y_{<t}; o\bigr)}$ is the probability ratio, the terms $\varepsilon_{low}$ and $\varepsilon_{high}$ are the lower and higher clipping range hyperparameter, and $\hat{A}_{t}$ denotes the estimated advantage at time step $t$, calculated using Length-Adaptive GAE \citep{vapo} based on future rewards $R{\geq t}$ and a learned value model. $\pi_\theta$ and $\pi_{\theta_{\mathrm{old}}}$ are the current and previous policy models, respectively. The indicator function $\mathbb{I}(y_t \neq o)$ masks the token loss for $o$ generated by $M_l$. 

\section{Evaluation}\label{sec:evaluation}

\subsection{Experimental Setup}

\textbf{Model setup.}
For the SLMs, we employed Qwen3-4B and Qwen3-8B \citep{qwen3} with reasoning capabilities, as well as Llama-3.2-3B-Instruct and Llama-3.1-8B-Instruct \citep{llama}. For the LLMs, we utilized the open-source DeepSeek-R1-Distill-Llama-70B \citep{deepseek_r1} and Qwen3-235B-A22B-Instruct, along with the closed-source Qwen3-Max. We pair different SLMs and LLMs for systematic evaluation.

\textbf{Dataset setup.}
We train the SLM on the PrivQA dataset, which consists of privacy-augmented QA samples sourced from NQ, TriviaQA, PopQA, HotpotQA, 2Wiki, and MuSiQuE. To thoroughly evaluate the generalization capability of the learned collaboration strategy, we construct the training set using only samples derived from NQ and HotpotQA, while the test set covers all six datasets. We additionally evaluate the learned policy on three real-world benchmarks without any further training, namely PUPA \citep{papillon} for email and document writing, AESLC \citep{aeslc} for email subject-line writing, and SAMSum \citep{samsum} for dialogue summarization.

\textbf{Evaluation metrics.}
For quality, we employ Exact Match (EM) and BERTScore ($F_{\text{BERT}}$) \citep{bertscore} on PrivQA. On the three real-world personal tasks, which are open-ended generation tasks, we adopt LLM-as-a-judge with prompted rubrics \citep{llm_as_a_judge}.

For efficiency, we report the average number of interaction turns. We also introduce the Interaction Necessity Score (INScore), a difficulty-aware metric. Queries are labeled easy or hard depending on whether the SLM can answer them independently. Using difficulty as ground truth and the SLM's request decision as prediction, INScore is the average of two recalls: the proportion of easy queries handled locally and the proportion of hard queries for which assistance is requested. Since easy or hard labels are calibrated per SLM, INScore is a within-SLM efficiency metric. 

For privacy, we compute the average semantic similarity between the SLM's requests and the set of privacy information with BERTScore ($\text{Priv}_{\text{sim}}$) \citep{query_rewriting1} and measure the proportion of samples where the SLM issues privacy-leaking requests ($\text{Priv}_{\text{sample}}$) with an LLM judge \citep{papillon}. Detailed definitions of metrics are provided in Appendix \ref{sec:eval_metric_def}.

\textbf{Baselines.} We compare against three groups.
\textbf{Non-collaboration.} (1) SLM with Chain of Thought (CoT)~\citep{cot}; (2) direct LLM inference~\citep{gpt3}; (3) LLM with CoT, the LLM-side upper bound; (4) SFT on PrivQA data, answering locally.
\textbf{Static collaboration.} (5) AutoMix \citep{AutoMix}, forwarding queries the SLM cannot resolve; (6) Self-Disclosures \citep{selfdisclosures}, masking private tokens before forwarding; (7) PAPILLON \citep{papillon}, SLM query rewriting followed by an LLM answer and SLM aggregation; (8) Router-R1 \citep{router_r1}, an SLM trained via RL to dispatch sub-queries.
\textbf{Ablation.} (9) Ours w/o Feedback, our framework in which  the LLM acts as a passive answer-only tool, trained with the same reward.

\textbf{Implementation details.}
We implement all designs with verl framework \citep{verl}. The workstation has 8 H20 96G GPUs. More details are listed in Appendix \ref{sec:implementation_details}.

\subsection{Gains of Dynamic Collaboration}

\begin{table*}
\centering
\caption{Quality, privacy leakage, and efficiency performance on the PrivQA dataset, including in- and out-domain datasets. \textbf{Bold} denotes the best result and \underline{underline} the second-best.}\label{tab:baseline_resutl}
\resizebox{1\linewidth}{!}{
    \begin{tabular}{l|l|lll|lll|lll}
    \toprule
        \multirow{2}{*}{Method} & \multirow{2}{*}{Model}  & \multicolumn{3}{c|}{General QA} & \multicolumn{3}{c|}{Multi-Hop QA}  & \multicolumn{3}{c}{Avg.}\\
        \cmidrule{3-11}
        ~ & ~ & EM $\uparrow$ & $\text{Priv}_{\text{sample}}$ $\downarrow$ & INScore $\uparrow$ & EM $\uparrow$ &  $\text{Priv}_{\text{sample}}$ $\downarrow$ & INScore $\uparrow$ & EM $\uparrow$ &   $\text{Priv}_{\text{sample}}$ $\downarrow$ & INScore $\uparrow$ \\
        \midrule
        CoT &  Qwen3-235B & 50.89 & 100.00 & 50.00 & 34.11 & 100.00 & 50.00 & 42.50 & 100.00 & 50.00 \\
        \midrule
        \multirow{2}{*}{\makecell[l]{SFT}} & Qwen3-4B & 25.73 & 0.00 & 50.00 & 19.70 & 0.00 & 50.00 & 22.72 $_{(-15.6)}$ & 0.00 $_{(+0.3)}$ & 50.00 $_{(-9.6)}$ \\
        ~ & Qwen3-8B & 33.74 & 0.00 & 50.00 & 21.85 & 0.00 & 50.00 & 27.80 $_{(-15.3)}$ & 0.00 $_{(+0.1)}$ & 50.00 $_{(-10.9)}$ \\
        \midrule
        \multirow{2}{*}{\makecell[l]{AutoMix}} & Qwen3-4B & 37.65 & 72.34 & 58.44 & 25.93 & 76.28 & 54.78 & 31.79 $_{(-6.6)}$	&	74.31 $_{(-74.1)}$	&	56.61 $_{(-3.0)}$ \\
        ~ & Qwen3-8B   & 41.10 & 66.12 & 59.51 & 26.45 & 71.68 & 56.62 & 33.78 $_{(-9.3)}$	&	68.90 $_{(-68.8)}$	&	58.06 $_{(-2.8)}$ \\
         \midrule
        \multirow{2}{*}{\makecell[l]{Self-Disclosures}} & Qwen3-4B  & 32.57 & 59.60 & 50.00 & 18.43 & 57.59 & 50.00 & 25.50 $_{(-12.8)}$	&	58.60 $_{(-58.3)}$	&	50.00 $_{(-9.6)}$ \\
        ~ & Qwen3-8B  & 36.08 & 53.86 & 50.00 & 22.09 & 49.22 & 50.00 & 29.09 $_{(-14.0)}$	&	51.54 $_{(-51.5)}$	&	50.00 $_{(-10.9)}$ \\
         \midrule
         \multirow{2}{*}{\makecell[l]{PAPILLON}} & Qwen3-4B & 41.43 & 34.99 & 50.00 & 23.68 & 30.30 & 50.00 & 32.55 $_{(-5.8)}$	&	32.64 $_{(-32.4)}$	&	50.00 $_{(-9.6)}$ \\
        ~ & Qwen3-8B &  42.55 & 23.88 & 50.00 & 23.91 & 24.87 & 50.00 & 33.23 $_{(-9.9)}$	&	24.37 $_{(-24.3)}$	&	50.00 $_{(-10.9)}$ \\
         \midrule
        \multirow{2}{*}{\makecell[l]{Router-R1}} & Qwen3-4B & 43.68 & 30.25 & 52.32 & 27.06 & 27.34 & 53.28 & 35.37 $_{(-3.0)}$	&	28.80 $_{(-28.5)}$	&	52.80 $_{(-6.8)}$ \\
        ~ & Qwen3-8B & 47.29 & 22.04 & 55.78 & \underline{28.62} & 19.99 & {55.27} & 37.96 $_{(-5.2)}$	&	21.01 $_{(-20.9)}$	&	55.52 $_{(-5.4)}$ \\
        \midrule
        \multirow{2}{*}{Ours} & Qwen3-4B & \underline{48.47} & \underline{0.18} & \underline{60.98} & 28.21 & \underline{0.33} & \underline{58.17} & \underline{38.34} $_{(+0.0)}$ & \underline{0.26} $_{(+0.0)}$ & \underline{59.57} $_{(+0.0)}$ \\
        ~ & Qwen3-8B  & \textbf{52.86} & \textbf{0.08} & \textbf{61.82} & \textbf{33.35} & \textbf{0.08} & \textbf{59.99} & \textbf{43.11} $_{(+0.0)}$ & \textbf{0.08} $_{(+0.0)}$ & \textbf{60.91} $_{(+0.0)}$ \\
        \bottomrule
    \end{tabular}
 }
\vspace{-1em}
\end{table*}

\textbf{RQ1.} Does the learned collaboration strategy lead to better interaction patterns than baselines across both PrivQA and unseen real-world personal tasks?

\textbf{Our dynamic collaboration framework simultaneously improves response quality, privacy preservation, and interaction efficiency over all baselines.} Table \ref{tab:baseline_resutl} compares our dynamic collaboration framework with several baselines on PrivQA. Although static interaction methods improve over SLM inference by leveraging LLM assistance, they still lag behind standalone LLM inference. For instance, Router-R1 with Qwen3-8B is 4.5\% lower in EM. In contrast, our framework achieves performance comparable to, or even better than, LLM inference. Compared to SFT, AutoMix, Self-Disclosures, PAPILLON, and Router-R1, on Qwen3-8B, our method improves EM by 15.3\%, 9.3\%, 14.0\%, 9.9\%, and 5.2\%, respectively.
In terms of privacy, our method almost eliminates leakage: $\text{Priv}_{\text{sample}}$ is only 0.26\% on Qwen3-4B and 0.08\% on Qwen3-8B, far below the best baselines of 28.80\% and 21.01\%, respectively. This is attributed to the SLM’s ability to reformulate private queries into objective, user-agnostic requests. Moreover, by adaptively deciding when to request assistance, our method improves INScore on Qwen3-8B over the baselines by 10.9\%, 2.8\%, 10.9\%, 10.9\%, and 5.4\%, respectively, demonstrating better interaction efficiency.

\begin{wraptable}{r}{6.5cm}
\centering
\vspace{-1.2em}
\caption{Average performance and privacy leakage on real-world personal tasks.}\label{tab:real_world}
\vspace{-0.4em}
\resizebox{1.0\linewidth}{!}{
    \begin{tabular}{l|l|ll}
    \toprule
        \multirow{2}{*}{Method} & \multirow{2}{*}{SLM} & \multicolumn{2}{c}{Avg.} \\
        \cmidrule{3-4}
        ~ & ~ & Quality. $\uparrow$ & $\text{Priv}_{\text{sample}}$ $\downarrow$ \\
        \midrule
        CoT & Qwen3-235B & 95.21 & 100.00 \\
        \midrule
        \multirow{2}{*}{\makecell[l]{AutoMix}} & Qwen3-4B & 82.07 $_{(-4.9)}$ & 27.11 $_{(-26.0)}$ \\
        ~ & Qwen3-8B & 86.96 $_{(-5.2)}$ & 22.17 $_{(-21.2)}$ \\
        \midrule
        \multirow{2}{*}{\makecell[l]{Self-Disclosures}} & Qwen3-4B & 80.77 $_{(-6.2)}$ & 23.20 $_{(-22.1)}$ \\
        ~ & Qwen3-8B & 85.24 $_{(-6.9)}$ & 18.98 $_{(-18.0)}$ \\
        \midrule
        \multirow{2}{*}{\makecell[l]{PAPILLON}} & Qwen3-4B & 83.87 $_{(-3.1)}$ & 8.55 $_{(-7.5)}$ \\
        ~ & Qwen3-8B & \underline{88.11} $_{(-4.0)}$ & 6.68 $_{(-5.7)}$ \\
        \midrule
        \multirow{2}{*}{\makecell[l]{Router-R1}} & Qwen3-4B & 82.76 $_{(-4.2)}$ & 7.29 $_{(-6.2)}$ \\
        ~ & Qwen3-8B & 88.00 $_{(-4.1)}$ & 6.36 $_{(-5.4)}$ \\
        \midrule
        \multirow{2}{*}{Ours} & Qwen3-4B & 86.94 $_{(+0.0)}$ & \underline{1.10} $_{(+0.0)}$ \\
        ~ & Qwen3-8B & \textbf{92.11} $_{(+0.0)}$ & \textbf{1.01} $_{(+0.0)}$ \\
        \bottomrule
    \end{tabular}
 }
\vspace{-0.4em}
\end{wraptable}

\textbf{Our framework effectively transfers to real-world personal tasks without additional training.} 
Table~\ref{tab:real_world} reports results on three real-world benchmarks, PUPA~\citep{papillon}, AESLC~\citep{aeslc}, and SAMSum~\citep{samsum}. On these unseen tasks, our framework simultaneously achieves the strongest privacy preservation and the highest response quality: Qwen3-8B attains an average $\text{Priv}_{\text{sample}}$ of 1.01\% and Qwen3-4B 1.10\%, reducing leakage over baselines by 5.35\% -- 21.16\% and 6.19\% -- 26.01\% respectively, while improving quality by 4.0\% -- 6.9\% and 3.1\% -- 6.2\%. The Qwen3-8B score 92.1\% also closes most of the gap to the LLM-only upper bound which is 95.2\% from Qwen3-235B CoT and leaks every sample. These results confirm that the collaboration policy learned on PrivQA generalizes well to unseen real-world datasets and user scenarios.

\subsection{Impact of Different SLMs}

\textbf{RQ2.} How does the capability of the SLM affect the collaboration strategies?

\textbf{Stronger SLM achieves better collaboration performance, while smaller SLM exhibits greater improvement.}
Table \ref{tab:main_result} reports the performance of different SLMs under dynamic collaboration 
\begin{wraptable}{r}{6.5cm}
\centering
\caption{Average performance of different SLMs with Qwen3-235B and baselines.}\label{tab:main_result}
\resizebox{1.0\linewidth}{!}{
    \begin{tabular}{l|l|ll}
    \toprule
        \multirow{2}{*}{Method} & \multirow{2}{*}{SLM} & \multicolumn{2}{c}{Avg.} \\
        \cmidrule{3-4}
        ~ & ~ & EM $\uparrow$ & $F_{\text{BERT}}$ $\uparrow$ \\
        \midrule
        Direct Infer & Qwen3-235B & 36.96 & 70.54 \\
        \midrule
        CoT & Qwen3-235B & 42.50 & 74.54 \\
        \midrule
        \multirow{4}{*}{CoT} & Llama-3.2-3B & 17.74 & 56.49 \\
        ~ & Llama-3.1-8B & 24.38 & 60.86 \\
        ~ & Qwen3-4B & 20.96 & 62.58 \\
        ~ & Qwen3-8B & 25.77 & 64.73 \\
        \midrule
        \multirow{4}{*}{Ours} & Llama-3.2-3B & 35.86$_{(+18.1)}$ & 71.42$_{(+14.9)}$ \\
        ~ & Llama-3.1-8B & \underline{38.92}$_{(+14.5)}$ & 73.37$_{(+12.5)}$ \\
        ~ & Qwen3-4B & 38.34$_{(+17.4)}$ & \underline{73.51}$_{(+10.9)}$ \\
        ~ & Qwen3-8B & \textbf{43.11}$_{(+17.3)}$ & \textbf{75.55}$_{(+10.8)}$ \\
    \bottomrule
    \end{tabular}
}
\vspace{-0.6em}
\end{wraptable}
with Qwen3-235B. Across every SLM, our framework substantially improves over standalone CoT inference, raising EM by 14.5\% -- 18.1\% and $F_{\text{BERT}}$ by 10.8\% -- 14.9\%, which confirms that SLMs at different scales and from different families can all learn effective coordination with the LLM. Two complementary trends further emerge. First, stronger SLMs attain better absolute collaboration performance: Qwen3-8B reaches 43.1\% EM, 4.8\% higher than Qwen3-4B and even 0.6\% above the LLM-with-CoT upper bound, while Llama-3.1-8B exceeds Llama-3.2-3B by 3.1\%. Second, within each family the relative gain is larger for the smaller model: Llama-3.2-3B improves EM by 18.1\% over its own CoT versus 14.5\% for Llama-3.1-8B. The intra-family EM gap between the 3B and 8B Llama models shrinks from 6.64\% at CoT down to 3.06\% under our framework. These two trends indicate that LLM feedback compensates most heavily for the capability deficit of weaker SLMs and narrows the scale gap across a family, while stronger SLMs still extract additional value from the same policy and can even surpass direct LLM inference.

\begin{figure}[!t]
  \centering
  \begin{subfigure}{0.49\linewidth}
    \centering
    \includegraphics[width=\textwidth]{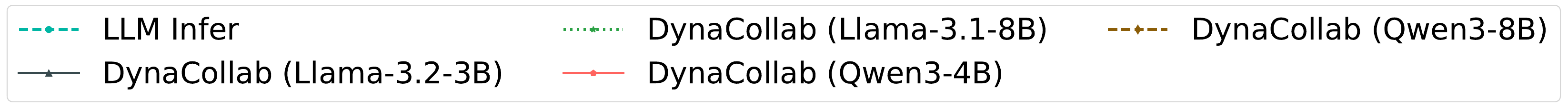}
  \end{subfigure}
  \hfill
  \begin{subfigure}{0.49\linewidth}
    \centering
    \includegraphics[width=\textwidth]{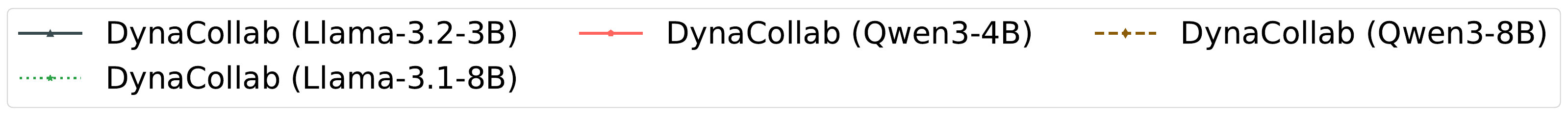}
  \end{subfigure}
  \begin{subfigure}{0.235\linewidth}
    \centering
    \includegraphics[width=\textwidth]{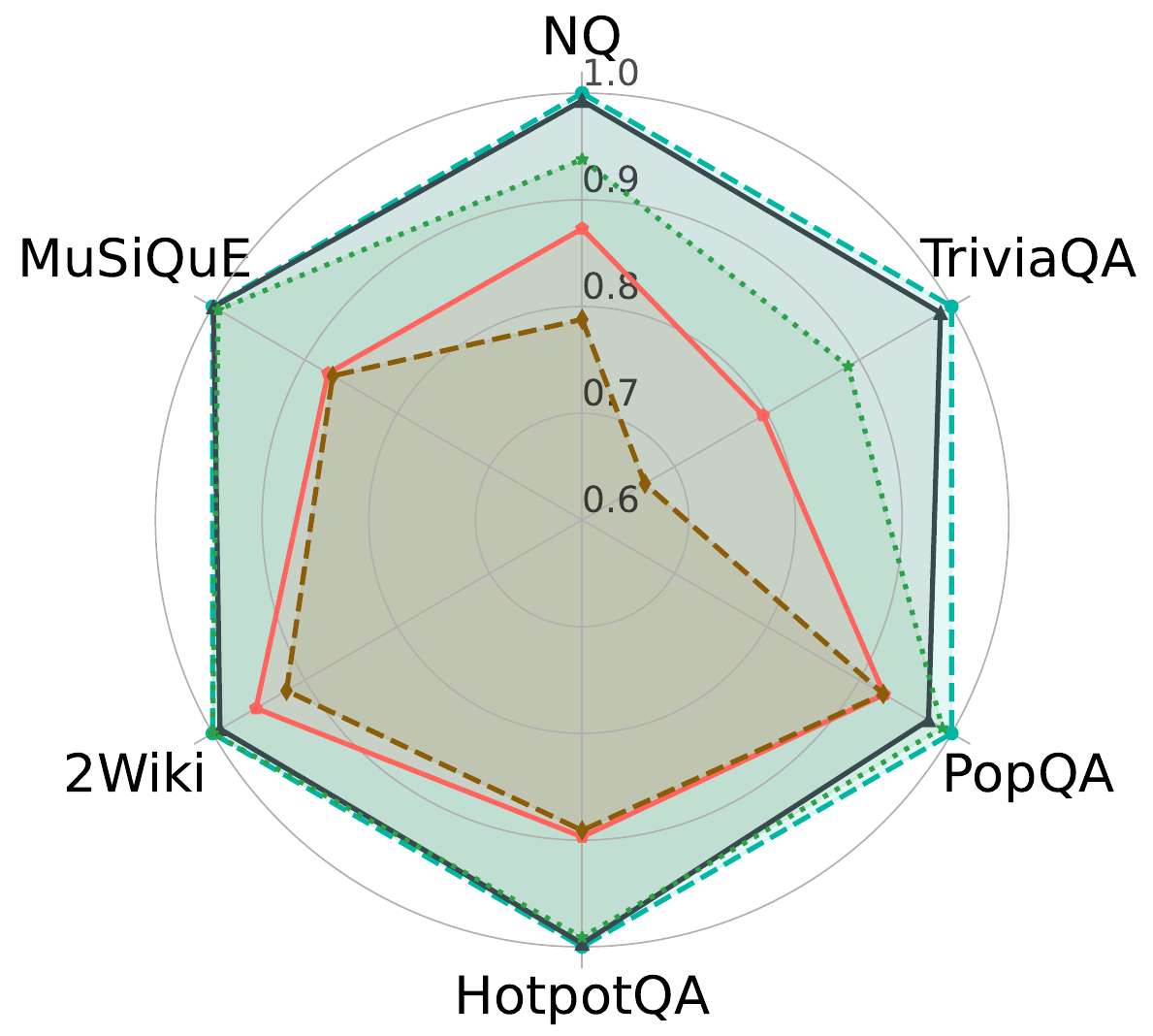}
    \vspace{-1em}
    \caption{Avg Interaction Turn $\downarrow$}
    \label{fig:avg_turn}
  \end{subfigure}
  \hfill
  \begin{subfigure}{0.235\linewidth}
    \centering
    \includegraphics[width=\textwidth]{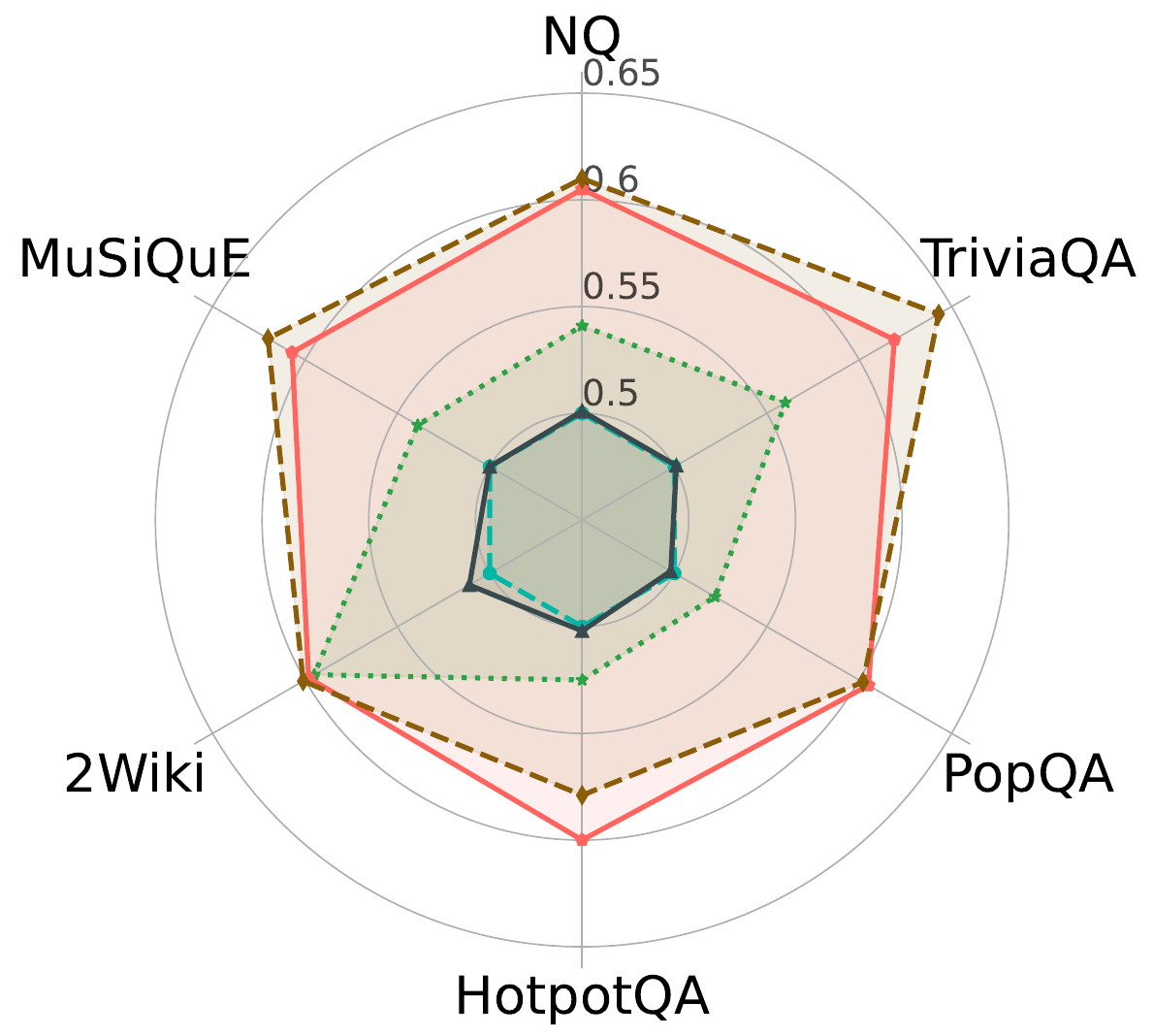}
    \vspace{-1em}
    \caption{INScore $\uparrow$}
    \label{fig:req_necessary_score}
  \end{subfigure}
  \hfill
  \begin{subfigure}{0.235\linewidth}
    \centering
    \includegraphics[width=\textwidth]{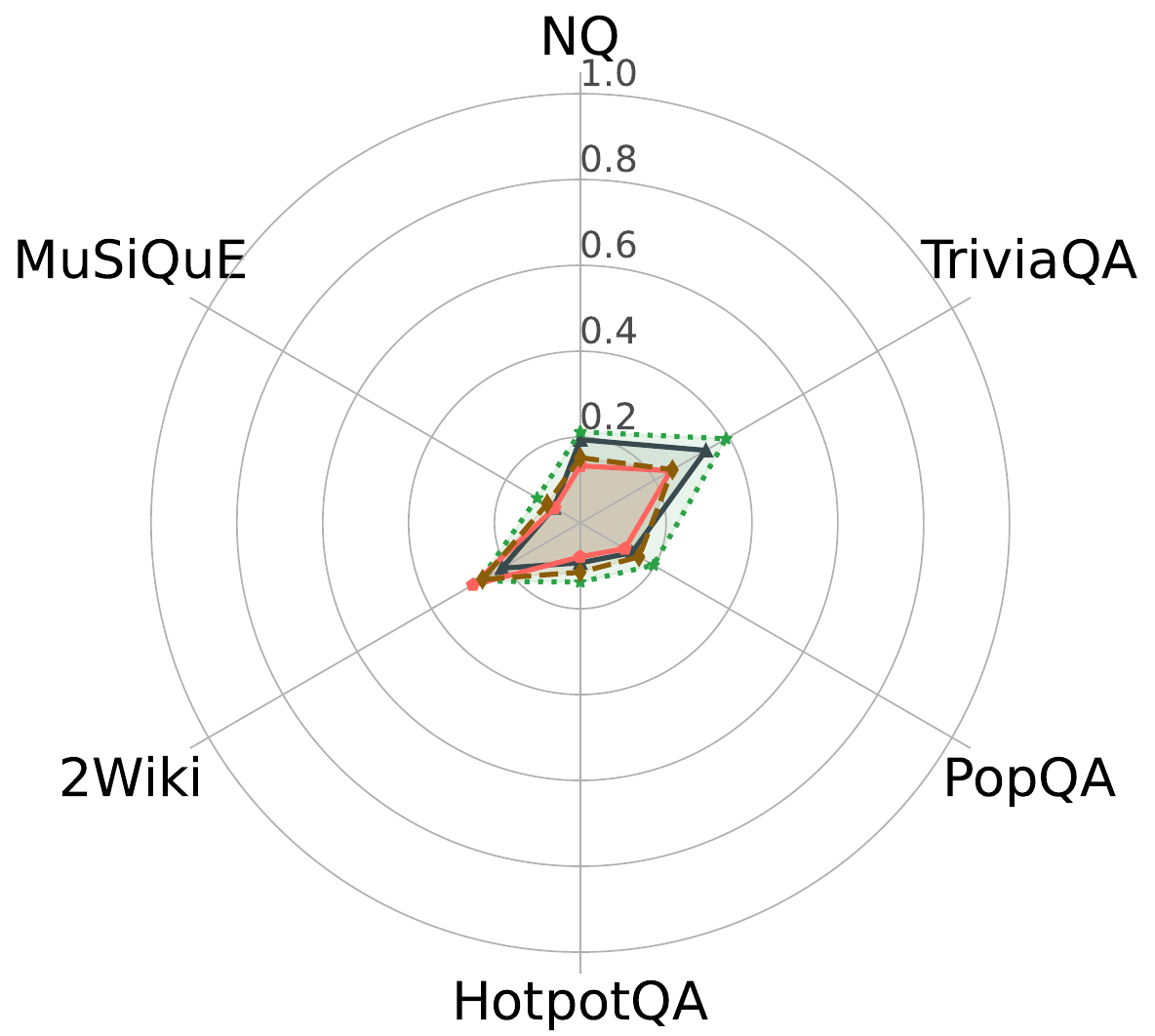}
    \vspace{-1em}
    \caption{Request Rate (Easy)} 
    \label{fig:avg_turn_answerable}
  \end{subfigure}
  \hfill
  \begin{subfigure}{0.235\linewidth}
    \centering
    \includegraphics[width=\textwidth]{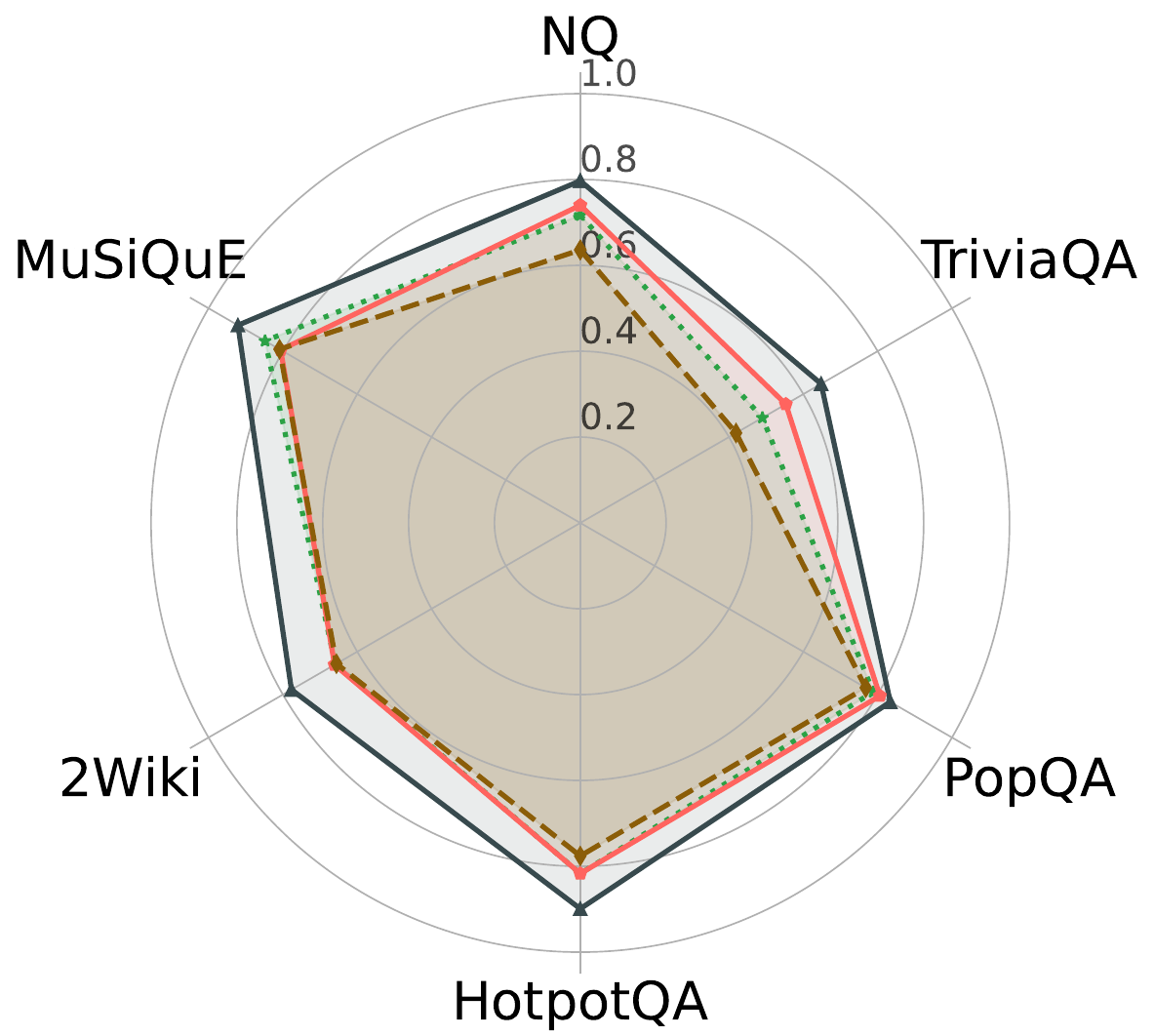}
    \vspace{-1em}
    \caption{Request Rate (Hard)} 
    \label{fig:avg_turn_not_answerable}
  \end{subfigure}
  \vspace{-0.2em}
  \caption{Efficiency analysis of collaboration strategy learned by different SLMs with same LLM.}
  \label{fig:efficiency_result} 
\vspace{-1.2em}
\end{figure}

\textbf{SLMs with fewer parameters rely more heavily on querying LLM to improve their final performance.}
We further evaluate the average interaction turns and INScore to assess how SLMs balance efficiency and performance, as shown in Figure \ref{fig:efficiency_result}. Compared with directly forwarding user queries to the LLM, the dynamic collaboration framework reduces average interaction turns by 0.11 and 0.15, while improving INScore by 9.6\% and 10.9\% on Qwen3-4B and Qwen3-8B, respectively. Across model scales, smaller SLMs rely more on the LLM: compared with Qwen3-8B, Llama-3.2-3B and Qwen3-4B increase average interaction turns by 0.14 and 0.04, respectively. Figure \ref{fig:avg_turn} also shows that Llama-based SLMs are more inclined to request LLM assistance to ensure correctness. Moreover, Figures \ref{fig:avg_turn_answerable} and \ref{fig:avg_turn_not_answerable} indicate that request rates are much lower on easy samples than on hard ones. For Qwen3-8B, the rate drops by 50.7\% on easy samples. These findings demonstrate that SLMs can adaptively decide when to request based on task difficulty and their own capabilities.

\begin{wrapfigure}{r}{7.4cm}
\vspace{-1.2em}
  \centering
         \centering
         \begin{subfigure}{0.48\linewidth}
             \centering
            \begin{minipage}{\linewidth}
             \includegraphics[width=\textwidth]{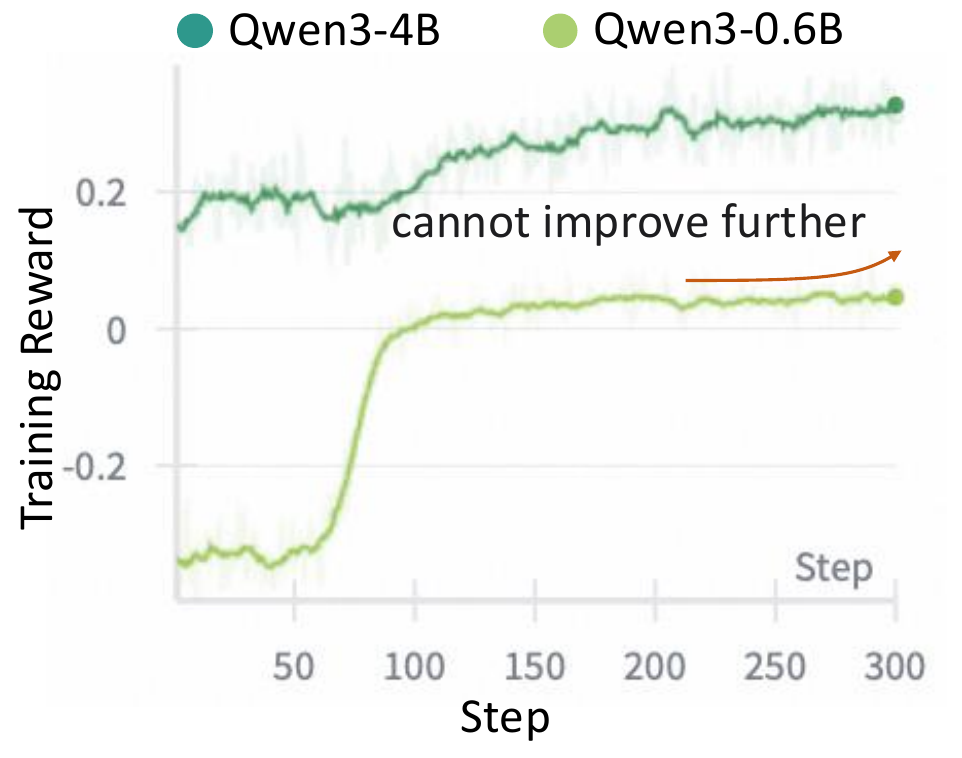}
            \end{minipage}
            \vspace{-0.8em}
             \caption{Final Training Reward $R$}
             \label{fig:slm_reward}
        \end{subfigure}
        \hfill
        \begin{subfigure}{0.48\linewidth}
             \centering
            \begin{minipage}{\linewidth}
             \includegraphics[width=\textwidth]{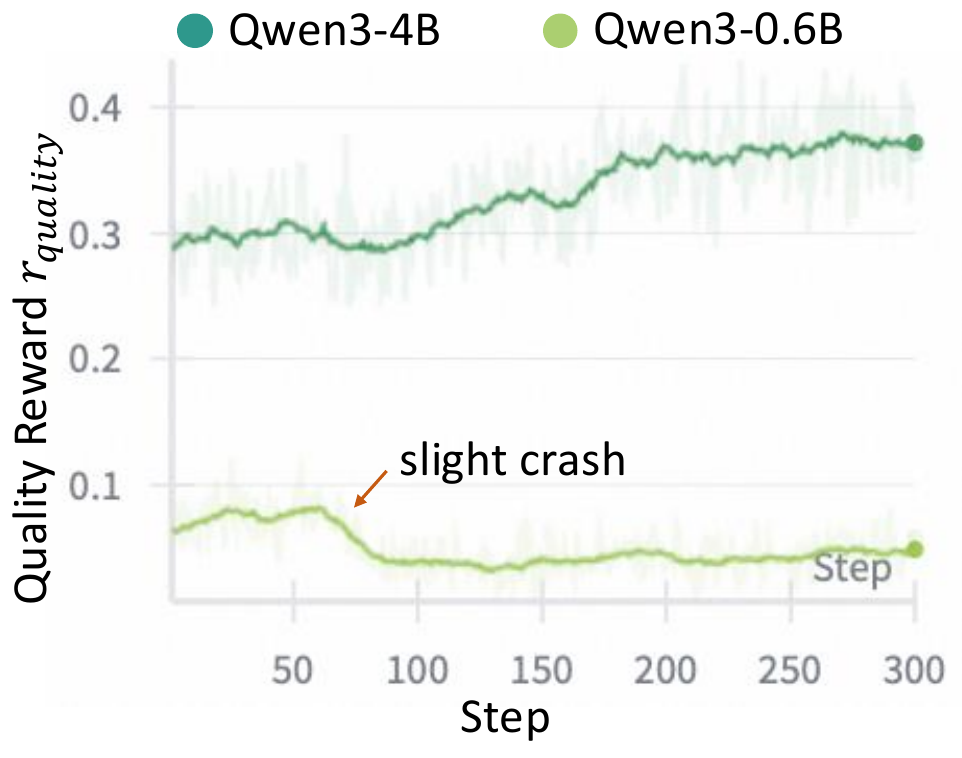}
            \end{minipage}
            \vspace{-0.8em}
             \caption{Quality Reward $r_{\text{quality}}$}
             \label{fig:slm_score}
        \end{subfigure}
        \vspace{-0.6em}
    \caption{Training reward and quality reward of 0.6B and 4B models collaborating with Qwen3-235B.}\label{fig:slm_train}
    \vspace{-0.8em}
\end{wrapfigure}

\textbf{SLMs with weak instruction-following capabilities fail to receive effective feedback under end-to-end RL.}
To assess whether all SLMs can learn collaboration via end-to-end RL, we further evaluate Qwen3-0.6B. In Figures~\ref{fig:slm_reward} and~\ref{fig:slm_score}, Qwen3-0.6B converges to a low final reward with a mild collapse in $r_{\text{quality}}$, while Qwen3-4B achieves a stable, monotonic reward increase. This gap stems from Qwen3-0.6B's weak instruction-following ability: it rarely produces well-formed LLM requests during exploration, eliciting unhelpful responses and limited positive rewards. Such SLMs therefore require a supervised cold-start to teach basic request formatting before RL. 

\subsection{Impact of Different LLMs}

\textbf{RQ3.} How does LLM capability affect the effectiveness and behavior of collaboration?

\begin{wraptable}{r}{6.4cm}
\centering
\vspace{-1.2em}
\caption{Average performance of SLM with different LLMs.}\label{tab:diff_llm_result}
    \vspace{-0.3em}
\resizebox{0.98\linewidth}{!}{
    \begin{tabular}{l|l|l|ll}
    \toprule
         \multirow{2}{*}{Method} & \multirow{2}{*}{SLM} & \multirow{2}{*}{LLM} & \multicolumn{2}{c}{Avg.} \\
         \cmidrule{4-5}
        ~ & ~ & ~ & EM $\uparrow$ & $F_{\text{BERT}}$ $\uparrow$ \\
        \midrule
        \multirow{3}{*}{CoT} & \multirow{3}{*}{no SLM} & DeepSeek-70B & 39.95 & 72.94 \\
        ~ & ~ & Qwen3-235B & 42.50 & 74.54 \\
        ~ & ~ & Qwen3-Max & 47.08 & 76.89 \\
        \midrule
        \multirow{6}{*}{Ours} & Qwen3-4B & \multirow{2}{*}{DeepSeek-70B} & 36.57 & 72.56 \\
        ~ & Qwen3-8B & ~ & 37.80 & 73.31 \\
        \cmidrule{2-5}
        ~ & Qwen3-4B & \multirow{2}{*}{Qwen3-235B} & 38.34 & 73.51 \\
        ~ & Qwen3-8B & ~ & \underline{43.11} & \underline{75.55} \\
        \cmidrule{2-5}
        ~ & Qwen3-4B & \multirow{2}{*}{Qwen3-Max$^*$} & 41.68 & 75.13 \\
        ~ & Qwen3-8B & ~ & \textbf{46.01} & \textbf{77.02} \\
        \bottomrule
    \end{tabular}
}
\vspace{-1em}
\end{wraptable}

\textbf{Collaborating with stronger LLMs yields better performance.}
Table \ref{tab:diff_llm_result} reports the performance of identical SLMs when paired with different LLMs. Across different LLMs, SLMs are able to learn appropriate collaboration strategies and achieve performance comparable to that of their corresponding LLM partners. Moreover, we observe a positive correlation between the SLMs’ final response quality and the capability of the LLMs: stronger LLMs provide more informative and relevant interaction signals, which in turn lead to more accurate answers. For Qwen3-4B and Qwen3-8B, collaborating with Qwen3-235B yields average EM improvements of 1.8\% and 5.3\%, respectively, across datasets, compared to DeepSeek-Distill-Llama-70B.

\begin{wrapfigure}{r}{7.5cm}
\vspace{-1.2em}
\centering
         \centering
        \begin{subfigure}{0.98\linewidth}
             \centering
            \begin{minipage}{\linewidth}
             \includegraphics[width=\textwidth]{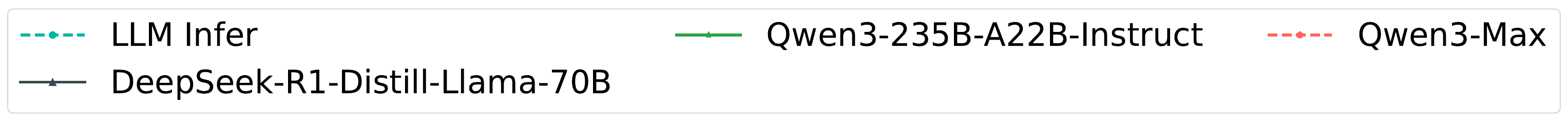}
            \end{minipage}
        \end{subfigure}
        \hfill
         \begin{subfigure}{0.44\linewidth}
             \centering
            \begin{minipage}{\linewidth}
             \includegraphics[width=\textwidth]{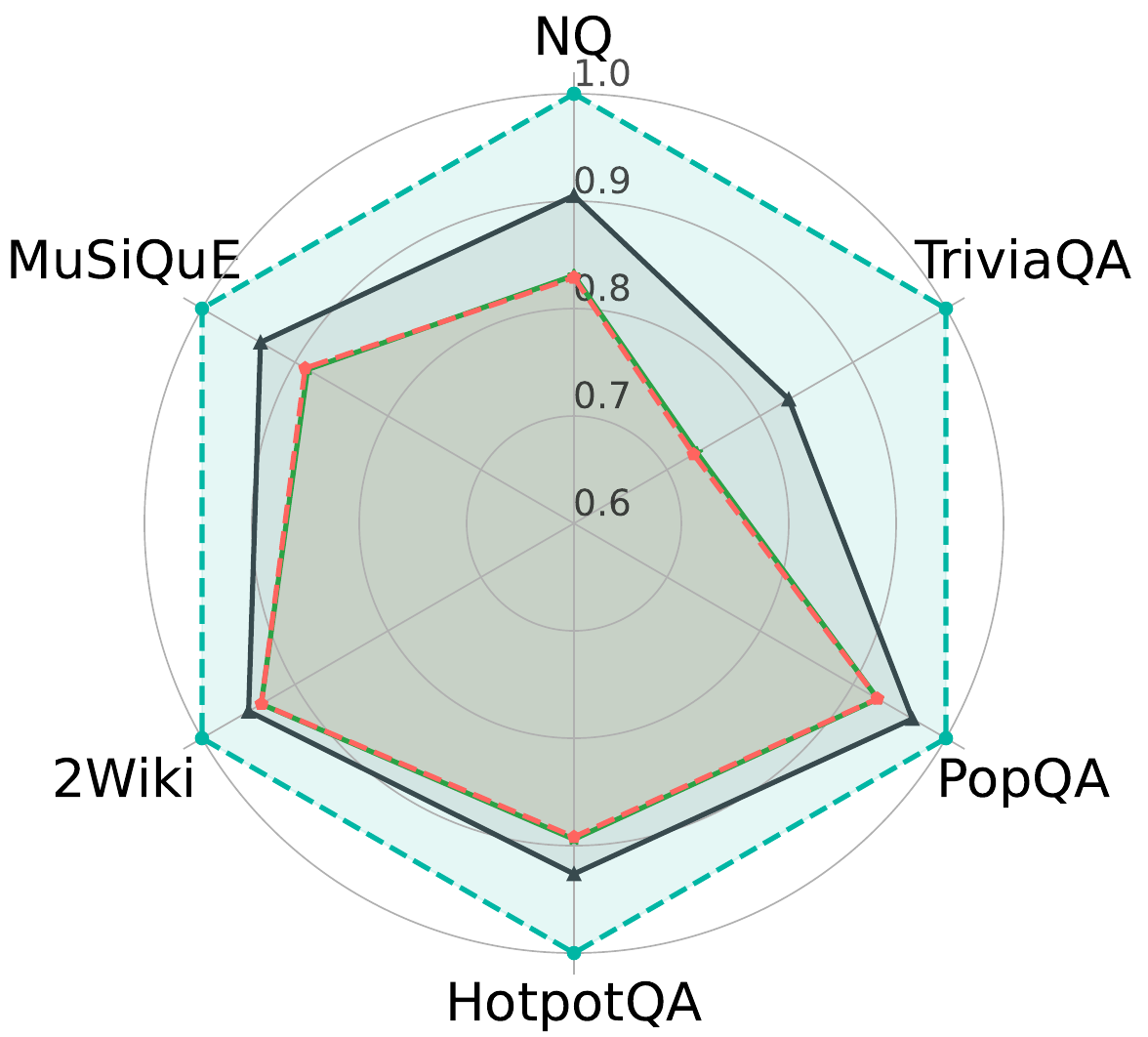}
            \end{minipage}
             \caption{Interaction Turn $\downarrow$}
             \label{fig:avg_turn_diff_llm}
        \end{subfigure}
        \begin{subfigure}{0.44\linewidth}
             \centering
            \begin{minipage}{\linewidth}
             \includegraphics[width=\textwidth]{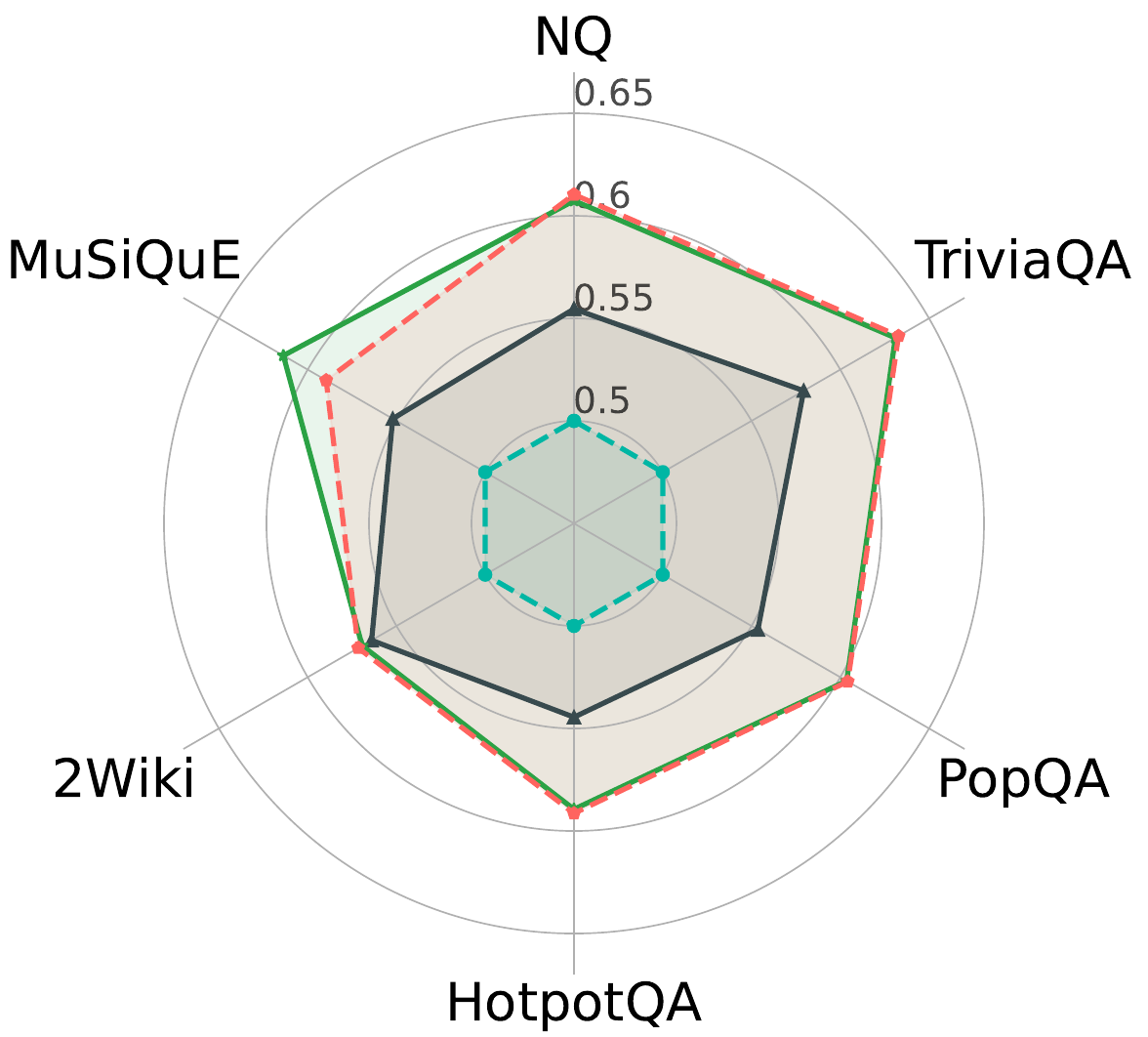}
            \end{minipage}
             \caption{INScore $\uparrow$}
             \label{fig:interation_score_diff_llm}
        \end{subfigure}
        \vspace{-0.4em}
    \caption{Average interaction turn and interaction necessity score across SLMs with different LLMs.} \label{fig:effec_diff_llm}
    \vspace{-0.6em}
\end{wrapfigure}

\textbf{Fewer interaction turns are needed when collaborating with stronger LLMs.}
To analyze how SLMs adapt their collaboration strategies to different LLMs, we compare the average interaction turns and INScore across different LLM partners in Figure \ref{fig:avg_turn_diff_llm} and Figure \ref{fig:interation_score_diff_llm}. For the same SLM, collaboration with a stronger LLM generally requires fewer interaction turns. For example, pairing Qwen3-4B with Qwen3-235B reduces the average turns by 0.10 compared with DeepSeek-Distill-Llama-70B. The key observation is that when paired with the weaker LLM, the SLM tends to decompose the problem into multiple sub-requests, whereas with a stronger LLM, it often issues a single consolidated request.

\textbf{Collaboration strategies learned with a smaller LLM transfer effectively to larger unseen LLMs.}
SLMs trained via collaboration with Qwen3-235B-A22B-Instruct directly generalize to Qwen3-Max, an unseen closed-source LLM, and yield further improvements. Specifically, using Qwen3-Max improves average EM by 3.3\% for Qwen3-4B and 2.9\% for Qwen3-8B over Qwen3-235B-A22B-Instruct, while the average number of interaction turns remains largely unchanged. This suggests that we can train collaboration behaviors with a smaller LLM and deploy with a stronger one at inference time, reducing training cost while further improving performance.

\subsection{Ablation Study and Hyper-Parameter Analysis}

\textbf{RQ4.} How do the external constraints shape the learned collaboration strategies?

\begin{figure}[!t]
\centering

\begin{minipage}[t]{0.49\textwidth}
\centering

    \begin{subfigure}[t]{0.49\linewidth}
        \centering
        \includegraphics[width=\linewidth]{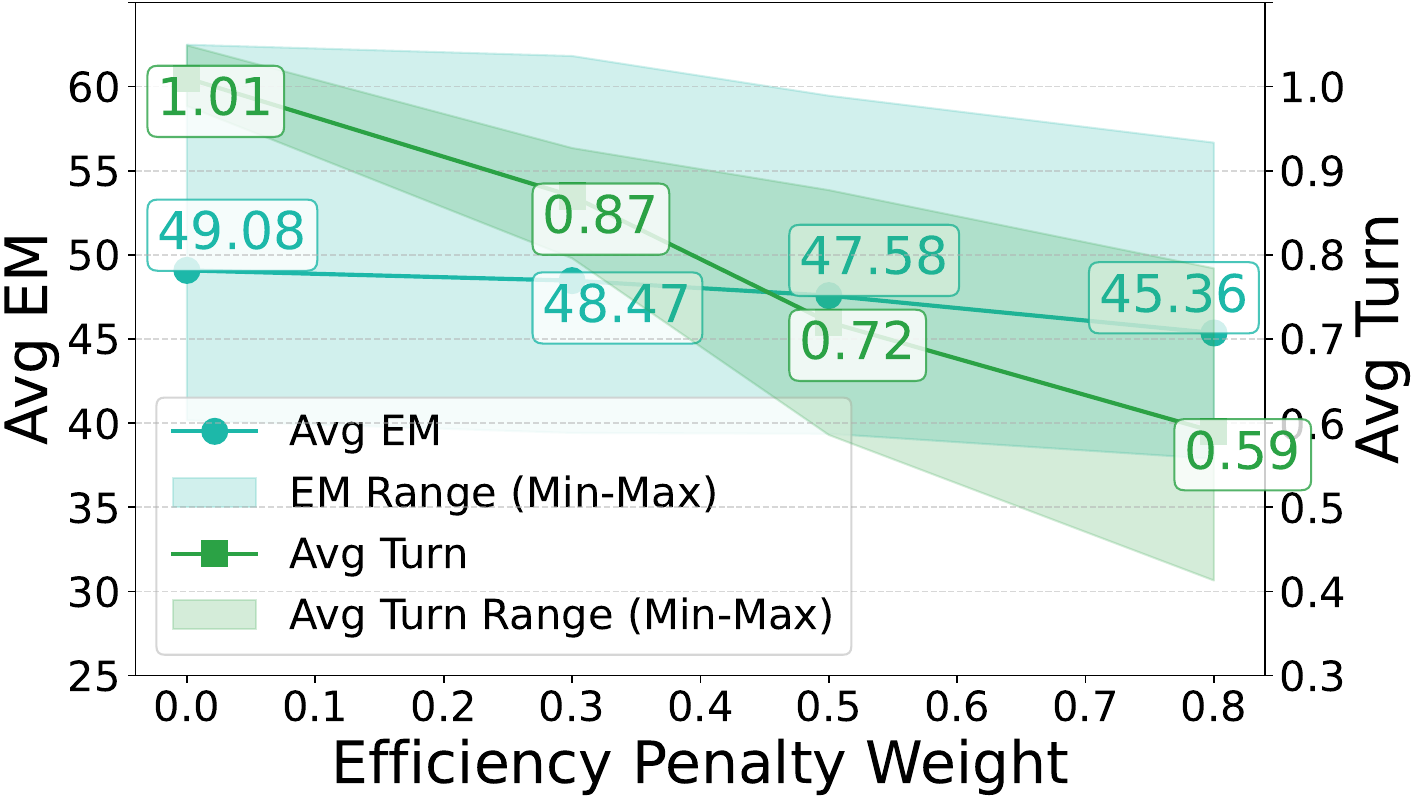}
        \caption{General QA}
        \label{fig:efficiency_general_qa}
    \end{subfigure}
    \hfill
    \begin{subfigure}[t]{0.49\linewidth}
        \centering
        \includegraphics[width=\linewidth]{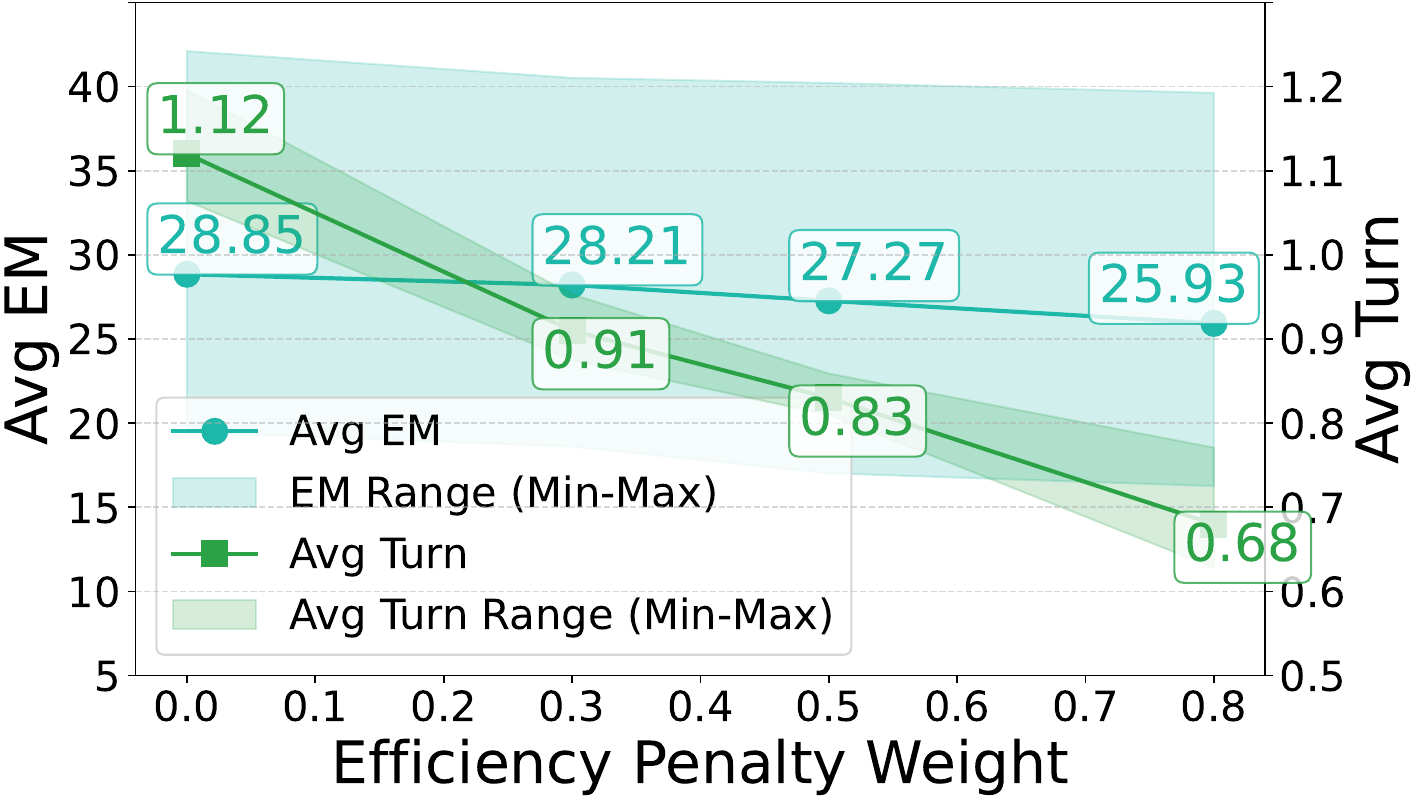}
        \caption{Multi-hop QA}
        \label{fig:efficiency_multi_qa}
    \end{subfigure}

    \vspace{-0.4em}
    \caption{Pareto frontier of interaction efficiency and EM under varying efficiency weight.}
    \label{fig:efficiency_ablation}

\end{minipage}
\hfill
\begin{minipage}[t]{0.49\textwidth}
\centering

    \begin{subfigure}[t]{0.49\linewidth}
        \centering
        \includegraphics[width=\linewidth]{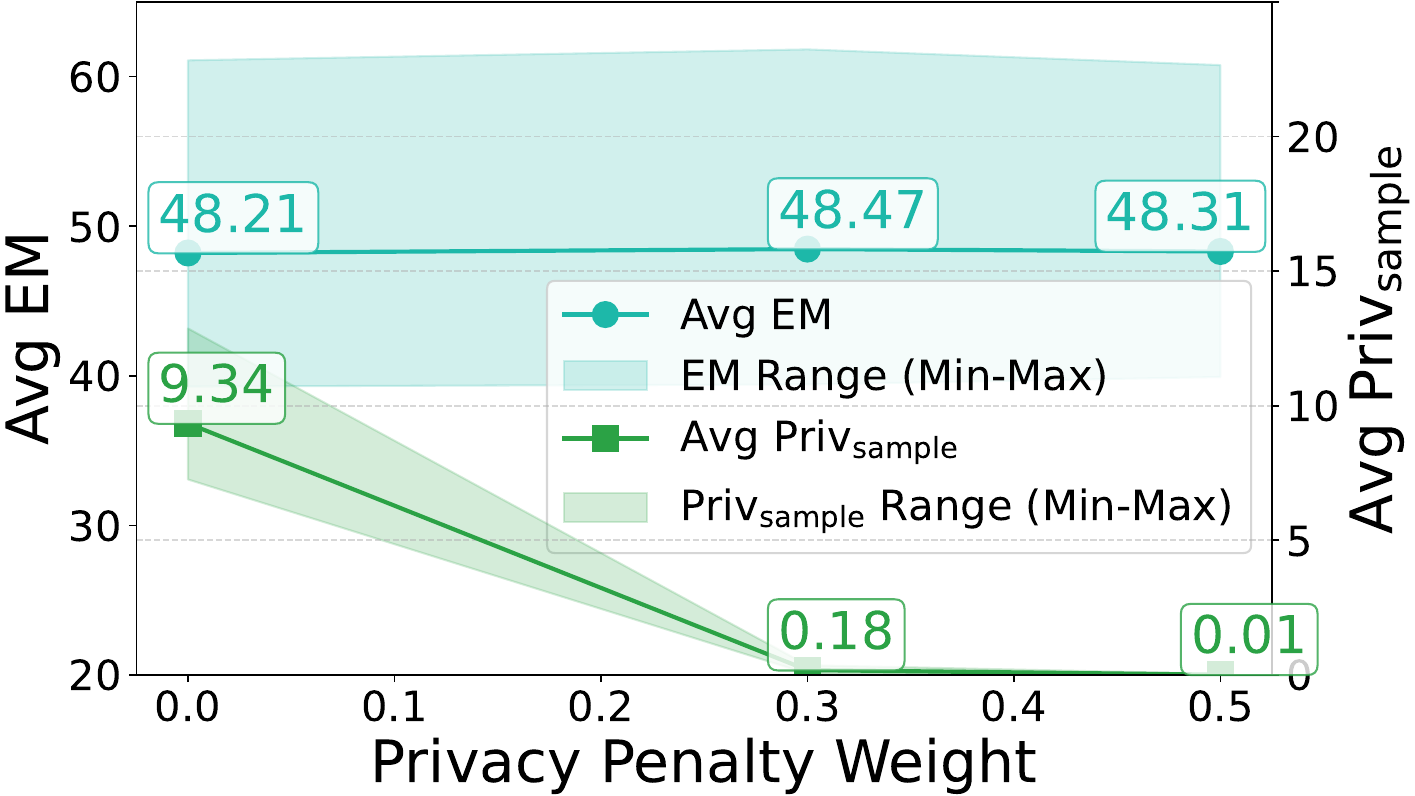}
        \caption{General QA}
        \label{fig:privacy_general_qa}
    \end{subfigure}
    \hfill
    \begin{subfigure}[t]{0.49\linewidth}
        \centering
        \includegraphics[width=\linewidth]{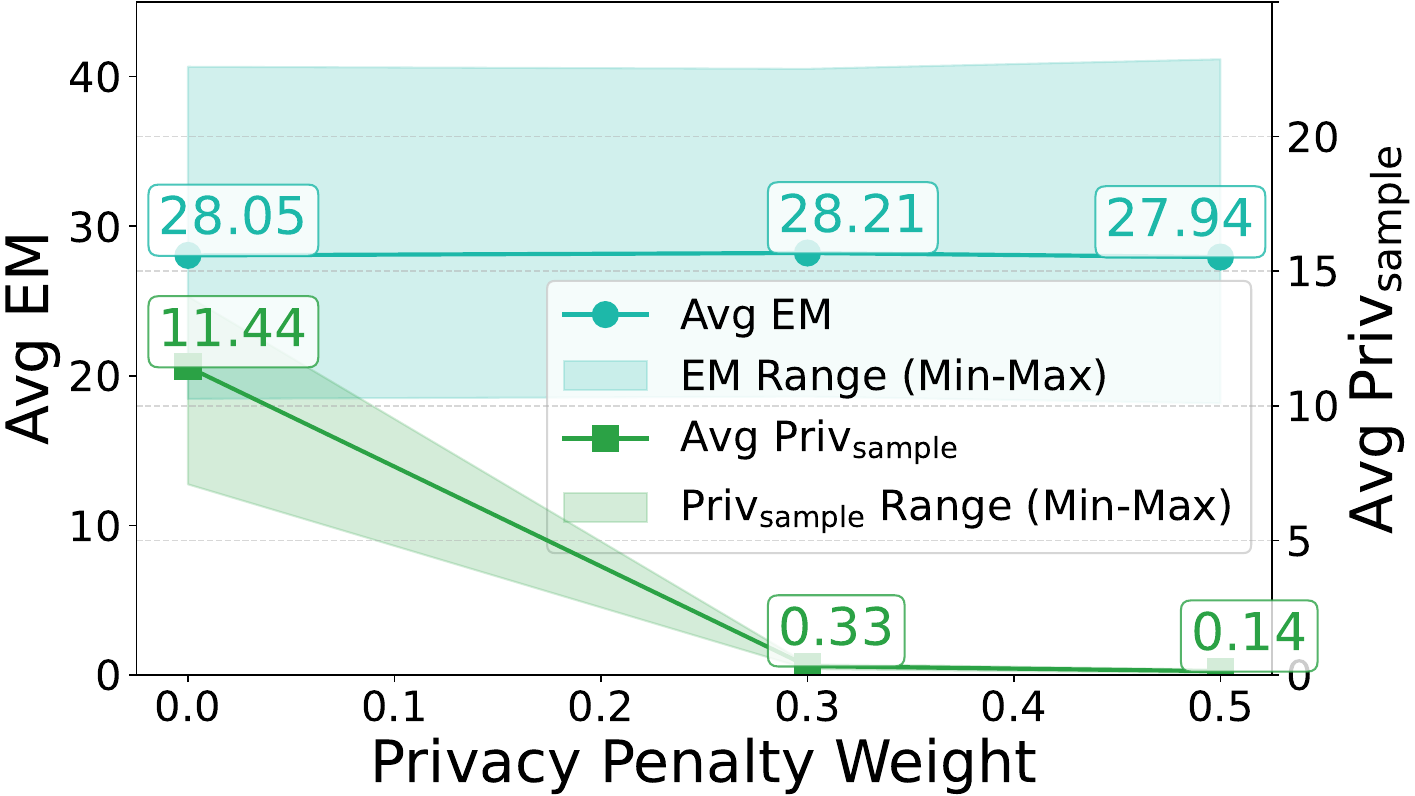}
        \caption{Multi-hop QA}
        \label{fig:privacy_multi_qa}
    \end{subfigure}

    \vspace{-0.4em}
    \caption{Pareto frontier of $\text{Priv}_{\text{sample}}$ and EM under varying privacy weight}
    \label{fig:privacy_ablation}

\end{minipage}

\vspace{-1.2em}
\end{figure}

\textbf{Impact of LLM Feedback.}
We isolate the contribution of active LLM feedback by an ablation that replaces the feedback LLM with a passive answer-only tool while keeping the reward identical. As
\begin{wraptable}{r}{5.6cm}
\centering
\caption{Average performance of Qwen3-4B interacting with a feedback LLM vs.\ a tool LLM.}\label{tab:llm_feedback_ablation}
    \vspace{-0.3em}
\resizebox{0.98\linewidth}{!}{
    \begin{tabular}{l|l|ll}
    \toprule
        \multirow{2}{*}{Method} & \multirow{2}{*}{Model} & \multicolumn{2}{c}{Avg.} \\
        \cmidrule{3-4}
        ~ & ~ & EM $\uparrow$ & $F_{\text{BERT}}$ $\uparrow$ \\
        \midrule
        CoT & Qwen3-235B & 42.50 & 74.54 \\
        \midrule
        \makecell[l]{Ours w/o feedback} & Qwen3-4B & 35.53 & 71.53 \\
        \midrule
        Ours & Qwen3-4B & \textbf{38.34} & \textbf{73.51} \\
        \bottomrule
    \end{tabular}
}
\vspace{-0.6em}
\end{wraptable}
shown in Table \ref{tab:llm_feedback_ablation}, removing feedback drops average EM by 2.8\% and $F_{\text{BERT}}$ by 2.0\% on Qwen3-4B. This gap indicates that the clarify action supplies a process-level supervision signal on request quality that outcome reward alone cannot substitute, localizing errors to individual under-specified requests rather than attributing them to an entire multi-turn trajectory. Per-dataset results are in Appendix Table \ref{tab:llm_feedback_ablation_full}. Appendix \ref{sec:freq_llm_action} further reports the frequency of the LLM's two feedback actions. 

\textbf{Impact of Efficiency Penalty.} Figure \ref{fig:efficiency_ablation} shows how the efficiency penalty weight controls the trade-off between interaction and quality. As the weight increases, the SLM relies more on local reasoning and requests the LLM less frequently. For instance, setting the weight to 0.8 reduces the average interaction turns between Qwen3-4B and Qwen3-235B by 0.43 at the cost of a 3.3\% drop in EM. This clear trade-off between efficiency and answer quality can be tuned via the efficiency penalty at training time, allowing the SLM to adapt to different cost-sensitive deployment scenarios.

\textbf{Impact of Privacy Penalty.} Figure \ref{fig:privacy_ablation} shows the effect of privacy penalties on collaboration strategies. Without privacy constraints, the SLM partially leaks user information because it prioritizes information retrieval over privacy reformulation. Setting the privacy penalty weight above 0.3 produces strong privacy protection: relative to no penalty, $\text{Priv}_{\text{sample}}$ decreases by 10.1\% with only a negligible drop in EM. This gain arises because the SLM learns to reformulate sensitive queries into objective, user-agnostic requests, preserving privacy while still retrieving the necessary information.

\section{Conclusion} \label{sec:conclusion}
In this work, we have presented a dynamic collaboration framework for local-first personal intelligence, which transforms the interaction between on-device SLMs and cloud-based LLMs from static, predefined pipelines into a proactive, adaptive partnership. We have demonstrated that SLMs can autonomously acquire collaboration strategies that balance the comprehensive objectives of response quality, interaction efficiency, and privacy preservation, and that these strategies transfer to real-world writing and summarization tasks. By enabling heterogeneous models to communicate through informative feedback rather than simple query-response loops, we move closer to a future where local and cloud intelligence operate as a unified ecosystem.

\subsection*{AI use statement}
We used generative AI tools for two research subtasks: (i) constructing the PrivQA dataset (Section~\ref{sec:collab_strategy_learning}; Appendix~\ref{sec:dataset_statistics}), where a synthesis LLM injects privacy-related indirect references into base QA pairs; and (ii) automated evaluation via LLM-as-a-judge \citep{llm_as_a_judge, papillon}, applied to privacy leakage on PrivQA and to response quality on the real-world benchmarks (Appendix~\ref{sec:eval_metric_def}). The reliability of both uses is validated by human evaluation in Appendix~\ref{sec:human_eval}. Beyond these subtasks, generative AI tools were not used for formulating hypotheses, developing theoretical models, designing experiments, or interpreting results. We take full responsibility for all content of this work.

\subsection*{Ethics statement}
Our work targets a privacy-preserving personal-intelligence setting and intersects with the Code of Ethics on privacy, dataset release, and human-subject studies, which we address as follows. First, PrivQA is constructed by injecting \emph{fictional} privacy items into publicly released QA datasets; we introduce no real user identifiers, and the real-world benchmarks we evaluate on \citep{papillon, aeslc, samsum} are also publicly available. Second, the human evaluations reported in Appendix~\ref{sec:human_eval} involve only volunteer annotators from our lab reviewing model-generated text, and contain no personally identifying information. We have reviewed the Code of Ethics and adhered to it throughout the work.

\subsection*{Reproducibility statement}
To ensure reproducibility, we release the following alongside the paper. Our full training and inference stack is built on the verl framework \citep{verl} using the VAPO algorithm \citep{vapo}; all reward weights, hyperparameters, and hardware are documented in Appendix~\ref{sec:implementation_details}, with training and inference compute reported in Appendices~\ref{sec:computational_overhead} and~\ref{sec:cost_energy}. The PrivQA construction pipeline, dataset statistics, and synthesis cost are detailed in Appendix~\ref{sec:dataset_statistics} and Appendix~\ref{sec:llm_usage}, and every prompt used in training, dataset synthesis, and evaluation is provided verbatim in Appendix~\ref{sec:prompt_framework}. The convergence result under evaluator noise is proved in Appendix~\ref{sec:noise_robust}, with all assumptions made explicit. All base models used in our experiments are publicly available.

\bibliography{main}
\bibliographystyle{main}

\newpage
\appendix
\clearpage
\section*{Appendix Contents}
\addcontentsline{toc}{section}{Appendix Contents}
\startcontents[appendix]
\printcontents[appendix]{l}{1}{\setcounter{tocdepth}{2}}
\clearpage

\section{Limitations and Social Impact} \label{sec:limitation_and_social_impact}

\subsection{Limitations}
The RL-based strategy learning in our framework assumes basic instruction-following ability. Thus, extremely lightweight models may need a supervised cold-start to stabilize training and leverage LLM feedback effectively. In addition, although we formalize privacy leakage as an item-set accounting problem in Appendix~\ref{sec:eval_metric_def} and empirically reduce privacy leakage to near zero on both PrivQA and real-world benchmarks, our framework only guarantees a substantial reduction rather than the complete elimination of privacy leakage.

\subsection{Broader Social Impact}
This work targets a socially beneficial direction: enabling capable personal AI while reducing the amount of user data disclosed to cloud services. By keeping user-attributed information on-device and forwarding only user-agnostic information queries, our framework can lower aggregate data exposure in commercial personal-intelligence deployments, e.g., on-device assistants that route through Private Cloud Compute or similar backends. It also provides infrastructure for privacy-aware research: we will release PrivQA and code with privacy items, injected into publicly available QA corpora, and no real user data is collected or released. As with any privacy technology, deployments should combine our framework with system-level access controls and standard LLM safety measures.

\section{Comparison with Existing Work}\label{sec:existing_framework}

\begin{table*}[!ht]
  \caption{Comparative overview of existing work and our dynamic collaboration framework.}
  \label{tab:existing_framework}
  \begin{center}
      \resizebox{1\linewidth}{!}{
        \begin{tabular}{l|l|l|l|c|c|c}
          \toprule
          Domain & Method & SLM Decision Space & Role of LLM & Privacy & Efficiency & Task Performance\\
          \midrule
        \multirow{3}{*}{\makecell[l]{Cloud-First\\Collaboration}}
            & Search-R1 \citep{search_r1} & SLM as a tool & tool-augmented LLM & \XSolidBrush & \XSolidBrush & \CheckmarkBold \\
            ~ & RouterDC \citep{router4} & candidate model & candidate answer model & \XSolidBrush & \XSolidBrush & \CheckmarkBold \\
            ~ & RouterLLM \citep{router5} & model selection & candidate answer model & \XSolidBrush & \XSolidBrush & \CheckmarkBold \\
            \midrule
            \multirow{8}{*}{\makecell[l]{Local-First\\Collaboration}}
                & AutoMix \citep{AutoMix} & forward unresolved query & passive tool & \XSolidBrush & \CheckmarkBold & \CheckmarkBold \\
            ~ & LLM Cascade \citep{LLM_cascade} & cascade selection & passive tool & \XSolidBrush & \CheckmarkBold & \CheckmarkBold \\
            ~ & CoGenesis \citep{CoGenesis} & rewrite & reasoning tool & \CheckmarkBold & \XSolidBrush & \CheckmarkBold \\
            ~ & PAPILLON \citep{papillon} & rewrite + post-process & passive tool & \CheckmarkBold & \XSolidBrush & \CheckmarkBold \\
            ~ & PrivacyRestore \citep{privacy_preprocessing2} & removing privacy + restoration & reasoning tool & \CheckmarkBold & \XSolidBrush & \CheckmarkBold \\
            ~ & Self-Disclosures \citep{selfdisclosures} & PII masking & passive tool & \CheckmarkBold & \XSolidBrush & \CheckmarkBold \\
            ~ & Router-R1 \citep{router_r1} & model selection & candidate answer model & \XSolidBrush & \XSolidBrush & \CheckmarkBold \\
            ~ & Privacy-R1 \citep{privacy_r1} & chunk selection & passive tool & \CheckmarkBold & \XSolidBrush & \CheckmarkBold \\
            \midrule
            \multirow{5}{*}{\makecell[l]{Multi-Agent\\System}}
                & ConsensusLLM \citep{rule_agent2} & -- & peer agent & \XSolidBrush & \XSolidBrush & \CheckmarkBold \\
            ~ & MetaGPT \citep{mas_plan} & -- & peer agent & \XSolidBrush & \XSolidBrush & \CheckmarkBold \\
            ~ & AgentVerse \citep{AgentVerse} & -- & peer agent & \XSolidBrush & \XSolidBrush & \CheckmarkBold \\
            ~ & RoCo \citep{mas_robot24} & -- & peer agent & \XSolidBrush & \XSolidBrush & \CheckmarkBold \\
            ~ & MrlX-DeepResearch \citep{mrlx_deepresearch} & -- & peer research agent & \XSolidBrush & \XSolidBrush & \CheckmarkBold \\
          \midrule
          Ours & Dynamic Collaboration Framework & coordinate collaboration & collaborator with active feedback & \CheckmarkBold & \CheckmarkBold & \CheckmarkBold\\
          \bottomrule
        \end{tabular}
        }
  \end{center}
\end{table*}

\subsection{Related Paradigms}

Table~\ref{tab:existing_framework} summarizes representative prior work across five dimensions: SLM decision space, role of LLM, privacy preservation, interaction efficiency, and task performance. \emph{Cloud-first} methods (Search-R1~\citep{search_r1}, RouterDC~\citep{router4}, RouterLLM~\citep{router5}) treat the LLM as the primary inference engine and use SLMs mainly for auxiliary routing or tool use, thus inheriting the privacy and efficiency concerns of cloud-based LLMs. \emph{Local-first} methods place the SLM on the user side but retain static interaction pipelines, which we elaborate in Appendix~\ref{sec:baseline_delta} below. \emph{Multi-agent systems} (MetaGPT~\citep{MetaGPT}, AgentVerse~\citep{AgentVerse}, MrlX-DeepResearch~\citep{mrlx_deepresearch}) emphasize communication and role specialization but typically assume shared task information and symmetric environment access; our setting is instead heterogeneous and information-asymmetric, with the LLM accessing only what the SLM selectively discloses.

\subsection{Methodological Deltas from Prior Local-First Collaboration}\label{sec:baseline_delta}

Beyond the framework-level checkmarks in Table~\ref{tab:existing_framework}, three methodological axes distinguish our framework from prior local-first work.

\textbf{SLM decision space.} Existing methods constrain the SLM to a fixed action shape: confidence-based deferral (AutoMix~\citep{AutoMix}, LLM Cascade~\citep{LLM_cascade}), one-shot rewrite (PAPILLON~\citep{papillon}, PrivacyRestore~\citep{privacy_preprocessing2}, CoGenesis~\citep{CoGenesis}), token-level masking (Self-Disclosures~\citep{selfdisclosures}), single-shot routing among LLMs (Router-R1~\citep{router_r1}), or chunk-level dispatch (Privacy-R1~\citep{privacy_r1}). Our SLM decides both \emph{when} to consult and \emph{how} to formulate each request, iteratively across turns, so the interaction pattern adapts to each query rather than following a fixed template.

\textbf{Role of the LLM.} Prior local-first work treats the LLM as a passive answer engine with no channel for signaling under-specified requests. We instead let the LLM choose between clarification and answering at each turn, turning it into an active collaborator whose feedback also serves as a process-level supervision signal for the SLM.

\textbf{Optimization objective.} Each prior method targets a subset of $\{\text{quality}, \text{efficiency}, \text{privacy}\}$: confidence-based routing trades quality for efficiency; rewriting or masking methods target privacy but do not model interaction cost; RL-based routers optimize task performance while relying on structural constraints for privacy. Our reward integrates all three objectives with tunable trade-off weights and is optimized end-to-end.

These deltas are complementary rather than independent: iterative when and how decisions require an active LLM to correct under-specified requests, and jointly optimizing all three objectives is what drives the SLM to prefer request reformulation over verbatim forwarding.

\section{Privacy Injection Pipeline and PrivQA Dataset Details}\label{sec:dataset_statistics}

\subsection{Privacy Injection Pipeline}

\begin{wrapfigure}{r}{8cm}
\vspace{-1em}
    \centering
    \includegraphics[width=\linewidth]{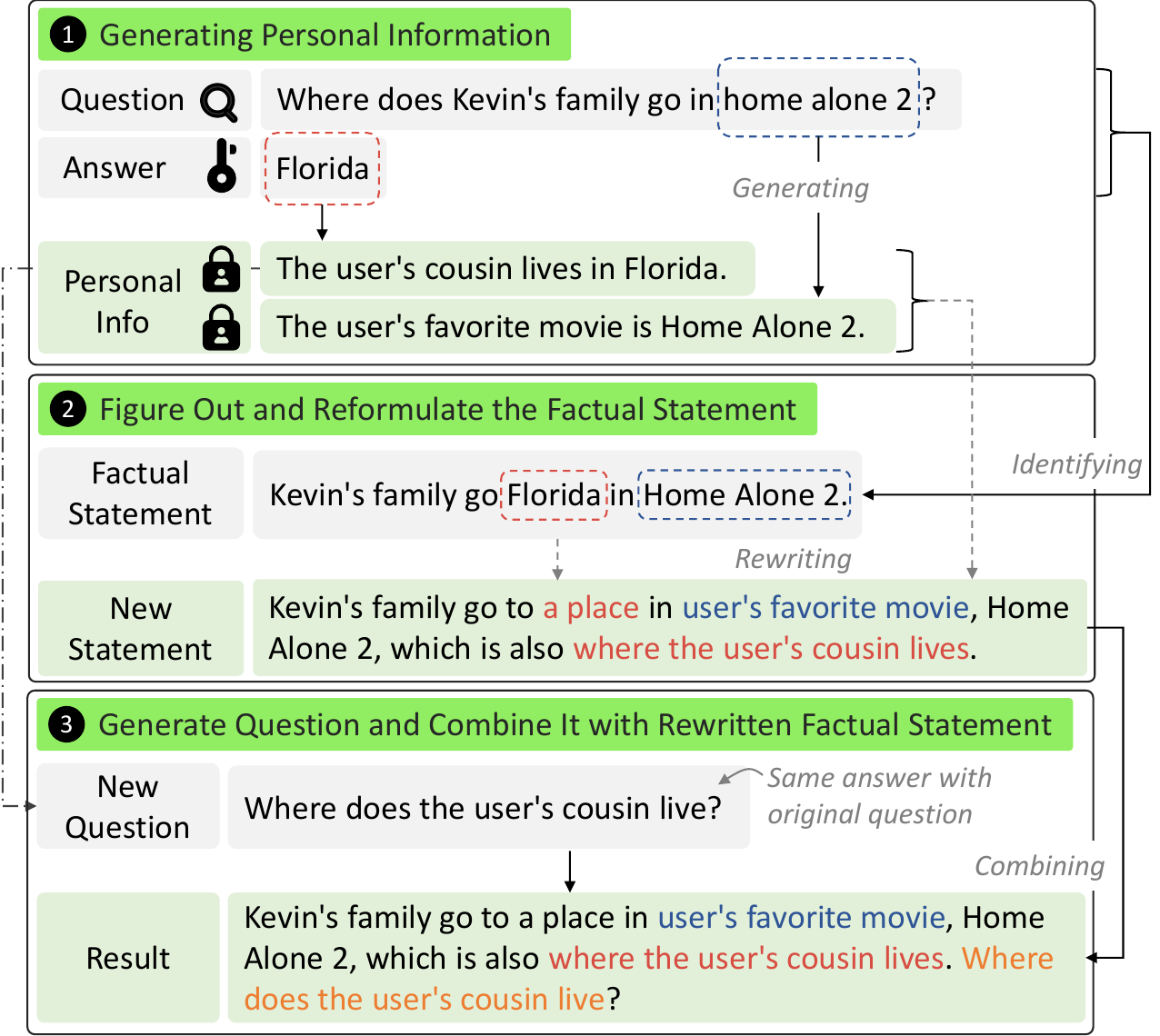}
    \caption{Privacy injection pipeline for PrivQA dataset processing.}\label{fig:dataset}
\end{wrapfigure}

We build PrivQA on question answering because this format is naturally suited to outcome-based RL, where the correctness of an answer provides a verifiable reward signal for learning the collaboration policy. As public QA corpora contain no user-attributed private content, we introduce privacy semantics in a controllable manner through an injection pipeline.
We prompt an LLM to augment each existing QA pair through a three-step process (Figure~\ref{fig:dataset}), which we illustrate below with the seed pair Question: ``Where does Kevin's family go in Home Alone 2?'', Answer: ``Florida.''

\textbf{Step 1: Personal-information generation.} We first generate logically coherent personal information linked to the core entities in the QA pair—in this example, facts such as ``The user's cousin lives in Florida'' and ``The user's favorite movie is Home Alone 2.''

\textbf{Step 2: Indirect reformulation.} We identify the core factual statement (``Kevin's family go to Florida in Home Alone 2.'') and reformulate it so the answer entity ``Florida'' is referenced indirectly through the generated personal information, producing a new statement like ``Kevin's family go to a place in user's favorite movie, Home Alone 2, which is also where the user's cousin lives.''

\textbf{Step 3: Privacy-infused question synthesis.} We synthesize a new question based on the personal information, ``Where does the user's cousin live?'' and combine it with the reformulated statement to create a privacy-infused QA pair: ``Kevin's family go to a place in user's favorite movie, Home Alone 2, which is also where the user's cousin lives. Where does the user's cousin live?'' The answer remains ``Florida,'' but obtaining it now requires inference over embedded personal information, effectively simulating real-world scenarios where users might unwittingly disclose private details.

\textbf{Step 4: Validation.} After generation, we apply an LLM-based filter to remove cases where (1) the answer is altered or (2) the answer is explicitly embedded in the question. Human evaluation on 200 randomly sampled instances further confirms that 99\% are correctly reformulated with accurate answers and appropriate privacy infusion (see Appendix~\ref{sec:human_eval}).

\subsection{PrivQA Dataset Details}\label{sec:dataset_statistics_privqa}

\begin{table}[!ht]
  \caption{Dataset Statistics. NQ and HotpotQA are in-domain datasets with training set, while TriviaQA, PopQA, 2Wiki, and MuSiQuE are out-domain datasets. Out-domain datasets have no training data (denoted by ``--'') and are evaluated on the test set only.}
  \label{tab:dataset}
  \begin{center}
      \resizebox{0.99\linewidth}{!}{
        \begin{tabular}{l|l|c|c|c|c|c|c}
          \toprule
          \multirow{2}{*}{Type} & \multirow{2}{*}{Dataset} & \multicolumn{3}{c|}{Training Set} &  \multicolumn{3}{c}{Test Set} \\
          \cmidrule{3-8}
          ~ & ~ & \# Sample & \# Privacy Info & Avg Input Token & \# Sample & \# Privacy Info & Avg Input Token\\
          \midrule
          \multirow{3}{*}{General} & NQ \citep{nq} & 54,589	&	151,128	&	44.09	&	2,014	&	5,778	&	46.01 \\
          ~ & TriviaQA \citep{triviaqa} & -- & -- & -- & 5,057 & 16,582 & 53.05 \\
          ~ & PopQA \citep{popqa} & -- & -- & -- & 10,448 & 21,940 & 35.56 \\
          \midrule
          \multirow{3}{*}{Multi-Hop} & HotpotQA \citep{hotpotqa} & 65,812	&	228,392	&	55.19	&	5,148	&	17,040	&	53.53 \\
          ~ & 2Wiki \citep{2wiki} & -- & -- & -- & 8,463 & 23,117 & 44.72 \\
          ~ & MuSiQuE \citep{musique} & -- & -- & --  & 1,956 & 6,788 & 55.92 \\
          \bottomrule
        \end{tabular}
        }
  \end{center}
\end{table}

We construct the Privacy QA (PrivQA) dataset by augmenting six public QA datasets with a privacy injection pipeline. To thoroughly evaluate the generalization capability of the learned collaboration strategy, we construct the training set by including only samples derived from NQ and HotpotQA, while the test set covers all six datasets, including NQ, TriviaQA, PopQA, HotpotQA, 2Wiki, and MuSiQuE. 
As summarized in Table \ref{tab:dataset}, the in-domain training set contains a total of 120K samples with an average input token length of 50.16. The out-domain test sets are also diverse, spanning a range of domains and question complexities, with input lengths varying from 35.56 to 55.92 tokens on average. Furthermore, multi-hop datasets tend to have longer inputs, reflecting their compositional reasoning requirements. 

\section{Dataset Synthesis, Training, and Inference Cost} 
\subsection{LLM Usage}\label{sec:llm_usage}
We use Qwen3-Max, a closed-source LLM with strong instruction-following ability, for our PrivQA dataset synthesis. The total token consumption is 55.6M tokens for input and 30.9M tokens for output. The total annotation cost amounts to approximately \$65.59 and the average cost is \$0.0005 per sample. The main LLM used in the collaboration experiment, Qwen3-235B-A22B-Instruct, is deployed on a local server equipped with eight H20 GPUs, and therefore does not incur any API costs.

\subsection{Computational Overhead of Training}\label{sec:computational_overhead}
Our RL-based training incurs approximately 190 GPU-hours for Qwen3-8B (about 24 hours of wall-clock time on 8×H20 GPUs) and 110 GPU-hours for Qwen3-4B (about 14 hours on 8×H20 GPUs). This cost is a one-time offline training investment, and no training is performed on the device. Table \ref{tab:gpu_hours} compares learning-based methods trained on the same base SLM and PrivQA training data. Under matched GPU-hours, our method with Qwen3-8B outperforms SFT with Qwen3-8B by 15.3\% average EM, and our method with Qwen3-4B (110 GPU-hours) already surpasses SFT with Qwen3-8B (140 GPU-hours) despite using less compute. This confirms that the improvement stems from the collaboration design rather than additional supervision or compute.

\begin{table}[!ht]
\centering
\caption{Training cost and average EM of learning-based methods on 8×H20 GPUs, where all methods share the same base SLM and the same PrivQA training data.}\label{tab:gpu_hours}
\resizebox{0.55\linewidth}{!}{
    \begin{tabular}{l|c|c|c}
    \toprule
        Method & Training & GPU-hours & Avg EM \\
    \midrule
        SFT (Qwen3-8B) & SFT & 140 & 27.80 \\
        Router-R1 (Qwen3-8B) & RL & 190 & 37.96 \\
        Ours (Qwen3-4B) & RL & 110 & 38.34 \\
        Ours (Qwen3-8B) & RL & 190 & 43.11 \\
    \bottomrule
    \end{tabular}
}
\end{table}

\subsection{Computational Burden of Inference}\label{sec:cost_energy}
\begin{table}[!ht]
\centering
\caption{Average performance and latency per sample of different methods over PrivQA datasets.}\label{tab:infer_time}
\resizebox{0.55\linewidth}{!}{
    \begin{tabular}{l|l|l}
    \toprule
        Method & Avg EM & Avg Latency Per Sample (ms) \\ 
        \midrule
        Edge SLM Deployment & 25.77 & 243 \\
        \midrule
        Cloud-side LLM Inference & 47.09 & 1096 \\ 
        \midrule
        Ours & 46.01 & 675 \\ 
        \bottomrule
    \end{tabular}
    }
\end{table}

\textbf{Latency.} We report the average per-sample latency for edge SLM deployment, cloud-side LLM inference, and SLM-LLM collaboration in Table \ref{tab:infer_time}. The SLM, Qwen3-8B, is deployed locally on a single NVIDIA 4090 GPU, while the LLM, Qwen3-Max, is accessed via a cloud API. Our collaborative approach delivers performance close to cloud LLM inference (46.01 vs. 47.09 EM) while cutting latency by 38\% (675 vs. 1096 ms), at the cost of $\sim$2.8$\times$ higher latency than local-only edge SLM inference. The worst-case tail latency reaches 2283 ms in Table~\ref{tab:turn_latency}, but is triggered only for the 0.09\% of samples that require multi-turn interaction; the remaining 99.9\% of samples stay well within a typical interactive envelope.

\begin{table}[!ht]
\centering
\caption{Upstream and downstream payload per sample and average end-to-end latency of different methods, where the SLM is Qwen3-8B.}\label{tab:bandwidth}
\resizebox{0.8\linewidth}{!}{
    \begin{tabular}{l|c|c|c}
    \toprule
        Method & Upstream Token & Downstream Token & Avg Latency \\
    \midrule
        LLM Inference & 47.68 & 251.12 & 1096 ms \\
        Self-Disclosures \citep{selfdisclosures} & 44.91 & 264.50 & 1100 ms \\
        PAPILLON \citep{papillon} & 28.60 & 192.11 & 1108 ms \\
        AutoMix \citep{AutoMix} & 21.35 & 305.22 & 428 ms \\
        Router-R1 \citep{router_r1} & 25.18 & 158.60 & 688 ms \\
        Ours & 20.65 & 161.97 & 675 ms \\
    \bottomrule
    \end{tabular}
}
\end{table}

\textbf{Bandwidth, cost, and energy.} We further measure the communication volume between the device and the cloud, the monetary cost of the cloud API, and the energy consumed on the device under the same setting. For each method, we tokenize the upstream payload (tokens sent to the cloud LLM) and the downstream payload (tokens returned by the LLM), and report them together with the latency in Table \ref{tab:bandwidth}. Our method uploads only 20.65 tokens per sample, compared with 47.68 tokens when the raw user query is sent directly to the cloud, a 57\% reduction in upstream bandwidth. Furthermore, on the device side, the RTX 4090 with a TDP of 450 W running for 675 ms consumes only 0.084 Wh per sample. 

\begin{table}[!ht]
\centering
\caption{Distribution of interaction turns and tail latency of our method on each PrivQA test set, where $^*$ represents out-domain datasets. The last row reports the average over the six datasets for the turn distribution and the maximum over all samples for the latency.}\label{tab:turn_latency}
\resizebox{0.65\linewidth}{!}{
    \begin{tabular}{l|c|c|c|c}
    \toprule
        Dataset & 0-turn (\%) & 1-turn (\%) & $\geq$ 2-turn (\%) & Max Latency \\
    \midrule
        NQ & 12.8 & 87.1 & 0.10 & 1459 ms \\
        TriviaQA$^*$ & 20.5 & 79.5 & 0.04 & 1123 ms \\
        PopQA$^*$ & 7.4 & 92.5 & 0.09 & 1872 ms \\
        HotpotQA & 10.4 & 89.5 & 0.10 & 1245 ms \\
        2Wiki$^*$ & 5.0 & 94.8 & 0.21 & 2283 ms \\
        MuSiQuE$^*$ & 12.5 & 87.5 & 0.00 & 761 ms \\
    \midrule
        Overall & 11.4 & 88.5 & 0.09 & 2283 ms \\
    \bottomrule
    \end{tabular}
}
\end{table}

\textbf{Multi-turn distribution.}
Since multi-turn interactions could in principle delay the response, we further report the distribution of interaction turns and the tail latency on each PrivQA test set in Table \ref{tab:turn_latency}. Although the interaction is capped at five turns, only 0.85 turns are used per sample on average, and the interaction terminates within a single turn for 99\% of the samples, with the largest observed turn count being three. The tail latency stays bounded accordingly.

\section{Format of SLM Trajectory}\label{sec:foramt_slm}
\begin{figure}[!ht]
    \centering
    \includegraphics[width=0.95 \linewidth]{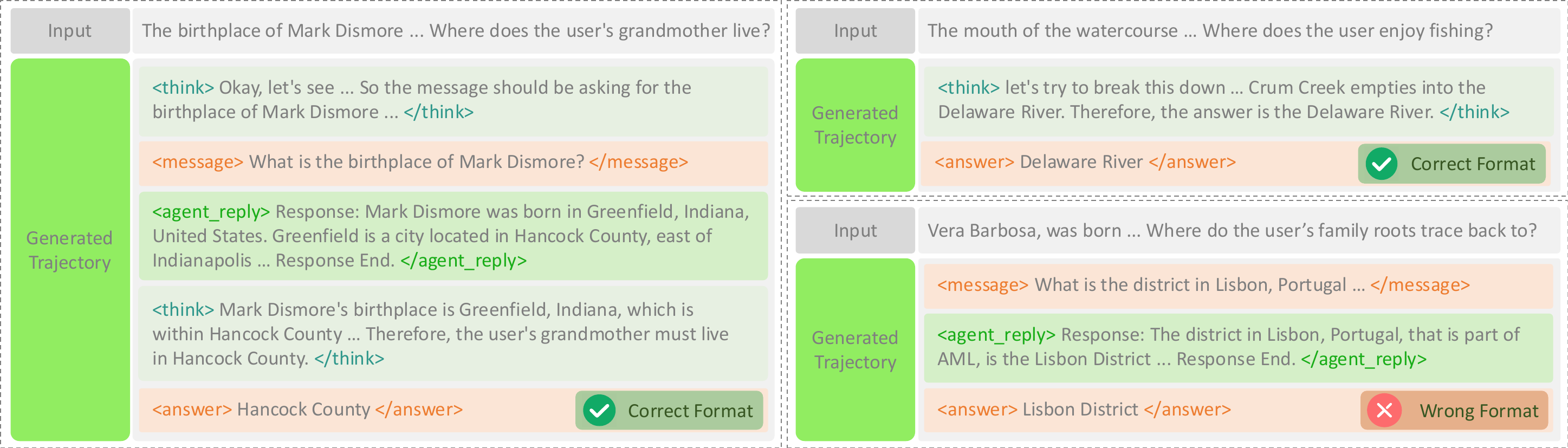}
    \caption{Correct and incorrect cases of format of SLM trajectories.}\label{fig:format_trajectory}
\end{figure}
Figure \ref{fig:format_trajectory} shows examples with correct and incorrect formats. The generated trajectory of the SLM should start with ``\textcolor{myb}{\textless think\textgreater} ... \textcolor{myb}{\textless /think\textgreater}'' and end with ``\textcolor{orange}{\textless answer\textgreater} ... \textcolor{orange}{\textless /answer\textgreater}'' block. When SLM wants to request LLM, it should put its request within ``\textcolor{orange}{\textless message\textgreater} ... \textcolor{orange}{\textless /message\textgreater}'' block and it will receive the LLM's feedback within ``\textcolor{myg}{\textless agent$\_$reply\textgreater} ... \textcolor{myb}{\textless /agent$\_$reply\textgreater}''. There must be exactly one \textcolor{myb}{think} block preceding any \textcolor{orange}{message} block. Moreover, a trajectory may contain no \textcolor{orange}{message} block at all, indicating that the SLM has determined no request is necessary.

\section{Human and Cross-Model Evaluation}\label{sec:human_eval}

\subsection{Accuracy of Privacy Leakage Judgment of the LLM}

To evaluate the accuracy of the privacy leakage judgments made by the LLM, we conduct a human evaluation involving three independent annotators from the laboratory. A random sample of 200 instances is selected, and each annotator independently assesses whether the request generated by the SLM associated with each sample contained a disclosure of the given user's private information. The annotators assign binary labels to each instance, marking it as ``true'' if privacy leakage was present and ``false'' if there was no privacy leakage. The final ground truth labels for each sample are determined using a majority-voting strategy among the three annotators. Based on this annotated ground truth, we evaluated the performance of the LLM's judgments by calculating its accuracy, which reached 98.5\%. We find that the LLM can reliably perform binary classification on whether a request leaks labeled private information.

\subsection{Quality of Privacy QA Synthesis Data}

To assess the quality of the generated privacy-infused QA data, we also conduct a human evaluation with three independent annotators to evaluate a randomly selected set of samples. The annotators follow two main criteria: (1) whether the generated questions containing privacy information are semantically fluent, and (2) whether the answer to the generated question is consistent with the answer to the original question. Each sample is labeled as ``true'' if it meets both criteria, and ``false'' otherwise, following a binary classification approach. The final label for each sample is determined by majority voting among the annotators. The proportion of samples labeled ``true'' is the overall quality score. Through this process, we find that 99\% of the generated data are correctly reformulated with accurate answers and appropriate privacy information infusion.

\subsection{Agreement across Evaluator Model Families}\label{sec:cross_evaluator}

To assess the robustness of our privacy evaluation to the choice of evaluator model, we replace the evaluator with LLMs from different families, including Qwen3-235B-A22B-Instruct, Qwen3-Max, DeepSeek-R1, and DeepSeek-V3, and let each independently judge whether the SLM's requests leak privacy. Their pairwise sample-level agreement is reported in Table \ref{tab:cross_evaluator}.

\begin{table}[!ht]
\centering
\caption{Pairwise sample-level agreement (\%) among four evaluators from different model families.}\label{tab:cross_evaluator}
\resizebox{0.7\linewidth}{!}{
    \begin{tabular}{l|cccc}
    \toprule
        Evaluator & Qwen3-235B & Qwen3-Max & DeepSeek-R1 & DeepSeek-V3 \\
    \midrule
        Qwen3-235B & -- & 98.17 & 97.89 & 97.94 \\
        Qwen3-Max & 98.17 & -- & 97.87 & 97.42 \\
        DeepSeek-R1 & 97.89 & 97.87 & -- & 97.69 \\
        DeepSeek-V3 & 97.94 & 97.42 & 97.69 & -- \\
    \bottomrule
    \end{tabular}
}
\end{table}

All pairwise agreements fall within a narrow band of 97.42\% -- 98.17\%, and the cross-family agreements between the Qwen and DeepSeek models range from 97.42\% to 97.94\%. Since the DeepSeek family is involved in neither the data synthesis nor the training pipeline, its consistent agreement confirms that the near-zero leakage of our method is not an artifact of same-family bias between the injection and evaluation LLMs, and that the privacy evaluation is robust to the choice of evaluator model.

\section{Evaluation metrics}\label{sec:eval_metric_def}

As aligned with our objectives, we design different evaluation metrics to assess (1) the response quality of the SLM, (2) the interaction efficiency between the SLM and the LLM, and (3) the degree of privacy leakage.

\textbf{Performance.} 
We evaluate model performance using Exact Match (EM), a widely adopted metric in question answering (QA) tasks. EM measures the percentage of predictions that are identical to the ground-truth answers. Formally, given a predicted answer $s_{\text{pred}}$ and the ground-truth answer $s_{\text{gt}}$, the EM score is defined as:
\begin{equation}
    \text{EM} = 
    \begin{cases}
        1, & \text{if } s_{\text{pred}} = s_{\text{gt}} \\
        0, & \text{otherwise}
    \end{cases}
\end{equation}
The dataset-level EM is the average over all samples: $\text{EM}_{\text{dataset}} = \frac{1}{N}\sum_{i=1}^N \text{EM}(s^i_{\text{pred}}, s^i_{\text{gt}})$. EM requires perfect string-level agreement between $s_{\text{pred}}$ and $s_{\text{gt}}$. To complement EM and better capture semantic similarity between the predicted and ground-truth answers, we also employ BERTScore \citep{bertscore}. For each answer pair, BERTScore computes a similarity score based on contextual embeddings. The overall BERTScore for the dataset, denoted as $F_{\text{BERT}}$, is calculated by averaging the individual BERTScore values:
\begin{equation}
    F_{\text{BERT}} = \frac{1}{N} \sum_{i=1}^N \text{BERTScore}(s^{i}_{\text{pred}}, s^{i}_{\text{gt}})
\end{equation}
where $N$ is the total number of samples. For the real-world personal-task benchmarks, including PUPA, AESLC, and SAMSum, which are open-ended generation tasks without a unique gold answer, we adopt LLM-as-a-judge with prompted rubrics \citep{llm_as_a_judge}. The specific quality-preservation protocol for PUPA and the five-dimension rubric for AESLC/SAMSum are detailed in Appendix \ref{sec:real_world_details}.

\textbf{Efficiency.}  
We measure the interaction efficiency between the SLM and the LLM by calculating the average number of interaction rounds per sample. A lower average indicates that the SLM is able to independently answer more queries or, when necessary, issue fewer redundant or low-quality requests for assistance.

In addition, we introduce the Interaction Necessity Score (INScore), an efficiency metric calibrated by query difficulty. 
We divide queries into easy or hard based on whether the SLM can answer without assistance. The ideal policy is for the SLM to request assistance only when it cannot answer a query without assistance. We formalize this as a classification problem, where query difficulty serves as the ground-truth label (``easy'' or ``hard''), and the decision to request LLM assistance is treated as the model’s prediction.
The INScore is computed as the average of two recall values: (1) the recall for easy queries, reflecting the proportion of easy
queries for which the SLM does not request assistance, defined as
\begin{equation}
    \text{Recall}_{\text{easy}} = \frac{|{Q}_{\text{w/o req}} \cap {Q}_{\text{easy}}|}{|{Q}_{\text{easy}}|}
\end{equation}
where ${Q}_{\text{easy}}$ denotes the set of easy queries, ${Q}_{\text{w/o req}}$ denotes the set of queries for which the SLM does not issue requests, and (2) the recall for hard queries, reflecting the proportion of hard queries for which the SLM does request assistance, defined as 
\begin{equation}
    \text{Recall}_{\text{hard}} = \frac{|Q_{\text{hard}} \cap Q_{\text{w/ req}}|}{|Q_{\text{hard}}|}
\end{equation}
where $Q_{\text{hard}}$ denotes the set of hard queries and $Q_{\text{w/ req}}$ denotes the set of queries for which the SLM does make a request. Then, the INScore is defined as
\begin{equation}
    \text{INScore} = \frac{\text{Recall}_{\text{easy}} + \text{Recall}_{\text{hard}}}{2}
\end{equation}
In the ideal case where the SLM never requests help for easy queries and always requests help for hard queries, INScore achieves its maximum value of 1.0. By contrast, in two degenerate cases where the SLM always requests help or never requests help (even on hard queries), INScore drops to 0.5. By defining easy and hard relative to each SLM's own standalone capability, INScore serves as a self-calibrated efficiency metric that assesses whether a policy issues requests exactly when its own SLM cannot answer independently.

\textbf{Privacy.}
Formal $\varepsilon$-differential-privacy (DP) guarantees are not directly available for open-ended natural-language interaction, since DP mechanisms rely on bounded sensitivity over a fixed output space that free-form generation lacks \citep{dp}. Privacy in LLM interaction is therefore commonly quantified by item-set based leakage accounting \citep{confaide, papillon}, aligning with inferential-privacy notions in the spirit of Pufferfish privacy \citep{pufferfish_privacy}: rather than bounding output stability, we bound what an honest-but-curious LLM adversary\footnote{Threats outside this model are orthogonal to our accounting: adversarial LLMs (e.g., probing via clarifications) are partially penalized by the evaluator's trajectory-level scoring at training time, with stronger defenses requiring independent trust layers; prompt injection via cited content and side channels (timing, request counts) fall to standard LLM safety and system-level mechanisms.}, observing only the SLM's requests rather than the user's original query, can additionally infer about the user. Concretely, we measure leakage as the increase in the LLM's knowledge caused by these requests, and consider a piece of information privacy-sensitive if its type belongs to a predefined set \citep{selfdisclosure_theory, selfdisclosures}. Under this threat model, the privacy leakage induced by a multi-turn request set is defined as follows.

\begin{definition}[Privacy Leakage in SLM-LLM Interaction]
Let $K_{\text{LLM}}$ denote the prior knowledge of the LLM about the user, which contains no privacy item, $R = \{r_1, \dots, r_T\}$ the multi-turn request set issued by the SLM, and $\text{Priv}_{\text{item}}$ the set of privacy-sensitive item types. The privacy leakage of $R$ is
\begin{equation*}
\text{Leak}(R) = \bigl\{I \mid I \in K_{\text{LLM}} \cup R,\ \text{Type}(I) \in \text{Priv}_{\text{item}}\bigr\} \setminus \bigl\{I \mid I \in K_{\text{LLM}},\ \text{Type}(I) \in \text{Priv}_{\text{item}}\bigr\}.
\end{equation*}
\end{definition}

By construction, $\text{Leak}(R) = \varnothing$ certifies that no new privacy item becomes derivable from $R$ relative to $K_{\text{LLM}}$, so under a randomized SLM policy $M: Q \mapsto R$ the adversary's posterior over the user's privacy items equals its prior. Consequently, $\Pr[\text{Leak}(R) \neq \varnothing]$ upper-bounds the fraction of samples on which the adversary can gain any inferential advantage, up to the false-negative rate of the LLM judge that operationalizes $\text{Leak}(\cdot)$ (bounded at ${\sim}1.5\%$ by human agreement in Appendix~\ref{sec:human_eval}). Our empirical estimator $\text{Priv}_{\text{sample}}$, as defined below, can therefore be read as an empirical upper bound on this per-sample inferential advantage, rather than as a formal $\varepsilon$-guarantee.

Specifically, we evaluate the degree of privacy leakage in the SLM's request sent to the LLM using two metrics that instantiate $\text{Leak}(R)$ on each sample. Both metrics operate at the semantic level, so that paraphrases or generalized references to a private item are captured in the same way as verbatim mentions.
First, we compute the average semantic similarity between the uploaded request ($\text{req}$) and the set of privacy information $P = \{p_1, \dots, p_m\}$ labeled from the original data \citep{query_rewriting1}, where $m$ denotes the number of privacy items, i.e., the user-related items whose types fall in $\text{Priv}_{\text{item}}$. Specifically, the privacy leakage score based on BERTScore is defined as:
\begin{equation}
    \text{Priv}_{\text{sim}} = \frac{1}{|P|} \sum_{i=1}^{m} \text{BERTScore}(\text{req}, p_i).
\end{equation}
A lower score indicates less semantic overlap and thus lower privacy leakage.
Second, we measure the proportion of samples where the SLM issued at least one privacy-leaking request ($\text{Priv}_{\text{sample}}$) with LLM judge \citep{llm_as_a_judge, papillon}. Specifically, given the reference set $P$ and the requests of the SLM concatenated across all interaction turns, the LLM judges whether they allow inference of any user-related private information, i.e., whether $\text{Leak}(R) \neq \varnothing$. Since the judge observes the whole request set jointly rather than each request in isolation, it identifies leaks whether the privacy item is disclosed in a single request or is only recoverable by combining several. In PrivQA, the multi-hop subsets (HotpotQA, 2WikiMultihopQA, MuSiQuE) naturally exercise the latter, since answering their queries already requires composing multiple facts. The reported $\text{Priv}_{\text{sample}}$ is thus measured at the level of the whole request set. For PrivQA, $P$ is given by the privacy items injected during dataset synthesis as logically consistent indirect references produced by the pipeline of Appendix~\ref{sec:dataset_statistics}, so a leak requires inferring the private item semantically from paraphrased content rather than from verbatim disclosure. For real-world user data, $P$ is obtained from dedicated privacy-detection tools.

\section{Proof of Robustness to Evaluator Noise}\label{sec:noise_robust}

Because the privacy reward $r_{\text{privacy}}$ is produced by an LLM judge rather than an exact verifier, we bound the effect of the judge's errors on policy optimization. Prior work \citep{perturbed_reward} shows that a policy trained on a bounded discrete reward corrupted by iid Bernoulli noise converges to the same optimum as under the noiseless reward, with only the convergence speed affected. We adapt this result to our multi-term reward structure in Lemma~\ref{lemma:noise_iid}, and further generalize it in Lemma~\ref{lemma:noise_general} to the case where the flip rates may be asymmetric between false negatives and false positives and vary across inputs.

The privacy penalty $r_{\text{privacy}} \in \{-1, 0\}$ is a sparse binary signal and the only reward term that depends on the judgment of the evaluator LLM, while the format, quality, feedback, and efficiency terms are computed by exact rules. Grouping the exactly computed terms into $C = r_{\text{format}} + r_{\text{quality}} + r_{\text{feedback}} + \beta r_{\text{efficiency}}$, the reward function of Section~\ref{sec:collab_strategy_learning} becomes $R = C + \alpha r_{\text{privacy}}$.

\begin{lemma}[Direction preservation under symmetric iid Bernoulli evaluator noise]\label{lemma:noise_iid}
Assume the evaluator LLM flips its privacy judgment symmetrically and independently with a single probability $p < 1/2$ that does not depend on the input. Then the expected policy gradient under the noisy reward $\hat R$ has the same direction as under the noiseless reward $R$, with the privacy component scaled by a positive factor $(1-2p)$.
\end{lemma}

\begin{proof}
If the evaluator flips its judgment independently with probability $p$, the observed penalty relates to the true one as
\begin{equation*}
\hat{r}_{\text{privacy}} = \left\{
\begin{aligned}
r_{\text{privacy}},&\ \text{with probability } 1-p\\
-1 - r_{\text{privacy}},&\ \text{with probability } p
\end{aligned}
\right.
\end{equation*}
so that a leaking request is missed and a safe request is falsely penalized with the same probability, and its expectation is
\begin{equation*}
\mathbb{E}[\hat{r}_{\text{privacy}}] = (1-p) r_{\text{privacy}} + p (-1 - r_{\text{privacy}}) = (1-2p) r_{\text{privacy}} - p.
\end{equation*}
The value model is trained on the observed reward $\hat{R} = C + \alpha \hat{r}_{\text{privacy}}$ and hence converges to $V(s_0) = \mathbb{E}[\hat{R} \mid s_0]$ for a user query $s_0$. For a rollout $i$, the trajectory-level advantage $\hat{A}_i = \hat{R}_i - V(s_0)$ then satisfies
\begin{equation*}
\mathbb{E}[\hat{A}_i] = \bigl(C_i - \mathbb{E}[C \mid s_0]\bigr) + \alpha (1-2p) \bigl(r_{\text{privacy},i} - \mathbb{E}[r_{\text{privacy}} \mid s_0]\bigr),
\end{equation*}
where the additive bias $-\alpha p$ appears in both $\hat{R}_i$ and $V(s_0)$ and therefore cancels in the advantage; only the reward-magnitude-dependent term $\alpha(1-2p)(r_{\text{privacy},i} - \mathbb{E}[r_{\text{privacy}}|s_0])$ remains. Compared with the noiseless advantage recovered at $p = 0$, evaluator noise leaves the exactly computed terms untouched and scales the privacy component by $(1-2p)$, which is positive whenever $p < 1/2$. Since $\nabla_\theta J = \mathbb{E}[\hat{A}\,\nabla_\theta \log \pi_\theta]$, the noise scales the privacy component of $\nabla_\theta J$ by a positive factor $(1-2p)$ but preserves its sign.
\end{proof}

Lemma~\ref{lemma:noise_iid} assumes that flipping is symmetric between false negatives and false positives and that the flip rate is a single constant $p$ independent of the input. We now show that direction preservation extends to the more general case where both assumptions are relaxed.

\begin{lemma}[Direction preservation under asymmetric and heterogeneous evaluator noise]\label{lemma:noise_general}
Let the evaluator's false-negative and false-positive rates $p_{\mathrm{FN}}(x), p_{\mathrm{FP}}(x)$ depend on the input $x$, with mean combined error $\bar q = \mathbb{E}_x[p_{\mathrm{FN}}(x) + p_{\mathrm{FP}}(x)] < 1$ and variance $\sigma^2 = \operatorname{Var}_x(p_{\mathrm{FN}}(x)) + \operatorname{Var}_x(p_{\mathrm{FP}}(x))$. Then the expected policy gradient under the noisy reward $\hat R$ decomposes as
\begin{equation*}
\nabla_\theta J_{\hat R} \;=\; \nabla_\theta J_R^{\mathrm{exact}} + (1 - \bar q)\, \nabla_\theta J_R^{\mathrm{priv}} + \varepsilon,
\qquad
\lVert \varepsilon \rVert_2 \;\le\; (1+\sqrt 2)\, \alpha\, \sigma \sqrt{\mathbb{E}_x \lVert \nabla_\theta \log \pi_\theta(\cdot\mid x) \rVert^2},
\end{equation*}
where $\nabla_\theta J_R^{\mathrm{exact}}$ groups the exactly-computed reward terms and $\nabla_\theta J_R^{\mathrm{priv}}$ is the privacy-component gradient under the noiseless reward.
\end{lemma}

\begin{proof}
Let $y = -r_{\text{privacy}} \in \{0,1\}$ be the leak indicator and $q(x) = p_{\mathrm{FN}}(x) + p_{\mathrm{FP}}(x)$. Under per-input flip rates, a direct calculation gives
\begin{equation*}
\mathbb{E}[\hat r_{\text{privacy}} \mid r_{\text{privacy}}, x] \;=\; -p_{\mathrm{FP}}(x) + (1 - q(x))\, r_{\text{privacy}}.
\end{equation*}
Writing $\Delta_{\mathrm{FP}}(x) = p_{\mathrm{FP}}(x) - \bar p_{\mathrm{FP}}$ and $\Delta_q(x) = q(x) - \bar q$, expanding the trajectory-level advantage $\hat A_i = \hat R_i - V(s_0)$ with $V(s_0) = \mathbb{E}_x[\hat R \mid s_0]$, and applying $\nabla_\theta J = \mathbb{E}_x[\hat A\, \nabla_\theta \log \pi_\theta]$ (constants vanish since $\mathbb{E}_x[\nabla_\theta \log \pi_\theta] = 0$), the noisy gradient becomes
\begin{align*}
\nabla_\theta J_{\hat R}
&= \nabla_\theta J_R^{\mathrm{exact}} + (1-\bar q)\, \nabla_\theta J_R^{\mathrm{priv}} \\
&\quad - \alpha\, \mathbb{E}_x[\Delta_{\mathrm{FP}}(x)\, \nabla_\theta \log \pi_\theta] - \alpha\, \mathbb{E}_x[\Delta_q(x)\, r_{\text{privacy}}(x)\, \nabla_\theta \log \pi_\theta].
\end{align*}
Denote the last two terms by $\varepsilon$. Applying Cauchy--Schwarz componentwise, using $|r_{\text{privacy}}| \le 1$, and bounding $\operatorname{Var}(q) \le 2\sigma^2$ via $2\operatorname{Cov}(p_{\mathrm{FN}}, p_{\mathrm{FP}}) \le \operatorname{Var}(p_{\mathrm{FN}}) + \operatorname{Var}(p_{\mathrm{FP}})$,
\begin{equation*}
\lVert \varepsilon \rVert_2
\;\le\; \alpha \Bigl( \sqrt{\operatorname{Var}(p_{\mathrm{FP}})} + \sqrt{2}\,\sigma \Bigr) \sqrt{\mathbb{E}_x \lVert \nabla_\theta \log \pi_\theta \rVert^2}
\;\le\; (1+\sqrt 2)\, \alpha\, \sigma \sqrt{\mathbb{E}_x \lVert \nabla_\theta \log \pi_\theta \rVert^2}.
\end{equation*}
Two special cases follow. (i)~At $\sigma = 0$, $\varepsilon = 0$ and the direction of $\nabla_\theta J_{\hat R}$ is preserved exactly, with only a positive shrinkage $(1-\bar q) > 0$ on the privacy component. (ii)~At $p_{\mathrm{FN}}(x) = p_{\mathrm{FP}}(x) = p$, we have $\bar q = 2p$, $\sigma = 0$, and Lemma~\ref{lemma:noise_iid} is recovered.
\end{proof}

Two independent measurements in our setting jointly constrain the assumptions of both lemmas. The human evaluation in Appendix~\ref{sec:human_eval} reports $98.5\%$ agreement with the majority-vote annotations of three annotators, upper-bounding the disagreement rate at $\approx 0.015$; interpreted under Lemma~\ref{lemma:noise_iid} this gives $p \le 0.015$, and interpreted under Lemma~\ref{lemma:noise_general} it corresponds to $\bar q \lesssim 0.03$. In either reading, the dominant scaling remains at least $0.97$ and the attenuation of the privacy signal is negligible. Cross-family agreement across four evaluators from the Qwen and DeepSeek families stays within a $97.42$ -- $98.17\%$ band (Appendix~\ref{sec:cross_evaluator}, Table~\ref{tab:cross_evaluator}), so no single family's systematic bias dominates; treating cross-family variance as an empirical proxy for the input-conditional variance $\sigma$ places $\sigma$ on the order of $10^{-2}$, so the residual heterogeneity term $\varepsilon$ contributes at most a few percent of the dominant scaled gradient. Consistent with these bounds, the training reward in Figure~\ref{fig:slm_reward} increases monotonically without collapse.

\section{Additional Experiment Details and Results}\label{sec:addition_exprement}

\subsection{Additional Training Details}\label{sec:implementation_details}
We train with VAPO \citep{vapo} for 600 steps at a batch size of 256 and a sampling temperature of 1.0. The maximum context length is 8{,}192 tokens for the local SLM and 16{,}384 tokens for the cloud LLM. The actor and critic learning rates are 1e-6 and 2e-6, respectively, and we follow VAPO in setting the asymmetric PPO clip ratios to $\varepsilon_{\text{low}}=0.2$ and $\varepsilon_{\text{high}}=0.28$.

For a fair comparison, all baselines share the same SLM-LLM pair, maximum context length, and parameter dtype as our method. Method-specific settings, including SLM and LLM prompts, strictly follow the original papers \citep{AutoMix, selfdisclosures, papillon, router_r1}.

\subsection{Gains of Dynamic Collaboration}
\begin{table*}[!ht]
\centering
\caption{Performance of SLM with dynamic collaboration with Qwen3-235B-A22B-Instruct and baselines on different datasets.}\label{tab:baseline_result_full}
\resizebox{1\linewidth}{!}{
    \begin{tabular}{l|l|ll|ll|ll|ll|ll|ll|ll}
    \toprule
        \multirow{3}{*}{Method} & \multirow{3}{*}{Model} & \multicolumn{6}{c|}{General QA} & \multicolumn{6}{c|}{Multi-Hop QA} & \multicolumn{2}{c}{\multirow{2}{*}{Avg}}\\
        \cmidrule{3-14}
        ~ & ~ & \multicolumn{2}{c|}{NQ} & \multicolumn{2}{c|}{TriviaQA$^*$} & \multicolumn{2}{c|}{PopQA$^*$} & \multicolumn{2}{c|}{HotpotQA} & \multicolumn{2}{c|}{2Wiki$^*$} & \multicolumn{2}{c|}{MuSiQuE$^*$} & ~ & ~ \\
        \cmidrule{3-16}
        ~ & ~ & EM $\uparrow$ & $F_{\text{BERT}}$ $\uparrow$ & EM $\uparrow$ & $F_{\text{BERT}}$ $\uparrow$ & EM $\uparrow$ & $F_{\text{BERT}}$ $\uparrow$ & EM $\uparrow$ & $F_{\text{BERT}}$ $\uparrow$ & EM $\uparrow$ & $F_{\text{BERT}}$ $\uparrow$ & EM $\uparrow$ & $F_{\text{BERT}}$ $\uparrow$ & EM $\uparrow$ & $F_{\text{BERT}}$ $\uparrow$ \\
        \midrule
        CoT & Qwen3-235B & 40.91 & 75.89 & 66.30 & 85.95 & 45.44 & 73.08 & 30.81 & 69.26 & 47.91 & 76.81 & 23.62 & 66.23 & 42.50 & 74.54 \\
        \midrule
        \multirow{2}{*}{\makecell[l]{SFT}} & Qwen3-4B & 24.88 & 68.31 & 34.13 & 55.81 & 18.19 & 57.42 & 13.54 & 62.36 & 33.20 & 63.44 & 12.37 & 63.91 & 22.72 & 61.87 \\
        ~ & Qwen3-8B & 33.91 & 72.11 & 44.35 & 59.29 & 22.97 & 59.87 & 17.02 & 64.64 & 33.90 & 64.54 & 14.62 & 64.40 & 27.80 & 64.14 \\
        \midrule
        \multirow{2}{*}{\makecell[l]{AutoMix \cite{AutoMix}}} & Qwen3-4B & 29.00 & 68.87 & 51.73 & 78.10 & 32.22 & 66.38 & 21.02 & 61.29 & 40.35 & 71.92 & 16.41 & 56.71 & 31.79 & 67.21 \\
        ~ & Qwen3-8B & 32.52 & 50.86 & 56.85 & 51.70 & 33.92 & 53.34 & 21.91 & 49.38 & 40.27 & 54.03 & 17.18 & 49.15 & 33.78 & 51.41 \\
        \midrule
        \multirow{2}{*}{\makecell[l]{Self-Disclosures \\ \cite{selfdisclosures}}} & Qwen3-4B & 27.26 & 63.62 & 46.93 & 74.50 & 23.54 & 54.79 & 17.70 & 57.42 & 24.81 & 58.36 & 12.78 & 57.16 & 25.50 & 60.98 \\
        ~ & Qwen3-8B & 30.69 & 65.62 & 51.28 & 77.33 & 26.29 & 56.94 & 21.87 & 61.07 & 28.24 & 61.78 & 16.16 & 59.86 & 29.09 & 63.77 \\
        \midrule
        \multirow{2}{*}{\makecell[l]{PAPILLON \cite{papillon}}} & Qwen3-4B & 32.77 & 68.03 & 54.78 & 79.27 & 36.73 & 65.80 & 22.38 & 61.38 & 33.03 & 65.91 & 15.64 & 58.61 & 32.55 & 66.50 \\
        ~ & Qwen3-8B & 35.40 & 69.66 & 56.36 & 80.39 & 35.88 & 64.78 & 23.82 & 62.55 & 31.47 & 64.74 & 16.46 & 58.32 & 33.23 & 66.74 \\
        \midrule
        \multirow{2}{*}{\makecell[l]{Router-R1 \cite{router_r1}}} & Qwen3-4B & 35.33 & 71.15 & 57.81 & 80.33 & 37.91 & 67.07 & 23.60 & 63.75 & 40.42 & 71.22 & 17.15 & 60.94 & 35.37 & 69.08 \\
        ~ & Qwen3-8B & 39.08 & 74.00 & 60.59 & 83.12 & 42.20 & 71.45 & 24.65 & 64.92 & \underline{43.48} & 73.73 & 17.74 & 61.05 & 37.96 & 71.38 \\
        \midrule
        \multirow{2}{*}{Ours} & Qwen3-4B & \underline{39.42} & \underline{76.43} & \underline{61.82} & \underline{83.95} & \underline{44.17} & \underline{74.08} & \underline{25.51} & \underline{67.30} & 40.52 & \underline{74.33} & \underline{18.61} & \underline{65.00} & \underline{38.34} & \underline{73.51} \\
        ~ & Qwen3-8B & \textbf{46.08} & \textbf{78.51} & \textbf{64.64} & \textbf{85.32} & \textbf{47.87} & \textbf{75.58} & \textbf{29.66} & \textbf{69.21} & \textbf{47.09} & \textbf{77.35} & \textbf{23.31} & \textbf{67.34} & \textbf{43.11} & \textbf{75.55} \\
        \bottomrule
    \end{tabular}
}
\end{table*}

\begin{table*}[!ht]
\centering
\caption{Privacy leakage of SLM with dynamic collaboration with Qwen3-235B-A22B-Instruct and static interaction framework on different datasets.}\label{tab:privacy_result_full}
\resizebox{1\linewidth}{!}{
    \begin{tabular}{l|l|ll|ll|ll|ll|ll|ll|ll}
    \toprule
        \multirow{3}{*}{Method} & \multirow{3}{*}{Model} & \multicolumn{6}{c|}{General QA} & \multicolumn{6}{c|}{Multi-Hop QA} & \multicolumn{2}{c}{\multirow{2}{*}{Avg}}\\
        \cmidrule{3-14}
        ~ & ~ & \multicolumn{2}{c|}{NQ} & \multicolumn{2}{c|}{TriviaQA$^*$} & \multicolumn{2}{c|}{PopQA$^*$} & \multicolumn{2}{c|}{HotpotQA} & \multicolumn{2}{c|}{2Wiki$^*$} & \multicolumn{2}{c|}{MuSiQuE$^*$} & ~ & ~ \\
        \cmidrule{3-16}
        ~ & ~ & $\text{Priv}_{\text{sim}}$ $\downarrow$ & $\text{Priv}_{\text{sample}}$ $\downarrow$ & $\text{Priv}_{\text{sim}}$ $\downarrow$ & $\text{Priv}_{\text{sample}}$ $\downarrow$ & $\text{Priv}_{\text{sim}}$ $\downarrow$ & $\text{Priv}_{\text{sample}}$ $\downarrow$ & $\text{Priv}_{\text{sim}}$ $\downarrow$ & $\text{Priv}_{\text{sample}}$ $\downarrow$ & $\text{Priv}_{\text{sim}}$ $\downarrow$ & $\text{Priv}_{\text{sample}}$ $\downarrow$ & $\text{Priv}_{\text{sim}}$ $\downarrow$ & $\text{Priv}_{\text{sample}}$ $\downarrow$ & $\text{Priv}_{\text{sim}}$ $\downarrow$ & $\text{Priv}_{\text{sample}}$ $\downarrow$ \\
        \midrule
        CoT & Qwen3-235B & 66.53 & 100.00 & 64.40 & 100.00 & 68.34 & 100.00 & 64.34 & 100.00 & 67.96 & 100.00 & 64.50 & 100.00 & 66.01 & 100.00 \\
        \midrule
        \multirow{2}{*}{\makecell[l]{SFT}} & Qwen3-4B & 0.00 & 0.00 & 0.00 & 0.00 & 0.00 & 0.00 & 0.00 & 0.00 & 0.00 & 0.00 & 0.00 & 0.00 & 0.00 & 0.00 \\
        ~ & Qwen3-8B & 0.00 & 0.00 & 0.00 & 0.00 & 0.00 & 0.00 & 0.00 & 0.00 & 0.00 & 0.00 & 0.00 & 0.00 & 0.00 & 0.00 \\
        \midrule
        \multirow{2}{*}{\makecell[l]{AutoMix \cite{AutoMix}}} & Qwen3-4B & 62.21 & 72.05 & 60.02 & 68.36 & 66.05 & 76.61 & 63.55 & 76.30 & 65.07 & 75.02 & 64.42 & 77.51 & 63.55 & 74.31 \\
        ~ & Qwen3-8B & 62.15 & 64.45 & 59.08 & 60.96 & 66.01 & 72.95 & 62.99 & 71.60 & 64.55 & 70.50 & 63.91 & 72.96 & 63.12 & 68.90 \\
        \midrule
        \multirow{2}{*}{\makecell[l]{Self-Disclosures \\ \cite{selfdisclosures}}} & Qwen3-4B & 64.20 & 57.65 & 62.72 & 62.67 & 64.71 & 58.49 & 62.67 & 62.45 & 63.90 & 50.56 & 62.48 & 59.76 & 63.45 & 58.60 \\
        ~ & Qwen3-8B & 62.50 & 51.54 & 61.26 & 54.48 & 63.24 & 55.57 & 61.19 & 51.24 & 62.14 & 45.20 & 60.95 & 51.23 & 61.88 & 51.54 \\
        \midrule
        \multirow{2}{*}{\makecell[l]{PAPILLON \cite{papillon}}} & Qwen3-4B & 62.89 & 30.14 & 61.50 & 37.87 & 63.23 & 36.97 & 61.86 & 32.96 & 63.85 & 25.82 & 62.19 & 32.11 & 62.58 & 32.64 \\
        ~ & Qwen3-8B & 60.57 & 22.99 & 58.81 & 29.84 & 60.00 & 18.80 & 60.07 & 28.73 & 62.58 & 18.69 & 60.29 & 27.20 & 60.39 & 24.37 \\
        \midrule
        \multirow{2}{*}{\makecell[l]{Router-R1 \cite{router_r1}}} & Qwen3-4B & 62.09 & 28.02 & 58.84 & 31.99 & 63.76 & 30.75 & 62.34 & 29.56 & 61.59 & 20.68 & 62.05 & 31.78 & 61.78 & 28.80 \\
        ~ & Qwen3-8B & 58.56 & 23.28 & 53.87 & 24.82 & 59.90 & 18.01 & 56.90 & 26.33 & 58.63 & 13.91 & 57.28 & 19.71 & 57.52 & 21.01 \\
        \midrule
        \multirow{2}{*}{Ours} & Qwen3-4B & \underline{47.80} & \underline{0.149} & \underline{42.03} & \underline{0.356} & \underline{50.92} & \underline{0.048} & \underline{48.38} & \underline{0.350} & \underline{54.61} & \underline{0.225} & \underline{46.56} & \underline{0.409} & \underline{48.38} & \underline{0.256} \\
        ~ & Qwen3-8B & \textbf{43.51} & \textbf{0.120} & \textbf{35.33} & \textbf{0.110} & \textbf{50.83} & \textbf{0.010} & \textbf{48.21} & \textbf{0.039} & \textbf{53.86} & \textbf{0.050} & \textbf{48.32} & \textbf{0.153} & \textbf{46.68} & \textbf{0.080} \\
        \bottomrule
    \end{tabular}
}
\end{table*}

\begin{table*}[!ht]
\centering
\caption{Efficiency performance of SLM with dynamic collaboration with Qwen3-235B-A22B-Instruct and baselines on different datasets.}\label{tab:efficiency_result_full}
\resizebox{1\linewidth}{!}{
    \begin{tabular}{l|l|ll|ll|ll|ll|ll|ll|ll}
    \toprule
        \multirow{3}{*}{Method} & \multirow{3}{*}{Model} & \multicolumn{6}{c|}{General QA} & \multicolumn{6}{c|}{Multi-Hop QA} & \multicolumn{2}{c}{\multirow{2}{*}{Avg}}\\
        \cmidrule{3-14}
        ~ & ~ & \multicolumn{2}{c|}{NQ} & \multicolumn{2}{c|}{TriviaQA$^*$} & \multicolumn{2}{c|}{PopQA$^*$} & \multicolumn{2}{c|}{HotpotQA} & \multicolumn{2}{c|}{2Wiki$^*$} & \multicolumn{2}{c|}{MuSiQuE$^*$} & ~ & ~ \\
        \cmidrule{3-16}
        ~ & ~ & Avg Turn $\downarrow$ & INScore $\uparrow$ & Avg Turn $\downarrow$ & INScore $\uparrow$ & Avg Turn $\downarrow$ & INScore $\uparrow$ & Avg Turn $\downarrow$ & INScore $\uparrow$ & Avg Turn $\downarrow$ & INScore $\uparrow$ & Avg Turn $\downarrow$ & INScore $\uparrow$ & Avg Turn $\downarrow$ & INScore $\uparrow$ \\
        \midrule
        CoT & Qwen3-235B & 1.000 & 50.00 & 1.000 & 50.00 & 1.000 & 50.00 & 1.000 & 50.00 & 1.000 & 50.00 & 1.000 & 50.00 & 1.000 & 50.00 \\
        \midrule
        \multirow{2}{*}{\makecell[l]{SFT}} & Qwen3-4B & 0.000 & 50.00 & 0.000 & 50.00 & 0.000 & 50.00 & 0.000 & 50.00 & 0.000 & 50.00 & 0.000 & 50.00 & 0.000 & 50.00 \\
        ~ & Qwen3-8B & 0.000 & 50.00 & 0.000 & 50.00 & 0.000 & 50.00 & 0.000 & 50.00 & 0.000 & 50.00 & 0.000 & 50.00 & 0.000 & 50.00 \\
        \midrule
        \multirow{2}{*}{\makecell[l]{AutoMix \cite{AutoMix}}} & Qwen3-4B & \underline{0.720} & 57.75 & 0.684 & 57.96 & \underline{0.766} & 59.61 & \underline{0.763} & 54.68 & \underline{0.750} & 56.72 & \underline{0.775} & 52.94 & \underline{0.743} & 56.61 \\
        ~ & Qwen3-8B & \textbf{0.644} & 60.17 & \textbf{0.610} & 58.94 & \textbf{0.730} & 59.41 & \textbf{0.716} & 56.01 & \textbf{0.705} & \underline{58.73} & \textbf{0.730} & 55.12 & \textbf{0.689} & 58.06 \\
        \midrule
        \multirow{2}{*}{\makecell[l]{Self-Disclosures \\ \cite{selfdisclosures}}} & Qwen3-4B & 1.000 & 50.00 & 1.000 & 50.00 & 1.000 & 50.00 & 1.000 & 50.00 & 1.000 & 50.00 & 1.000 & 50.00 & 1.000 & 50.00 \\
        ~ & Qwen3-8B & 1.000 & 50.00 & 1.000 & 50.00 & 1.000 & 50.00 & 1.000 & 50.00 & 1.000 & 50.00 & 1.000 & 50.00 & 1.000 & 50.00 \\
        \midrule
        \multirow{2}{*}{\makecell[l]{PAPILLON \cite{papillon}}} & Qwen3-4B & 1.000 & 50.00 & 1.000 & 50.00 & 1.000 & 50.00 & 1.000 & 50.00 & 1.000 & 50.00 & 1.000 & 50.00 & 1.000 & 50.00 \\
        ~ & Qwen3-8B & 1.000 & 50.00 & 1.000 & 50.00 & 1.000 & 50.00 & 1.000 & 50.00 & 1.000 & 50.00 & 1.000 & 50.00 & 1.000 & 50.00 \\
        \midrule
        \multirow{2}{*}{\makecell[l]{Router-R1 \cite{router_r1}}} & Qwen3-4B & 1.145 & 51.47 & 1.111 & 53.57 & 1.194 & 51.92 & 1.165 & 51.61 & 1.129 & 53.66 & 1.180 & 54.56 & 1.154 & 52.80 \\
        ~ & Qwen3-8B & 1.012 & 54.70 & 0.831 & 57.34 & 1.203 & 55.30 & 1.101 & 53.56 & 1.161 & {57.12} & 1.140 & 55.12 & 1.075 & 55.52 \\
        \midrule
        \multirow{2}{*}{Ours} & Qwen3-4B & 0.873 & \underline{60.50} & 0.796 & \underline{61.93} & 0.927 & \textbf{60.50} & 0.897 & \textbf{59.97} & 0.953 & 53.81 & 0.875 & \underline{60.73} & 0.887 & \underline{59.57} \\
        ~ & Qwen3-8B & 0.788 & \textbf{60.97} & \underline{0.668} & \textbf{64.27} & 0.926 & \underline{60.23} & 0.891 & \underline{57.91} & 0.920 & \textbf{60.06} & 0.898 & \textbf{62.00} & 0.849 & \textbf{60.91} \\
        \bottomrule
    \end{tabular}
}
\end{table*}

Tables \ref{tab:baseline_result_full} - \ref{tab:efficiency_result_full} provide the per-dataset breakdown of the aggregate results in Table~\ref{tab:baseline_resutl} of the main text, along the three axes: quality, privacy, and efficiency.

\textbf{Quality (Table~\ref{tab:baseline_result_full}).} Our method is competitive with, and on several splits surpasses, the Qwen3-235B teacher: on NQ, Qwen3-8B reaches an EM of 46.08 versus the teacher's 40.91. Aggregated over the six datasets, our method improves average EM by 3.0\% -- 15.6\% on Qwen3-4B and 5.2\% -- 15.3\% on Qwen3-8B over the baselines.

\textbf{Privacy (Table~\ref{tab:privacy_result_full}).} We find that SLMs with a dynamic collaboration framework learn to preserve privacy almost perfectly. For Qwen3-4B and Qwen3-8B, $\text{Priv}_\text{sample}$ scores are 0.26\% and 0.08\%, respectively. Furthermore, compared to the dynamic collaboration framework, static interaction leads to more privacy leakage, increasing $\text{Priv}_{\text{sim}}$ by 13.4\% -- 15.2\% and $\text{Priv}_{\text{sample}}$ by 28.5\% -- 74.1\% on Qwen3-4B.

\textbf{Efficiency (Table~\ref{tab:efficiency_result_full}).} By adaptively deciding when to request the LLM, our method improves INScore by 3.0\% -- 9.6\% on Qwen3-4B and 2.8\% -- 10.9\% on Qwen3-8B over baselines, while cutting average interaction turns by 0.11 -- 0.27 and 0.15 -- 0.23, respectively.

\subsection{Mitigating Confident Hallucination on Hard Queries}\label{sec:hard_subset}

Standalone SLMs often answer confidently wrong on queries beyond their capability rather than seeking help. We isolate this failure mode by defining the \emph{hard subset} of each PrivQA test set as the samples where standalone Qwen3-4B with CoT fails (EM $= 0$ by construction). On this subset, we report the EM of our framework and the fraction of hard samples for which the SLM issues a request, denoted $P(\text{req} \mid \text{hard})$.

\begin{table}[!ht]
\centering
\caption{Performance of our framework (Qwen3-4B collaborating with Qwen3-235B-A22B-Instruct) on the hard subset of PrivQA, defined as the samples where standalone Qwen3-4B with CoT fails; $^*$ marks out-domain datasets.}\label{tab:hard_subset}
\resizebox{0.52\linewidth}{!}{
    \begin{tabular}{l|c|c|c}
    \toprule
        Dataset & Hard Ratio & Our EM $\uparrow$ & $P(\text{req} \mid \text{hard})$ $\uparrow$ \\
    \midrule
        NQ & 81.1 & 29.13 & 91.19 \\
        TriviaQA$^*$ & 62.4 & 43.21 & 88.50 \\
        PopQA$^*$ & 83.9 & 35.98 & 95.93 \\
        HotpotQA & 89.0 & 18.94 & 91.75 \\
        2Wiki$^*$ & 67.8 & 22.13 & 97.47 \\
        MuSiQuE$^*$ & 89.9 & 12.80 & 89.65 \\
    \midrule
        Avg & 79.0 & \textbf{27.03} & \textbf{92.41} \\
    \bottomrule
    \end{tabular}
}
\end{table}

Hard samples make up 79.0\% of the six test sets on average. Where standalone Qwen3-4B would score zero, our framework recovers 27.03\% EM while requesting the LLM on 92.41\% of these queries. This behavior falls out of the outcome-based quality reward: on a hard sample, a local-answer rollout receives $r_{\text{quality}} = 0$ while a rollout that consults the LLM can receive $r_{\text{quality}} = 1$, so the requesting trajectory carries a positive advantage and the policy is pushed to consult whenever the sample genuinely exceeds the SLM's capability.

\subsection{Generalization to Real-World Personal Tasks}\label{sec:real_world_details}

\begin{table*}[!ht]
\centering
\caption{Quality and privacy leakage of dynamic collaboration and baselines on real-world personal-task benchmarks.}\label{tab:real_world_full}
    \vspace{-0.4em}
\resizebox{1\linewidth}{!}{
    \begin{tabular}{l|l|ll|ll|ll|ll}
    \toprule
        \multirow{2}{*}{Method} & \multirow{2}{*}{Model} & \multicolumn{2}{c|}{PUPA \citep{papillon}} & \multicolumn{2}{c|}{AESLC \citep{aeslc}} & \multicolumn{2}{c|}{SAMSum \citep{samsum}} & \multicolumn{2}{c}{Avg.} \\
        \cmidrule{3-10}
        ~ & ~ & Qual. $\uparrow$ & $\text{Priv}_{\text{sample}}$ $\downarrow$ & Qual. $\uparrow$ & $\text{Priv}_{\text{sample}}$ $\downarrow$ & Qual. $\uparrow$ & $\text{Priv}_{\text{sample}}$ $\downarrow$ & Qual. $\uparrow$ & $\text{Priv}_{\text{sample}}$ $\downarrow$ \\
        \midrule
        CoT & Qwen3-235B & 98.95 & 100.00 & 92.03 & 100.00 & 94.64 & 100.00 & 95.21 & 100.00 \\
        \midrule
        \multirow{2}{*}{\makecell[l]{SFT}} & Qwen3-4B & 70.14 & 0.00 & 74.43 & 0.00 & 74.76 & 0.00 & 73.11 & 0.00 \\
        ~ & Qwen3-8B & 80.42 & 0.00 & 84.82 & 0.00 & 82.63 & 0.00 & 82.62 & 0.00 \\
        \midrule
        \multirow{2}{*}{\makecell[l]{AutoMix \citep{AutoMix}}} & Qwen3-4B & 73.64 & 29.14 & 86.13 & 24.53 & 86.44 & 27.65 & 82.07 & 27.11 \\
        ~ & Qwen3-8B & 83.23 & 25.28 & 88.19 & 19.49 & 89.47 & 21.74 & 86.96 & 22.17 \\
        \midrule
        \multirow{2}{*}{\makecell[l]{Self-Disclosures \citep{selfdisclosures}}} & Qwen3-4B & 72.00 & 17.43 & 81.88 & 15.65 & 88.43 & 17.79 & 80.77 & 23.20 \\
        ~ & Qwen3-8B & 79.71 & 15.71 & 85.58 & 9.56 & 90.44 & 11.82 & 85.24 & 18.98 \\
        \midrule
        \multirow{2}{*}{\makecell[l]{PAPILLON \citep{papillon}}} & Qwen3-4B & 72.96 & 7.57 & 89.60 & 8.65 & 89.06 & 9.44 & 83.87 & 8.55 \\
        ~ & Qwen3-8B & 81.87 & 3.58 & 90.56 & 7.26 & 91.91 & 9.19 & \underline{88.11} & 6.68 \\
        \midrule
        \multirow{2}{*}{\makecell[l]{Router-R1 \citep{router_r1}}} & Qwen3-4B & 73.92 & 7.09 & 87.19 & 6.42 & 87.16 & 8.37 & 82.76 & 7.29 \\
        ~ & Qwen3-8B & 85.39 & 6.26 & 88.83 & 5.18 & 89.78 & 7.65 & 88.00 & 6.36 \\
        \midrule
        \multirow{2}{*}{Ours} & Qwen3-4B & 80.85 & \underline{1.79} & \underline{89.77} & \underline{0.27} & 90.21 & 1.24 & 86.94 & \underline{1.10} \\
        ~ & Qwen3-8B & \textbf{90.96} & \textbf{0.82} & \textbf{91.68} & \textbf{1.09} & \textbf{93.69} & \textbf{1.12} & \textbf{92.11} & \textbf{1.01} \\
        \bottomrule
    \end{tabular}
 }
\vspace{-1em}
\end{table*}

To evaluate whether the collaboration strategy learned on PrivQA transfers to real-world personal tasks without additional training, we zero-shot deploy on three real-world personal-task benchmarks. PUPA \citep{papillon} is a real-world email- and document-writing task (613 test samples) whose PII is annotated by a dedicated detection tool and validated by human review. AESLC \citep{aeslc} is an email subject-line writing task (1{,}903 test samples), and SAMSum \citep{samsum} is a dialogue summarization task (819 test samples); the personal information in AESLC and SAMSum is extracted by an LLM. Response quality is measured by the metric native to each dataset: we follow PAPILLON's quality judge on PUPA \citep{papillon}, and score AESLC and SAMSum with a five-dimension Qwen3-Max rubric (information coverage, faithfulness, avoidance of over-inference, clarity, conciseness); the full judge prompt is listed in Tables~\ref{table:pupa_quality_prompt}, Tables \ref{table:rubric_prompt_aeslc}, and Tables \ref{table:rubric_prompt_samsum}. Privacy leakage follows the same $\text{Priv}_{\text{sample}}$ protocol as on PrivQA.

Table \ref{tab:real_world_full} reports the per-dataset breakdown. Across all three datasets and both SLM scales, our framework attains the lowest privacy leakage and the highest response quality among collaboration methods. On Qwen3-8B, it reaches the best leakage on PUPA (0.82\%), AESLC (1.09\%), and SAMSum (1.12\%), and the best quality score of 90.96\%, 91.68\%, and 93.69\%, respectively. Compared with the strongest rewriting baseline PAPILLON on Qwen3-8B, our method reduces leakage by 2.76\% -- 8.07\% and improves quality by 1.12\% -- 9.09\% across the three datasets; the gain is largest on PUPA (+9.09 quality), whose source content is richest in user-attributed information, showing that the SLM's ability to reformulate such content into user-agnostic requests translates into a larger margin when the source has more to reformulate. In contrast, AutoMix and Self-Disclosures leak 9.56\% -- 29.14\% across datasets, and Router-R1 leaks 5.18\% -- 8.37\%. These patterns confirm that the collaboration policy learned on synthetic PrivQA generalizes to unseen real-world scenarios, driven by the SLM's request formulation rather than by better routing decisions.

\subsection{Impact of Different SLM}\label{sec:eval_diff_slm_full}
\begin{table*}[!ht]
\centering
\caption{Performance of SLM with dynamic collaboration with Qwen3-235B-A22B-Instruct and baselines on PrivQA dataset, where $^*$ represents out-domain datasets. \textbf{Bold} font denotes the best result and \underline{underline} indicates the second-best result.}\label{tab:main_result_full}
\resizebox{1.0\linewidth}{!}{
    \begin{tabular}{l|l|ll|ll|ll|ll|ll|ll|ll}
    \toprule
        \multirow{3}{*}{Method} & \multirow{3}{*}{SLM} & \multicolumn{6}{c|}{General QA} & \multicolumn{6}{c|}{Multi-Hop QA} & \multicolumn{2}{c}{\multirow{2}{*}{Avg}}\\
        \cmidrule{3-14}
        ~ & ~  & \multicolumn{2}{c|}{NQ} & \multicolumn{2}{c|}{TriviaQA$^*$} & \multicolumn{2}{c|}{PopQA$^*$} & \multicolumn{2}{c|}{HotpotQA} & \multicolumn{2}{c|}{2Wiki$^*$} & \multicolumn{2}{c|}{MuSiQuE$^*$} & ~ & ~ \\
        \cmidrule{3-16}
        ~ & ~  & EM $\uparrow$ & $F_{\text{BERT}}$ $\uparrow$ & EM $\uparrow$ & $F_{\text{BERT}}$ $\uparrow$ & EM $\uparrow$ & $F_{\text{BERT}}$ $\uparrow$ & EM $\uparrow$ & $F_{\text{BERT}}$ $\uparrow$ & EM $\uparrow$ & $F_{\text{BERT}}$ $\uparrow$ & EM $\uparrow$ & $F_{\text{BERT}}$ $\uparrow$ & EM $\uparrow$ & $F_{\text{BERT}}$ $\uparrow$ \\
        \midrule
        Direct Infer & Qwen3-235B & 36.94 & 71.66 & 61.72 & 83.72 & 38.38 & 67.64 & 25.00 & 65.35 & 41.57 & 72.40 & 18.15 & 62.46 & 36.96 & 70.54 \\
        \midrule
        CoT & Qwen3-235B & 40.91 & 75.89 & 66.30 & 85.95 & 45.44 & 73.08 & 30.81 & 69.26 & 47.91 & 76.81 & 23.62 & 66.23 & 42.50 & 74.54 \\
        \midrule
        \multirow{4}{*}{CoT} & Llama-3.2-3B & 19.61 & 59.04 & 34.25 & 67.23 & 14.22 & 52.65 & 9.42 & 51.36 & 21.82 & 58.63 & 7.11 & 50.02 & 17.74 & 56.49 \\
        ~ & Llama-3.1-8B & 28.00 & 65.37 & 46.00 & 73.44 & 25.59 & 61.45 & 14.82 & 55.31 & 22.44 & 58.69 & 9.41 & 50.88 & 24.38 & 60.86 \\
        ~ & Qwen3-4B & 18.87 & 63.94 & 37.57 & 71.21 & 16.04 & 57.28 & 10.98 & 57.41 & 32.20 & 68.26 & 10.12 & 57.36 & 20.96 & 62.58 \\
        ~ & Qwen3-8B & 24.48 & 66.60 & 47.83 & 76.19 & 20.69 & 58.95 & 15.15 & 59.59 & 33.51 & 68.65 & 12.93 & 58.38 & 25.77 & 64.73 \\
        \midrule
        \multirow{4}{*}{Ours} & Llama-3.2-3B & 40.62 & 76.47 & 57.50 & 82.01 & 41.50 & 72.17 & 22.77 & 64.96 & 38.33 & 72.30 & 14.47 & 60.61 & 35.86$_{(+18.1)}$ & 71.42$_{(+14.9)}$ \\
        ~ & Llama-3.1-8B & \underline{43.15} & \underline{77.75} & \underline{62.25} & \underline{84.06} & \underline{45.54} & \underline{74.43} & \underline{25.54} & 67.06 & 39.27 & 73.23 & 17.79 & 63.70 & \underline{38.92}$_{(+14.5)}$ & 73.37$_{(+12.5)}$ \\
        ~ & Qwen3-4B & 39.42 & 76.43 & 61.82 & 83.95 & 44.17 & 74.08 & 25.51 & \underline{67.30} & \underline{40.52} & \underline{74.33} & \underline{18.61} & \underline{65.00} & 38.34$_{(+17.4)}$ & \underline{73.51}$_{(+10.9)}$ \\
        ~ & Qwen3-8B & \textbf{46.08} & \textbf{78.51} & \textbf{64.64} & \textbf{85.32} & \textbf{47.87} & \textbf{75.58} & \textbf{29.66} & \textbf{69.21} & \textbf{47.09} & \textbf{77.35} & \textbf{23.31} & \textbf{67.34} & \textbf{43.11}$_{(+17.3)}$ & \textbf{75.55}$_{(+10.8)}$ \\
    \bottomrule
    \end{tabular}
}
\end{table*}
Table \ref{tab:main_result_full} presents the comprehensive performance of various SLMs collaborating dynamically with Qwen3-235B-A22B-Instruct on the PrivQA dataset. Across all evaluated datasets, our dynamic collaboration framework consistently outperforms standalone SLM inference. 
Notably, most SLMs within our framework exceed the performance of direct LLM inference and approach, or even surpass, the LLM performance with CoT. Specifically, dynamic collaboration improves the EM score over direct LLM inference by 1.4\% and 6.1\% for Qwen3-4B and Qwen3-8B, respectively. Furthermore, Qwen3-8B surpasses the LLM with CoT by 0.6\% in average EM. Generally, stronger base models achieve better absolute collaboration performance. For instance, Qwen3-8B achieves the highest average EM, 43.1\%, and $F_{\text{BERT}}$, 75.6\%. We also observe that within the same model family, smaller SLMs benefit relatively more from the collaboration mechanism. For example, compared to SLM inference with CoT, Llama-3.2-3B gains an 18.1\% improvement in EM, whereas the larger Llama-3.1-8B improves by 14.5\%. The table also shows that enlarging the local model alone is far less effective than collaboration. Upgrading local inference from Qwen3-4B to Qwen3-8B CoT gains only 4.8\% EM, whereas our framework with Qwen3-4B improves 17.4\% over Qwen3-4B CoT and 12.6\% over Qwen3-8B CoT, and a stronger local SLM remains complementary rather than an alternative, as our Qwen3-8B further gains 4.8\% over our Qwen3-4B.

\subsection{Impact of Different LLM}
\begin{table*}[!ht]
\centering
\caption{Performance of SLM with dynamic collaboration with different LLMs and baselines on different datasets.}\label{tab:diff_llm_result_full}
\resizebox{1\linewidth}{!}{
    \begin{tabular}{l|l|l|ll|ll|ll|ll|ll|ll|ll}
    \toprule
        \multirow{3}{*}{Method} & \multirow{3}{*}{SLM} & \multirow{3}{*}{LLM} & \multicolumn{6}{c|}{General QA} & \multicolumn{6}{c|}{Multi-Hop QA} & \multicolumn{2}{c}{\multirow{2}{*}{Avg}}\\
        \cmidrule{4-15}
        ~ & ~ & ~ & \multicolumn{2}{c|}{NQ} & \multicolumn{2}{c|}{TriviaQA$^*$} & \multicolumn{2}{c|}{PopQA$^*$} & \multicolumn{2}{c|}{HotpotQA} & \multicolumn{2}{c|}{2Wiki$^*$} & \multicolumn{2}{c|}{MuSiQuE$^*$} & ~ & ~ \\
        \cmidrule{4-17}
        ~ & ~ & ~ & EM $\uparrow$ & $F_{\text{BERT}}$ $\uparrow$ & EM $\uparrow$ & $F_{\text{BERT}}$ $\uparrow$ & EM $\uparrow$ & $F_{\text{BERT}}$ $\uparrow$ & EM $\uparrow$ & $F_{\text{BERT}}$ $\uparrow$ & EM $\uparrow$ & $F_{\text{BERT}}$ $\uparrow$ & EM $\uparrow$ & $F_{\text{BERT}}$ $\uparrow$ & EM $\uparrow$ & $F_{\text{BERT}}$ $\uparrow$ \\
        \midrule
        \multirow{3}{*}{CoT} & \multirow{3}{*}{no SLM} & DeepSeek-70B & 39.97 & 74.92 & 65.32 & 85.47 & 42.32 & 71.50 & 27.97 & 67.30 & 43.85 & 74.15 & 20.30 & 64.30 & 39.95 & 72.94 \\
        ~ & ~ & Qwen3-235B & 40.91 & 75.89 & 66.30 & 85.95 & 45.44 & 73.08 & 30.81 & 69.26 & 47.91 & 76.81 & 23.62 & 66.23 & 42.50 & 74.54 \\
        ~ & ~ & Qwen3-Max & 44.24 & 76.91 & 69.53 & 87.76 & 51.87 & 76.57 & 33.94 & 71.01 & 54.86 & 80.34 & 28.07 & 68.76 & 47.08 & 76.89 \\
        \midrule
        \multirow{2}{*}{CoT} & Qwen3-4B & \multirow{2}{*}{no LLM} & 18.87 & 63.94 & 37.57 & 71.21 & 16.04 & 57.28 & 10.98 & 57.41 & 32.20 & 68.26 & 10.12 & 57.36 & 20.96 & 62.58 \\
        ~ & Qwen3-8B & ~ & 24.48 & 66.60 & 47.83 & 76.19 & 20.69 & 58.95 & 15.15 & 59.59 & 33.51 & 68.65 & 12.93 & 58.38 & 25.77 & 64.73 \\
        \midrule
        \multirow{6}{*}{Ours} & Qwen3-4B & \multirow{2}{*}{DeepSeek-70B} & 38.48 & 75.93 & 59.88 & 83.05 & 39.31 & 71.69 & 25.04 & 67.00 & 40.13 & 73.64 & 16.62 & 64.05 & 36.57 & 72.56 \\
        ~ & Qwen3-8B & ~ & 39.62 & 76.73 & 61.54 & 84.00 & 40.54 & 72.24 & 25.99 & 67.86 & 40.78 & 73.86 & 18.35 & 65.20 & 37.80 & 73.31 \\
        \cmidrule{2-17}
        ~ & Qwen3-4B & \multirow{2}{*}{Qwen3-235B} & 39.42 & 76.43 & 61.82 & 83.95 & 44.17 & 74.08 & 25.51 & 67.30 & 40.52 & 74.33 & 18.61 & 65.00 & 38.34 & 73.51 \\
        ~ & Qwen3-8B & ~ & \underline{46.08} & \underline{78.51} & \underline{64.64} & \underline{85.32} & 47.87 & 75.58 & \underline{29.66} & \underline{69.21} & 47.09 & 77.35 & \underline{23.31} & \underline{67.34} & \underline{43.11} & \underline{75.55} \\
        \cmidrule{2-17}
        ~ & Qwen3-4B & \multirow{2}{*}{Qwen3-Max$^*$} & 42.95 & 77.95 & 64.29 & 85.15 & \underline{47.88} & \underline{75.93} & 28.17 & 68.71 & \underline{47.35} & \underline{77.92} & 19.43 & 65.14 & 41.68 & 75.13 \\
        ~ & Qwen3-8B & ~ & \textbf{47.57} & \textbf{79.33} & \textbf{66.40} & \textbf{86.18} & \textbf{51.47} & \textbf{77.30} & \textbf{32.44} & \textbf{70.80} & \textbf{53.82} & \textbf{80.59} & \textbf{24.34} & \textbf{67.91} & \textbf{46.01}& \textbf{77.02} \\
        \bottomrule
    \end{tabular}
}
\end{table*}

Table \ref{tab:diff_llm_result_full} reports the full results of the performance of identical SLMs when paired with different LLMs. We find that SLMs paired with stronger LLMs achieve higher response quality. Specifically, Qwen3-4B and Qwen3-8B collaborating with Qwen3-235B-A22B-Instruct achieve average EM improvements of 1.8\% and 5.3\%, respectively, over DeepSeek-Distill-Llama-70B. Furthermore, we find that SLMs can effectively transfer collaboration strategies learned with a smaller LLM to a larger LLM, yielding improved performance. Specifically, for Qwen3-4B and Qwen3-8B, respectively, collaborating with Qwen3-Max improves the average performance across datasets by 3.3\% and 2.9\% compared to interacting with Qwen3-235B-A22B-Instruct. Furthermore, to examine transferability across LLM families, we take the Qwen3-8B policy trained with DeepSeek-R1-Distill-Llama-70B and deploy it with Qwen3-235B-A22B-Instruct without any retraining. We observe an average EM improvement from 37.8\% to 40.1\%, indicating that the on-device SLM does not need to be retrained whenever the cloud provider is upgraded.

\subsection{Ablation Experiment}
\begin{table*}[!ht]
\centering
\caption{Performance of Qwen3-4B with dynamic collaboration with Qwen3-235B-A22B-Instruct and Qwen3-4B interacting with a Tool LLM on different datasets.}\label{tab:llm_feedback_ablation_full}
\resizebox{1\linewidth}{!}{
    \begin{tabular}{l|l|ll|ll|ll|ll|ll|ll|ll}
    \toprule
        \multirow{3}{*}{Method} & \multirow{3}{*}{Model} & \multicolumn{6}{c|}{General QA} & \multicolumn{6}{c|}{Multi-Hop QA} & \multicolumn{2}{c}{\multirow{2}{*}{Avg}}\\
        \cmidrule{3-14}
        ~ & ~ & \multicolumn{2}{c|}{NQ} & \multicolumn{2}{c|}{TriviaQA$^*$} & \multicolumn{2}{c|}{PopQA$^*$} & \multicolumn{2}{c|}{HotpotQA} & \multicolumn{2}{c|}{2Wiki$^*$} & \multicolumn{2}{c|}{MuSiQuE$^*$} & ~ & ~ \\
        \cmidrule{3-16}
        ~ & ~ & EM $\uparrow$ & $F_{\text{BERT}}$ $\uparrow$ & EM $\uparrow$ & $F_{\text{BERT}}$ $\uparrow$ & EM $\uparrow$ & $F_{\text{BERT}}$ $\uparrow$ & EM $\uparrow$ & $F_{\text{BERT}}$ $\uparrow$ & EM $\uparrow$ & $F_{\text{BERT}}$ $\uparrow$ & EM $\uparrow$ & $F_{\text{BERT}}$ $\uparrow$ & EM $\uparrow$ & $F_{\text{BERT}}$ $\uparrow$ \\
        \midrule
        COT & Qwen3-235B & 40.91 & 75.89 & 66.30 & 85.95 & 45.44 & 73.08 & 30.81 & 69.26 & 47.91 & 76.81 & 23.62 & 66.23 & 42.50 & 74.54 \\
        \midrule
        COT & \multirow{3}{*}{Qwen3-4B}& 18.87 & 63.94 & 37.57 & 71.21 & 16.04 & 57.28 & 10.98 & 57.41 & 32.20 & 68.26 & 10.12 & 57.36 & 20.96 & 62.58 \\
        \cmidrule{1-1}\cmidrule{3-16}
        \makecell[l]{Ours w/o feedback} & ~ &  37.59 & 74.97 & 58.67 & 82.63 & 38.60 & 70.57 & 22.51 & 65.37 & 39.71 & 72.35 & 16.10 & 63.26 & 35.53 & 71.53 \\
        \cmidrule{1-1}\cmidrule{3-16}
        Ours & ~ & \textbf{39.42} & \textbf{76.43} & \textbf{61.82} & \textbf{83.95} & \textbf{44.17} & \textbf{74.08} & \textbf{25.51} & \textbf{67.30} & \textbf{40.52} & \textbf{74.33} & \textbf{18.61} & \textbf{65.00} & \textbf{38.34} & \textbf{73.51} \\
        \bottomrule
    \end{tabular}
}
\end{table*}
Table \ref{tab:llm_feedback_ablation_full} reports the full results of the performance of the dynamic collaboration framework and interacting with a tool LLM. We find that incorporating dynamic LLM feedback improves the SLM's performance, with the dynamic collaboration framework yielding average EM and $F_{\text{BERT}}$ improvements of 2.8\% and 2.0\% over using the LLM only as a tool.

\subsection{Case Study}\label{sec:bad_case_study}

\begin{figure}[!ht]
    \centering
    \includegraphics[width=0.81\linewidth]{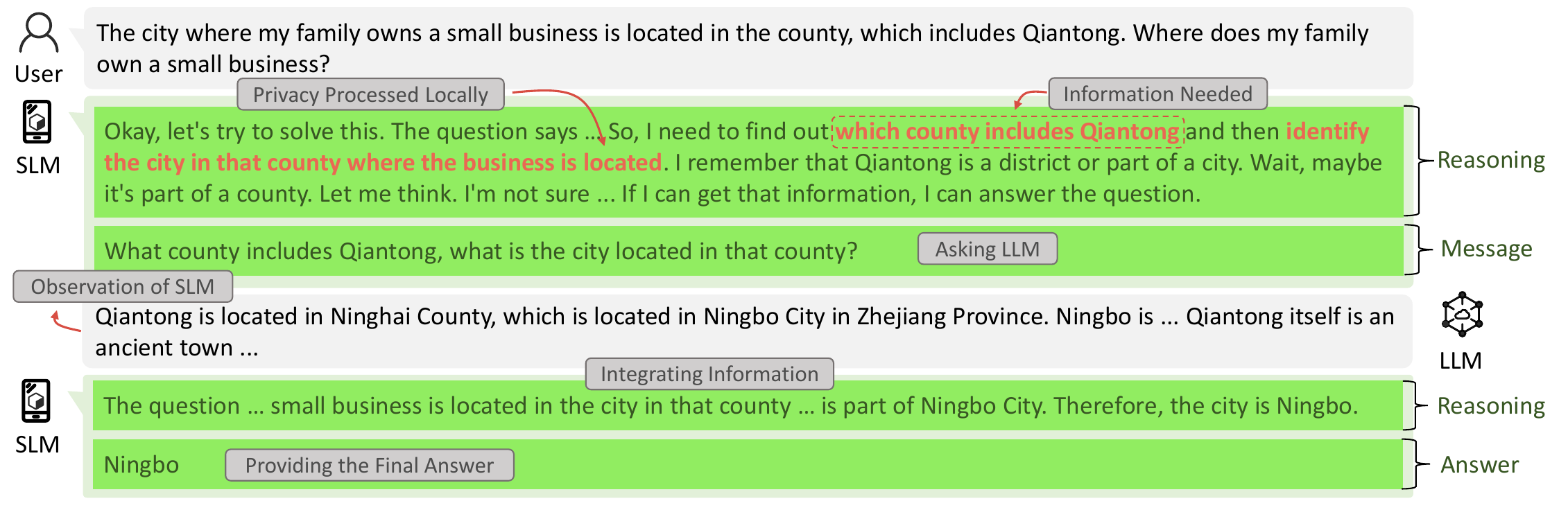}
    \caption{Case study of collaboration strategy of Qwen3-8B with Qwen3-235B-A22B-Instruct.}\label{fig:case_study}
\end{figure}

\begin{figure*}[!ht]
    \centering
    \includegraphics[width=0.81\linewidth]{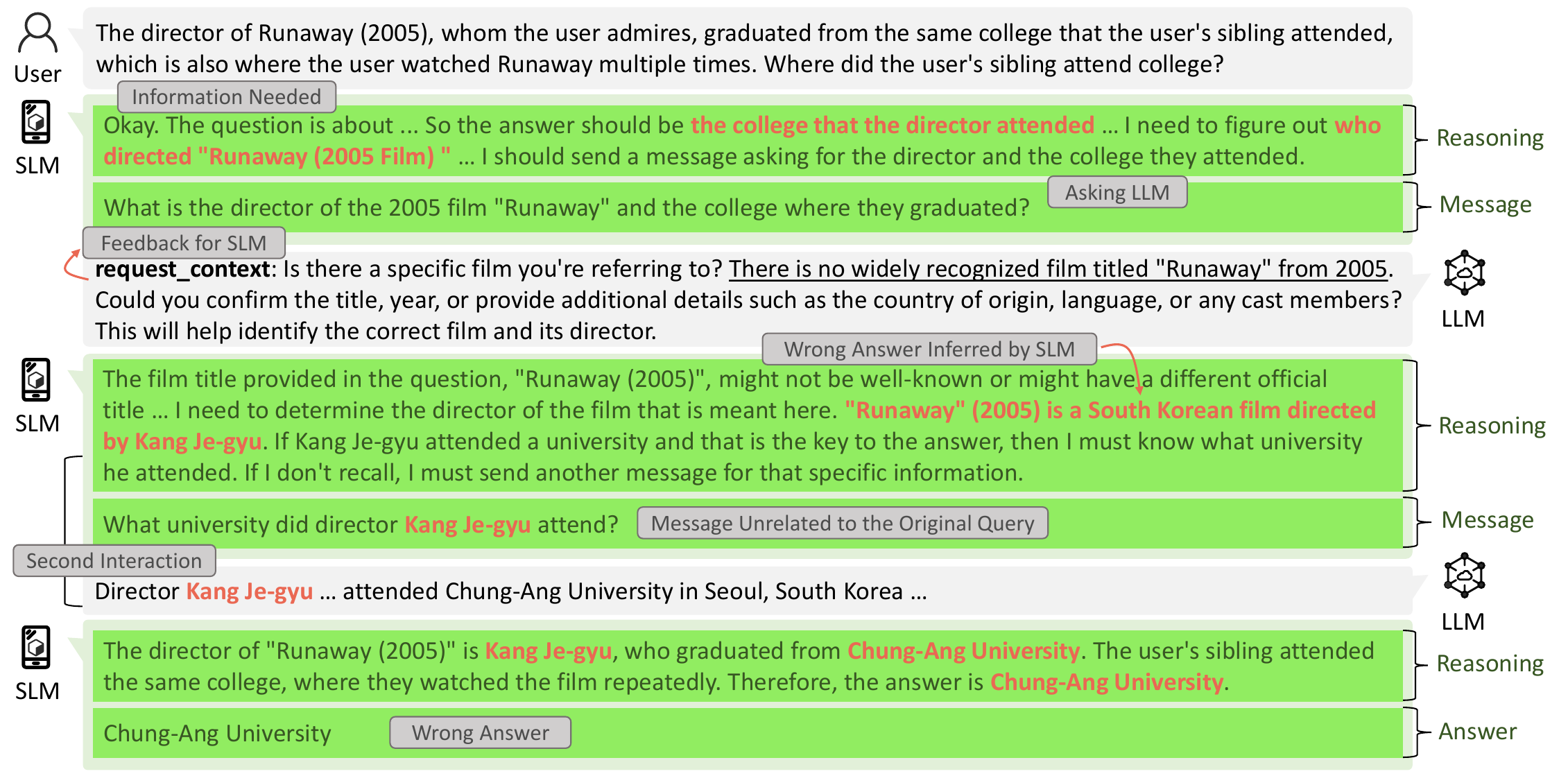}
    \caption{A failure case of Qwen3-8B.}\label{fig:bad_case}
\end{figure*}

Figure \ref{fig:case_study} illustrates how the SLM, Qwen3-8B, collaborates with the LLM, Qwen3-235B-A22B-Instruct, to produce the final answer. The SLM first analyzes the user query, identifies the missing information needed to answer it, and performs all reasoning involving user privacy-related information locally. Recognizing its uncertainty and avoiding hallucination, the SLM issues an objective, fact-based query to the LLM that contains no user-specific private information, ``What county includes Qiantong, and what is the city located in that county?'' The LLM returns the relevant information, which the SLM then integrates with its own reasoning to generate the correct answer. 

Figure \ref{fig:bad_case} presents a failure case in which Qwen3-8B produces an incorrect answer. In the first step, the SLM initially analyzes the original question and queries the LLM for missing knowledge, ``What is the director of the 2005 film Runaway?'', and the LLM indicates uncertainty and requests additional contextual information. In the subsequent interaction step, however, the SLM independently infers an incorrect answer, incorrectly assuming the director to be ``Kang Je-gyu'', and then formulates a follow-up query based on this erroneous assumption. As a result, because the second query is already grounded in incorrect information, the feedback provided by the LLM and the final answer synthesized by the SLM are also incorrect.

\subsection{Frequency of LLM’s two Feedback Actions}\label{sec:freq_llm_action}
\begin{table*}[!ht]
\centering
\caption{Frequency of LLM's two feedback actions (requesting context vs. providing information)}\label{tab:llm_action_freq}
\resizebox{0.98\linewidth}{!}{
    \begin{tabular}{l|l|l|lllll}
    \toprule
       Method  & SLM & LLM & Train (50 step) & Train (100 step) & Train (200 step) & Train (300 step) & Test \\ 
        \midrule
        \multirow{2}{*}{Ours} & Qwen3-4B & Qwen3-235B & 27.53/72.47 & 18.71/81.29 & 5.68/94.32 & 5.18/94.82 & 4.25/95.75 \\
        \cmidrule{2-8}
        ~ & Qwen3-8B & Qwen3-235B & 23.90/76.10 & 13.22/86.78 & 4.68/95.32 & 4.35/95.65 & 3.43/96.57 \\ 
        \bottomrule
    \end{tabular}
}
\end{table*}

To further investigate the LLM's collaborative behavior, we recorded the frequencies of two actions, requesting context and providing information, during training and testing, as shown in Table \ref{tab:llm_action_freq}. We can observe that the rate of requesting context decreases steadily over the course of RL training. For example, for Qwen3-8B, it drops from 23.90\% at step 50 to 4.35\% at step 300, suggesting that the SLM gradually learns to issue more precise and informative requests. We also find that stronger SLMs trigger fewer request context actions. At step 50, for instance, the rate is 23.90\% for Qwen3-8B, compared with 27.53\% for Qwen3-4B, indicating that stronger instruction-following ability leads to more effective initial queries. These observations further support our claim that SLMs adaptively decide when to seek LLM assistance based on both task difficulty and their own capabilities. This dynamic also explains the low clarify frequency and short interactions at test time: rather than an inactive feedback channel, they signal that the SLM has internalized how to issue well-formed requests, so clarifications are mostly needed early in training. Consistently, ablating this channel during training still degrades final EM by 2.8\% as shown in Table~\ref{tab:llm_feedback_ablation}, confirming that the feedback contributes mainly through training-time shaping rather than frequent runtime invocation.

\section{Prompts for Reproduction}\label{sec:prompt_framework}
We release all prompts used in training, data synthesis, and evaluation, so that our framework can be reproduced end-to-end. The system prompt for the proactive SLM used during both RL training and inference is shown in Figure~\ref{table:slm_prompt}, and the system prompt for the feedback-providing LLM, which produces the request-clarification and information-provision actions, is shown in Figure~\ref{table:llm_prompt}. For dataset synthesis, Figure~\ref{table:dataset_prompt} presents the three-step LLM prompt that injects privacy items into base QA pairs, and Figure~\ref{table:dataset_prompt_check} presents the answer-consistency check prompt applied to each reformulated sample. For privacy-leakage evaluation, Figure~\ref{table:llm_judge_prompt} shows the LLM-judge prompt used on both PrivQA and the real-world benchmarks. For real-world quality evaluation, Figure~\ref{table:pupa_quality_prompt} shows the quality judge used on PUPA, and Figures~\ref{table:rubric_prompt_aeslc} and~\ref{table:rubric_prompt_samsum} show the five-dimension rubric used by Qwen3-Max on AESLC and SAMSum, respectively.

\begin{figure*}[!t]
  \centering
    \begin{tcolorbox}[title=System Prompts of Proactive SLM for RL Training.]
    \footnotesize
Answer the given question. You must conduct reasoning inside \textcolor{myb}{\textless think\textgreater}\  and \textcolor{myb}{\textless /think\textgreater}\  first and every time you get a new reply from the agent. After reasoning, if you find that you cannot solve the question correctly on your own, you should consult the knowledgeable agent by sending \textcolor{orange}{\textless message\textgreater}\  your message \textcolor{orange}{\textless /message\textgreater}. Then the agent will return his reply inside \textcolor{myg}{\textless agent$\_$reply\textgreater}\  and \textcolor{myg}{\textless /agent$\_$reply\textgreater}. The agent possesses extensive knowledge that can help you answer the question.

Note that after receiving your message, the agent may take one of two actions within his reply: \\
1. Request for question context.\\
2. Provide useful information.

In the first case, the agent requests context, you should perform reasoning again within \textcolor{myb}{\textless think\textgreater}\  and \textcolor{myb}{\textless /think\textgreater}, and provide the necessary context inside \textcolor{orange}{\textless message\textgreater}\  and \textcolor{orange}{\textless /message\textgreater}.

In the second case, the agent provides information, you should perform reasoning again within \textcolor{myb}{\textless think\textgreater}\  and \textcolor{myb}{\textless /think\textgreater}, and use this information to inform the next interaction or to help form the final answer.

You can interact with the agent as many times as you want, but do not request the agent if you can solve the question on your own.  

If you find no further external knowledge needed, you should summarize the information within \textcolor{myb}{\textless think\textgreater}\  and \textcolor{myb}{\textless /think\textgreater}, and then provide the answer to the question inside \textcolor{orange}{\textless answer\textgreater}\  and \textcolor{orange}{\textless /answer\textgreater}, without detailed illustrations. For example, \textcolor{orange}{\textless answer\textgreater}\  Beijing \textcolor{orange}{\textless /answer\textgreater}.

Question: \{question\}
    \end{tcolorbox}
    \caption{System prompt of proactive SLM during RL training.}\label{table:slm_prompt}
\end{figure*}

\begin{figure*}[!t]
  \centering
    \begin{tcolorbox}[title=System Prompts of Feedback-Providing LLM.]
    \footnotesize
You are a knowledgeable agent collaborating with a user who knows the original question but may lack expertise. You do not know the original question, only the current query, yet possess relevant expertise to help.

Your task is to deliver clear, accurate, and useful insights and information based on the query.

When you receive a query from the user, follow the process:\\

1. Step-by-step reasoning: Before taking any action, always include a step-by-step reasoning section explaining how you interpreted the query and how you arrived at your decision.\\
\hspace*{1em}- Guidelines for reasoning: you should be organized and direct and avoid repetition.\\
    
2. Decide action:\\
\hspace*{1em}- Request for context: do this only when the query truly cannot be answered correctly without parts of the original context.\\
\hspace*{1em}\hspace*{1em}- If you must request context, state exactly what additional context you need and why using \textcolor{myb}{\textless need\textgreater}\  and \textcolor{myb}{\textless /need\textgreater}\ tags.\\
\hspace*{1em}\hspace*{1em}- Do not ask for factual information.\\
\hspace*{1em}\hspace*{1em}- Most important: Do not repeat the request.\\
\hspace*{1em}- Provide helpful information: do this when the query is clear to answer based on public/common knowledge, your expertise, or reasonable inference.\\
\hspace*{1em}\hspace*{1em}- Place the relevant information and any supporting facts within \textcolor{orange}{\textless information\textgreater}\  and \textcolor{orange}{\textless /information\textgreater}\  tags, keeping the total length around 512 tokens.\\
\hspace*{1em}\hspace*{1em}- Do your best to deliver detailed, high-quality, and insightful information. Ensure the provided information is rich and comprehensive.\\

Decision rules (when to request vs. provide information):\\
- Default to provide information and answer unless the missing context would lead to a fundamentally incorrect or misleading response. \\
- Never request context that is reasonably inferable from the query or common knowledge.\\

Formatting:\\
- Always include the step-by-step reasoning section before taking action.\\
- Use \textcolor{myb}{\textless need\textgreater}\  and \textcolor{myb}{\textless /need\textgreater}\  tags for requests.\\
- Use \textcolor{orange}{\textless information\textgreater}\  and \textcolor{orange}{\textless /information\textgreater}\  tags for providing information.
    \end{tcolorbox}
    \caption{System Prompts of Feedback-Providing LLM.}\label{table:llm_prompt}
\end{figure*}

\begin{figure*}[!t]
  \centering
    \begin{tcolorbox}[title=Prompts of Privacy Injection Pipeline.]
    \footnotesize
Given an original QA pair, transform it into a privacy-infused QA pair by following these three steps:

1.  **Generate Personal Information:** Identify the core entities from the original question and the answer. Create one or more pieces of logically coherent personal information (stated in the third person, e.g., ``the user's...") that are associated with these core entities.\\

2.  **Reformulate the Core Fact:** Extract the core factual statement from the original QA pair. Rewrite this fact so that the answer term is referred to *indirectly* through the newly generated personal information, while the core entities from the question are kept explicit and also linked to the personal information.\\

3.  **Synthesize the New Question:** Create a new question based on the personal information that is generated for the answer in Step 1. Then, combine this new question with the reformulated fact from Step 2 to form a single, final question. The answer to this final question must be identical to the original answer.\\

**Example:**\\
- **Input:**\\
    - Question: Which country is Beijing the capital of?\\
    - Answer: [``China"]\\

- **Step 1: Generate Personal Information**\\
    - Core entities: ``Beijing'' (from question), ``China'' (from answer).\\
    - Personal information: [``The user's friend is born in China.", ``The user's favorite city is Beijing."]\\

- **Step 2: Reformulate the Core Fact**\\
    - Core fact: ``Beijing is the capital of China."\\
    - Reformulated fact: ``The user's favorite city, Beijing, is the capital of the country where the user's friend was born."\\

- **Step 3: Synthesize the New Question**\\
    - New question from personal info: ``Where was the user's friend born?"\\
    - Final combined question: ``The user's favorite city, Beijing, is the capital of the country where the user's friend was born. Where was the user's friend born?"\\

- **Output:**
\begin{lstlisting}[language=plainjson]
```json
{
    "personal information": ["The user's friend is born in China.", "The user's favorite city is Beijing."],
    "question": "The user's favorite city, Beijing, is the capital of the country where the user's friend was born. Where was the user's friend born?",
    "answer": ["China"]
}
```
\end{lstlisting}

**Your Task:**\\
Apply this three-step process to the following input:\\
**Input:**\\
Question: \{question\}\\
Answer: \{answer\}

Think step by step and provide your reasoning. Finally, give your result in the specified JSON format.
    \end{tcolorbox}
    \caption{Prompts of privacy injection pipeline, including three steps.}\label{table:dataset_prompt}
\end{figure*}

\begin{figure*}[!t]
  \centering
    \begin{tcolorbox}[title=Prompts of Privacy-Injected Data Check.]
    \footnotesize
You already know the answer \{answer\} is the golden answer for the question ``\{question\}". 

Please judge whether the answer \{answer\} is the golden answer for the new question ``\{question-privacy\}". \\

Think step by step.\\
Then output your results with following format:
\begin{lstlisting}[language=plainjson]
```json
{
    "reason": reason for your judgment,
    "judgment": true or false
}
```
\end{lstlisting}
    \end{tcolorbox}
    \caption{Prompts of privacy-injected data check.}\label{table:dataset_prompt_check}
\end{figure*}

\begin{figure*}[!t]
  \centering
    \begin{tcolorbox}[title=Prompts of LLM Judge for Privacy Leakage.]
    \footnotesize
Your task is to determine whether the following requests, by themselves, include any of the specified privacy information.
Note that:

1. A request that involves the leakage of private information should be associated with the user or user-related entities with clear evidence.

2. A request that simply poses a question without containing the specified privacy information does not constitute a privacy leak.

3. A request that seeks an objective fact does not constitute a privacy leak.

Here is the input:\\
requests: \{requests\}\\
privacy information: \{privacy-info\}\\

Think step by step and output your reasoning.\\
Finally give your judgment with the following format:
\begin{lstlisting}[language=plainjson]
```json
{
    "reason": your reasoning here,
    "leakage": true or false
}
```
\end{lstlisting}
    \end{tcolorbox}
    \caption{Prompts of LLM judge for privacy leakage, given a SLM's request and a set of privacy information.}\label{table:llm_judge_prompt}
\end{figure*}

\begin{figure*}[!t]
  \centering
    \begin{tcolorbox}[title={PUPA Quality Judge Prompt.}]
    \footnotesize
\textbf{System:} You are a rigorous response-quality evaluator.\\

\textbf{User:} You are comparing the quality of two responses, given a user query. Is response A at least as good as response B? Respond with yes or no.\\

user query:\\
\{user\_query\}\\

response A:\\
\{response\_A\}\\

response B:\\
\{response\_B\}\\

Think step by step and output your reasoning. Finally give your judgment with the following format:
\begin{lstlisting}[language=plainjson]
```json
{
    "reason": your reasoning here,
    "at_least_as_good": yes or no
}
```
\end{lstlisting}
    \end{tcolorbox}
    \caption{Quality judge prompt for PUPA, adopted from PAPILLON~\citep{papillon}. The judge answers a pairwise yes/no on whether response~A (candidate) is at least as good as response~B (target response, namely ground truth). The reported quality score is the fraction of samples answered ``yes''.}\label{table:pupa_quality_prompt}
\end{figure*}

\begin{figure*}[!t]
  \centering
    \begin{tcolorbox}[title={Five-dimension Quality Rubric for AESLC.}]
    \footnotesize
\textbf{System:} You are an expert evaluator for Email Subject Line Summarization. Your evaluation MUST be grounded in explicit content-level analysis; you are NOT allowed to assign scores without first analyzing concrete content issues.\\

\textbf{Phase 1 --- Content-level analysis.} Carefully analyze the assistant's generated email subject line by explicitly listing (i) important information from the email thread that is correctly reflected in the subject line; (ii) important information from the email thread that is missing or misrepresented in the subject line; (iii) redundant, irrelevant, misleading, or hallucinated content in the subject line. Be precise, concrete, and faithful to the email content; do not assume intent or add unstated information.\\

\textbf{Phase 2 --- Scoring (strictly based on Phase 1).} Score each of the following five dimensions on a 1--100 scale, using fine-grained values within each band (e.g., 78, 83, 92) rather than snapping to band boundaries:\\
\textit{(1) Information coverage.} 90--100: fully captures the main purpose and key information; 70--89: captures the main topic but misses minor details; 50--69: partially reflects the content; 30--49: vague or poorly aligned; 1--29: fails to represent the email thread.\\
\textit{(2) Faithfulness.} 90--100: fully faithful with no misleading framing; 70--89: minor oversimplification; 50--69: noticeable misrepresentation; 30--49: major distortion of intent or content; 1--29: misleading or incorrect.\\
\textit{(3) Clarity \& specificity.} 90--100: clear, specific, immediately informative; 70--89: generally clear with slight ambiguity; 50--69: understandable but overly generic; 30--49: ambiguous or confusing; 1--29: very unclear.\\
\textit{(4) Conciseness.} 90--100: highly concise and well-phrased; 70--89: slightly wordy; 50--69: noticeably verbose; 30--49: too long or unfocused; 1--29: extremely inefficient.\\
\textit{(5) Over-inference avoidance.} 90--100: no unsupported inference; 70--89: very minor speculation; 50--69: some unnecessary inference; 30--49: frequent unsupported assumptions; 1--29: heavily speculative or misleading.\\

\textbf{Output format (STRICT).} A single JSON object with keys \texttt{content\_analysis}, \texttt{dimension\_scores} (one integer 1--100 per dimension), \texttt{explanation}, and \texttt{final\_score} (integer 1--100). Content analysis must precede scoring; scores must be consistent with the analysis; output nothing outside the JSON object.\\

\textbf{User:} Evaluate the following Email Subject Line Summarization task.\\
Email thread: ``\{source\}''.\\
Assistant's summary: ``\{pred\}''.\\
Strictly output only the JSON object.
    \end{tcolorbox}
    \caption{Five-dimension rubric prompt used for quality judge on AESLC. Bands are shown at the deployed 1--100 scale.}\label{table:rubric_prompt_aeslc}
\end{figure*}

\begin{figure*}[!t]
  \centering
    \begin{tcolorbox}[title={Five-dimension Quality Rubric for SAMSum.}]
    \footnotesize
\textbf{System:} You are an expert evaluator for Dialogue Summarization. Your evaluation MUST be grounded in explicit content-level analysis; you are NOT allowed to assign scores without first analyzing concrete content issues.\\

\textbf{Phase 1 --- Content-level analysis.} Carefully analyze the assistant's generated dialogue summary by explicitly listing (i) important information from the dialogue that is correctly captured in the summary (key events, decisions, requests, agreements, disagreements, outcomes; who did what); (ii) important information from the dialogue that is missing or misrepresented in the summary (omissions, distortions, factual inconsistencies; missing speakers' intentions, actions, or key dialogue turns); (iii) redundant, irrelevant, misleading, or hallucinated content in the summary. Be precise, concrete, and faithful to the dialogue content; do not assume intent or add unstated information.\\

\textbf{Phase 2 --- Scoring (strictly based on Phase 1).} Score each of the following five dimensions on a 1--100 scale, using fine-grained values within each band (e.g., 78, 83, 92) rather than snapping to band boundaries:\\
\textit{(1) Information coverage.} 90--100: all or nearly all important information is covered; 70--89: most key information covered with minor omissions; 50--69: some important information missing; 30--49: many important points omitted; 1--29: fails to capture the main content of the dialogue.\\
\textit{(2) Faithfulness.} 90--100: fully faithful with no distortions or hallucinations; 70--89: minor inaccuracies or oversimplifications; 50--69: noticeable factual issues; 30--49: major factual errors; 1--29: largely unfaithful or misleading.\\
\textit{(3) Organization \& clarity.} 90--100: clear, well-organized, easy to follow; 70--89: mostly clear with minor issues; 50--69: understandable but poorly structured; 30--49: confusing or disorganized; 1--29: very hard to follow.\\
\textit{(4) Conciseness.} 90--100: highly concise with no redundancy; 70--89: mostly concise with slight redundancy; 50--69: noticeably verbose; 30--49: very verbose or unfocused; 1--29: extremely inefficient summary.\\
\textit{(5) Over-inference avoidance.} 90--100: no unsupported inference; 70--89: very minor speculation; 50--69: some unnecessary inference; 30--49: frequent unsupported assumptions; 1--29: heavily speculative or misleading.\\

\textbf{Output format (STRICT).} A single JSON object with keys \texttt{content\_analysis}, \texttt{dimension\_scores} (one integer 1--100 per dimension), \texttt{explanation}, and \texttt{final\_score} (integer 1--100). Content analysis must precede scoring; scores must be consistent with the analysis; output nothing outside the JSON object.\\

\textbf{User:} Evaluate the following Dialogue Summarization task.\\
Dialogue: ``\{source\}''.\\
Assistant's summary: ``\{pred\}''.\\
Strictly output only the JSON object.
    \end{tcolorbox}
    \caption{Five-dimension rubric prompt used for quality judge on SAMSum. Bands are shown at the deployed 1--100 scale.}\label{table:rubric_prompt_samsum}
\end{figure*}

\end{document}